\documentclass[11pt]{article}

\usepackage[final]{acl}

\usepackage{times}
\usepackage{latexsym}

\usepackage[T1]{fontenc}

\usepackage[utf8]{inputenc}

\usepackage{microtype}

\usepackage{inconsolata}

\usepackage{graphicx}
\usepackage{float}
\usepackage{listings}
\usepackage{booktabs}
\usepackage{amsmath}
\usepackage{amsthm}
\usepackage{booktabs}
\usepackage{algorithm}
\usepackage{caption}
\usepackage{subcaption}
\usepackage{cuted}
\usepackage{capt-of} 
\usepackage{multirow}
\usepackage[noend]{algpseudocode}
\DeclareMathOperator*{\argmax}{arg\,max}
\title{In-Context Learning for Diagram Code Generation: What is the Effect on Faithfulness and Quality?}

\author{
 \textbf{Evanfiya Logacheva\textsuperscript{1}},
 \textbf{Arto Hellas\textsuperscript{1}},
 \textbf{Tsvetomila Mihaylova\textsuperscript{1}},
 \textbf{Juha Sorva\textsuperscript{1}},
\\
 \textbf{Ava Heinonen\textsuperscript{1}},
 \textbf{Juho Leinonen\textsuperscript{1}}
\\
\\
 \textsuperscript{1}Aalto University,
\\
 \small{
   {\{evanfiya.logacheva, arto.hellas, tsvetomila.mihaylova, juha.sorva, ava.heinonen, juho.2.leinonen\}@aalto.fi}
 }
}

\begin{document}
\maketitle
\begin{abstract}
We investigate how provision of in-context examples based on discourse similarity affects quality and hallucination in diagram code generation from educational materials. Using a sample of texts as sources for diagram generation, we test two methods using a conventional text-to-diagram task format, instruction-based zero-shot generation and ICL with text-to-diagram examples, and an unconventional ICL method with a discourse analysis-to-diagram example, which requires a model to generate a diagram from a discourse analysis. We evaluate AI-generated diagrams for faithfulness to provided context and their quality outcomes using both human and automated evaluation. Bayesian data analysis of the human diagram evaluation shows that there is significant evidence that the unconventional task format improves context faithfulness in diagram generation compared to the zero-shot instruction with reducing the odds of context faithfulness hallucination by nearly $70\%$; however, we do not find the ICL having a significant effect on the diagram aesthetic presentation and superficial coherence. Evaluation with the LLM-as-a-judge G-Eval metric shows nearly 4.5-fold increased odds of the perfect faithfulness outcome for the novel ICL task. Our study provides a novel insight into ICL format effect on superficial fluency and faithfulness in context-grounded structured diagram code generation.
\end{abstract}

\section{Introduction}
Diagram code generation is a multifaceted task for large language models (LLMs) as it often involves input context parsing into a semantically fluent presentation and generation of syntactically correct code, which makes it a structured output task~\cite{yang2026structeval}. While in-context learning (ICL) has been used in diagram generation applications~\cite{mondal-etal-2024-scidoc2diagrammer}, there has been limited research on how ICL affects diagram code generation compared to zero-shot generation and how variations in ICL formats shape LLMs performance for the task. 

Additionally, diagram code generation is often context-grounded, i.e., generated from a given context, which makes it vulnerable to both intrinsic and extrinsic hallucination \citep{lei2025, ravichander-etal-2025-halogen}. This presents a unique challenge for diagram generation in certain contexts such as education, as hallucination in generated learning materials presents a considerable risk to learners \citep{genaiedu}. While AI-generated diagrams may appear superficially acceptable, factual hallucination or unfaithfulness to course materials lead to irrelevant or inaccurate information presented to students, which is a highly undesirable behavior. 

To our best knowledge, there has been no dedicated study examining how ICL affects various quality dimensions and hallucination in diagram code generation, as any comprehensive examination requires costly human assessment with predefined criteria. We address the issue by introducing a diagram evaluation rubric for logical organization, connectivity, and layout quality used for expert diagram evaluation based on previous research on layout quality effect on diagram understanding~\cite{Störrle2011135, bergstrom2022evaluating}. Next, we examine each diagram for both intrinsic and extrinsic hallucination applying hallucination topology defined by \citet{lei2025}, revealing cases of factual hallucination and context unfaithfulness. Additionally, we measure G-Eval~\cite{liu-etal-2023-g} and Faithfulness~\cite{Ip_Vongthongsri_deepeval_2026} to automatically assess diagram context faithfulness and BERTScore~\cite{Zhang*2020BERTScore:} and ROUGE-L~\cite{lin-2004-rouge} to measure diagram text similarity to its input.

As established earlier by \citet{Jiang2023a, mondal-etal-2024-scidoc2diagrammer}, LLMs require an underlying framework for organizing nodes and edges of a diagram. We propose Rhetorical Structure Theory (RST)~\cite{MANNTHOMPSON+1988+243+281} as an organizational strategy as it has been earlier adopted for creation and analysis of multimedia content \citep{AndreRist1995, Bateman2001, Hiippala2020} and can be used for selecting documents based on their discourse similarity \citep{10.1162/COLI_a_00298}. Discourse structure representation with RST serves for both selection of relevant in-context examples through discourse structure similarity, which would guide LLMs towards the correct type of a diagram, and implicit demonstration of the hierarchical discourse structure a text should be mapped into. By providing a nested diagrammatic representation as an example, it gives a model a consistent pattern for ICL. 

To estimate how in-context examples affect diagram quality, hallucination, and overall faithfulness, we compare LLMs' output from three various input prompts: two variations with ICL examples and a baseline zero-shot method. All three receive a set of similar instructions but the in-context prompts are tested with various input formats. A standard format consists of a text-to-diagram example, while an unconventional one provides a discourse-to-diagram example. Both are compared to the zero-shot instruction-based generation across the quality, factual and context faithfulness outcomes using Bayesian models for estimating effect sizes of the prompts and model configurations as treatments. We used 150 diagrams generated with the o3 and GPT-4o models in our human assessment, which is almost double the size of the number of diagrams evaluated by human annotators in earlier research~\cite{mondal-etal-2024-scidoc2diagrammer}, in addition to 600 diagrams generated with the GPT-4.1 and GPT-5.6 models for automated assessment. In total, we used 10 model configurations in our assessment.

Our findings suggest that provision of examples significantly improves the fidelity of ICL-generated diagrams to provided context expressed as both the reduction of context inconsistency hallucination, higher G-Eval faithfulness scores, and improved textual similarity. The effect is significantly more pronounced for the unconventional format of discourse-to-diagram. We also observe that a more advanced model, o3~\cite{openai2025competitiveprogramminglargereasoning}, offers a significantly better performance with ICL for the novel task format compared to GPT-4o, which might be due to it possibly having a higher task diversity during its pretraining and thus being able to generalize on the task that the other model, GPT-4o, fails on~\cite{pmlr-v267-goddard25a}. Our results show that prompting formats and task definitions produce varying effects on different aspects of diagram code generation with zero-shot generation being the least faithful to context and most riddled with factual hallucination. The Bayesian analysis shows that the unconventional discourse-to-diagram mapping certainly has a positive impact on diagram faithfulness.

\section{Background}
\subsection{Diagram Code Generation with LLMs}
Since diagram code generation requires a model to appropriately translate the underlying structural system of a concept it represents, it often requires an intermediate approach for extracting relevant information and allocating it into nodes, which are then connected based on some principles. Recent works for diagram code generation with Mermaid.js and GraphViz use such techniques as entity and relation extraction for flowchart generation in a chatbot~\cite{Jiang2023a} and information extraction using a user's intent alongside in-context learning (ICL)~\cite{mondal-etal-2024-scidoc2diagrammer} for science diagram construction. 

Automated assessment of various features of AI-generated diagrams remains challenging. Existing metrics measuring logical organization have low to moderate correlation with human evaluation~\cite{liang-you-2025-evaluating}, with CLIPScore being negligibly so ($r\approx0.08-0.11$). Automated evaluation of layout quality is similarly uncertain when it comes to its alignment with human assessment, as ~\citet{mondal-etal-2024-scidoc2diagrammer} reported up to a one point difference between human and AI evaluation of diagram layout quality (on a 5-point scale), which makes it challenging to assess its reliability. \citet{Jiang2023a} report a user survey on their interactive diagram tool implemented in a chat setting, which offers individual insights but does not scale well. Human perception of  diagram layout aesthetics is largely subjective and requires a curated rubric for any meaningful assessment~\cite{bergstrom2022evaluating}.
\subsection{Rhetorical Structure Theory}
RST was initially developed for automatic generation of texts \cite{Taboada2006}, making it a promising approach for automatic generation of diagrams. The theory explains text coherence through hierarchical relations between parts of a text \cite{MANNTHOMPSON+1988+243+281,Taboada2006}. In an RST analysis, a coherent piece of text can be represented as a tree whose leaves are ``elementary discourse units'' (EDUs) connected with coherence relations \cite{stede2017annotation}. Relations vary depending on a specific RST framework \cite{Taboada2006}, and in this work, we chose the RST annotation guide by Stede et al. \cite{stede2017annotation}. Their approach separates relations into multinuclear and mononuclear, where the latter can be pragmatic, semantic, or textual. Multinuclear relations consist of two or more nuclei, whereas mononuclear ones have only one nucleus. A nucleus is the main central unit of a span, and a satellite is secondary \cite{stede2017annotation}. 

\citet{AndreRist1995} and \citet{Bateman2001} apply the RST for generation of coherent multimedia presentations.  \citet{AndreRist1995} argue that for sufficient understanding of a multimedia document, one needs to analyze relations between visual elements and between visual and textual elements. In their approach, they utilize rhetorical structure for planning multimedia presentations. \citet{Bateman2001} also note that visual and textual parts need to work in synergy rather than being simply placed in proximity. They demonstrate that the coherent expression of communicative intent and layout can be achieved by designing the latter on the basis of an RST-organized scheme. RST has been adopted for multimodal diagram annotation~\cite{Hiippala2020}. Similarly to those in texts \cite{stede2017annotation}, discourse relations in diagrams can be asymmetric (mononuclear) and symmetric (multinuclear) \cite{Hiippala2020}. 
\subsection{In-context Learning}
ICL has emerged as a popular method to teach a model to perform complex tasks without fine-tuning and updating model parameters~\cite{minaee2024largelanguagemodelssurvey, min2022}. The performance of ICL relies on the ability of LLMs to overcome the distributional mismatch between their pretraining prior and a given prompt, which is achieved when examples within the prompt are informative enough for a model to infer concepts present in its pretraining \citep{xie2022an}. ICL success also depends on whether a given task itself is in or out of distribution (OOD) of its pretraining task types, failing when LLMs cannot retrieve a relevant type \citep{wang2025can}. Additionally, their performance for OOD tasks depends on the pretraining task diversity of a model, as higher diversity allows for OOD generalization \citep{pmlr-v267-goddard25a}. When faced with OOD prompts, models tend to hallucinate \citep{kalai2025languagemodelshallucinate}, which makes ICL application for novel tasks complicated. 

\citet{xie2022an} propose that ICL functions as implicit Bayesian inference as follows,
\begin{equation}\label{icl_bayes}
    p(O|P)=\sum_cP(O|C,P)p(C|P)d(C)
\end{equation}
where $O$ is output, $P$ is a prompt, $C$ is a concept. For a model to infer a shared concept from the prompt examples, they need to be of sufficient length and clarity. Empirical studies show that ICL suffers from positional bias \citep{cobbina-zhou-2025-show} and vulnerability to prompt perturbations \cite{chen-etal-2023-relation}, which means that simply increasing the number of examples does not guarantee an improvement and makes their selection and placement important. While ICL can perform on new data, it struggles to overcome its pretraining \citep{kossen2024incontext}. 

\subsection{Taxonomy of Hallucination in LLM-Generated Content}\label{taxonomyhal}
Hallucination in LLM-generated content can be defined as misalignment with provided input context (intrinsic) or factual knowledge (extrinsic) \cite{lei2025, ravichander-etal-2025-halogen}. A more detailed taxonomy by \citet{lei2025} distinguishes between factuality and faithfulness hallucination. The first consists of factual contradiction, which is when a model's output contains a fact that contradicts world knowledge, and factual fabrication, which constitutes cases when we cannot verify facts present in generated content against existing information. Faithfulness hallucination can be of three subtypes: instruction inconsistency in case of deviation from user instructions; context inconsistency in the event of misalignment with input context; and logical inconsistencies for when a model's output internally contradicts itself.

\section{Proposed Method}
Our ICL method relies on LLM-generated RST analysis and the inclusion of examples for diagram code generation\footnote{The implementation can be found at https://github.com/evalog-a/llm-diagram-generator}. We implemented two variations (RST1 and RST2) and compare them to a zero-shot (0-shot) pipeline (Appendix E). Our examples for diagram generation from an input text are not simple format mappings between an input and its label as they represent a complex nested structure, mirroring the hierarchical discourse organization of their textual sources. As RST provides an encompassing taxonomy of existing relations, it allows us to identify similar texts based on their discourse structure, e.g., texts with multinuclear sequential or list relations. Using discourse structure to measure similarity is preferable since surface-level semantic similarity does not match texts based on their underlying organization \citep{10.1162/COLI_a_00298}. Discourse hierarchies signal the type of visualization appropriate for their structure: flowchart-type diagrams displaying mainly sequential relations, e.g., algorithms, and tree-types representing lists of items \citep{Tversky_2012, Tversky_2014, Hiippala2020}. Our examples also include complex, combined types, which often include elaborations (see Appendix B). The examples include discourse cues in the comments to help LLMs correctly visualize relations without explicitly mentioning RST. 

The two variations, RST1 and RST2, require LLMs to perform fundamentally different tasks: while RST1 includes a pairing of a simple text with its diagram code, RST2 provides a model with an RST analysis instead. We argue that while LLMs have seen in their pretraining educational materials, GraphViz code, and RST corpora, the specific pair `RST analysis' $\rightarrow$ `diagram code' requires a model to perform a new mapping. We compare the ICL to the 0-shot generation, where a model receives nearly identical set of instruction but without a mention of educational purposes and examples to establish a baseline. The instructions focus on visualization design principles that are later used in our diagram quality evaluation criteria (Eq. \ref{c3}) based on the diagram aesthetic research by \citet{Störrle2011135, bergstrom2022evaluating}. We do not include any explicit guidelines on context faithfulness or factual hallucination. For those diagrams that were evaluated by experts, we initially included a self-refinement step, where models were given a rendered diagram and asked to correct any present layout issues. This step was omitted in the generation of the diagrams that were automatically evaluated for faithfulness and similarity to prevent LLMs from introducing any possible modifications to texts within diagrams.

\section{Evaluation}
We investigated the following research questions. 
\textbf{RQ1:} How do in-context examples impact diagram quality in terms of logical organization, connectivity, layout aesthetics compared to zero-shot generation?
\textbf{RQ2:} How do in-context examples impact diagram context faithfulness compared to zero-shot generation?
\subsection{Human Evaluation Methodology}\label{methods}
  The authors manually evaluated the generated RST analyses and the 150 generated diagrams according to a quality rubric (Appendix C). The previous work by \citet{mondal-etal-2024-scidoc2diagrammer} used 80 diagrams in human evaluation in comparison. In total, 25 source texts were used for generation, each for 6 diagrams: the 3 methods $\times$ 2 models\footnote{The resulting dataset is openly available at https://github.com/evalog-a/diagram-evaluation}. The 0-shot diagrams were generated from the identical prompt consisting of a short instruction and a source text. For RST1 and RST2, RST analyses and examples were model specific, i.e., an analysis generated by each tested model was specific to that model. The cost of the human evaluation was high, with the total of 100 person-hours. The rubric consists of three main criteria graded on a 5-point ordinal scale, $F = \{1, 2, 3, 4, 5 \}$. A $C_{1}$ (logical organization) score is defined as follows,
\begin{equation}\label{c1}
C_1=[0.6 \times L_{1} + 0.3 \times L_{2} + 0.1 \times L_{3}]
\end{equation}
where $L_{1}$ is a diagram's flow and structure being logical, $L_{2}$ is its language clarity, $L_{3}$ is its adherence to flowchart conventions ; $L_1, L_2, L_3\in F$. $C_{1}$ is rounded half to odd. A $C_{2}$ (connectivity) score is assigned holistically, based on the diagram's uniformity and cohesion across connections between its elements, e.g., the score is reduced for orphan nodes. $C_{3}$ (layout aesthetic) assessment is based on prior research on graph aesthetic~\cite{bergstrom2022evaluating,Störrle2011135}; where a $C_{3}$ score is as follows,
\begin{equation}\label{c3}
C_3=
\begin{cases}
=5, \sum_{n=1}^{7}k_n<1\\
=4, 1\leq\sum_{n=1}^{7}k_n<3\\
=3, \sum_{n=1}^{7}k_n=3\\
=2, \sum_{n=1}^{7}k_n=4\\
=1, \sum_{n=1}^{7}k_n>4
\end{cases}
\end{equation}
where $k_n\in\{0,1\}$. $C_3$ is determined by the presence ($k_n=1$) or absence ($k_n=0$) of the following: ($k_1$) excessive line crossings or bends; ($k_2$) elements being obscured; ($k_3$) elements being incomprehensible due to their colors, size, or shape; ($k_4$) asymmetry; ($k_5$) vertical and horizontal misalignment; ($k_6$) excessive width; ($k_7$) dishomogeneity in appearance. As a unified quality metric, we use $G$, defined as follows,
\begin{equation}\label{g_outcome}
G=
\begin{cases}
=1, C_1, C_2, C_3  > 3\\
=0, C_1, C_2, C_3  \leq 3\\
\end{cases}
\end{equation}
The rubric evaluation was performed by 2 raters per each diagram, and its reliability estimated with Krippendorff's $\alpha$ and Kendall's $W$ coefficients for IRR for ordinal data~\cite{RJ-2021-046,kendallW,gibbons1993}. 

In addition to the diagram quality, instances of both factual and faithfulness hallucination were noted. We provide a detailed description of each type in Appendix G, but for our analysis, we namely report $F_{fact}$, which is factual fabrication or contradiction, and $F_c$, which is deviation from a given source text, which includes a model either omitting or adding information not present in a context given to it. Both those metrics are binary, i.e., we do not distinguish between a diagram containing several inaccuracies versus only one, both are marked as a $1$. No hallucination is denoted as a $0$.
\subsection{Automated Evaluation}
We generated additional 600 diagrams with same set of the source texts as in Section 4.1. We used 8 model configurations with and without reasoning effort, with models of various sizes. To account for any variability in LLM performance for RST analysis and similarity, we used the same analyses and chosen examples for RST1 and RST2 generated with GPT-5.6 Sol for every input instance, so each model received exactly the same prompt message and query per method\footnote{The diagrams evaluated by the human experts received varied inputs per each model.}.

As \citet{mondal-etal-2024-scidoc2diagrammer} used LLM-as-a-judge for assessment of completeness, faithfulness, and layout, we attempted to evaluate the $G$ outcome using the same criteria as the human evaluators. We provided GPT-5.6 Sol on the `high' reasoning setting with a prompt generated by the o3 model containing instructions generated from the rubric, a rendered diagram image, and its corresponding source text. The model was instructed to output 3 numerical scores for $C_1, C_2, C_3$, which were then used to calculate $G$ as defined in Eq. \ref{g_outcome}.

Available deterministic metrics and LLM-as-a-judge approaches for measuring similarity and faithfulness are text-based, and since structured outputs such as diagram DOT code contain text, we base our faithfulness evaluations not on LLM-generated captions of diagrams but on actual text in diagrams. In the evaluation, a diagram DOT snippet was given to GPT-5.6 Luna with a simple instruction to extract any text within the diagram's nodes and organize it following the edges between them. These re-constructed texts were compared to the inputs that the LLMs received during the generation process, where the 0-shot and RST1 cases used the original source texts. For RST2, the models received the RST analyses during the generation, which is why we asked GPT-5.6 Luna to extract EDUs from the analyses that were then compared to the texts obtained from the generated DOT.

Using the contexts as references and their corresponding extracted texts from the diagrams' DOT snippets, we evaluated faithfulness of the generated diagrams using G-Eval~\cite{liu-etal-2023-g} on a scale of $[0,1]$ implemented by ~\citet{Ip_Vongthongsri_deepeval_2026}. We additionally used DeepEval's Faithfulness metric that measures the proportion of truthful claims in LLM output over the number of claims in provided context~\cite{Ip_Vongthongsri_deepeval_2026}, alongside BERTScore~\cite{Zhang*2020BERTScore:} and ROUGE-L~\cite{lin-2004-rouge}, which do not measure straightforwardly context faithfulness but offer insight into overall similarity between texts within diagrams and their source contexts. 

We do not compare the human evaluation of faithfulness hallucination rates to the automated scores due to them measuring related but different concepts. Our human evaluation does not measure the proportion of faithfulness treating any unfaithful instance within a diagram as a hallucination, which means that an otherwise faithful representation gets flagged as a positive hallucination instance if it contains any claim that is not present in its source context, whereas G-Eval and Faithfulness scores represent a proportion $[0,1]$, which is challenging to translate into human assessment. Practically, it means that it is difficult to empirically locate a threshold of G-Eval and Faithfulness that corresponds to the human assessment of hallucination.

\subsection{Bayesian Data Analysis}
To estimate the effects of the methods ($p_{me}$) and models ($p_{mo}$) on the diagrams' hallucination and quality, we fit 3 Bayesian generalized linear mixed models (GLMMs) with random intercepts for each source text $s_i$ with two weakly informative priors~\cite{gelman2008weakly} for a prior sensitivity comparison (details in Appendix D and E).  The GLMMs use the linear predictor as follows,
\begin{equation}\label{model}
\begin{split}
     \eta_{wi} = \beta_0+\sum_{t\in P}\beta_{p_t}X_{p_twi}+u_i, \\u_i\sim N(0, \sigma^2),\\
     P=\{p_{me},p_{mo}\}\\
     P(Y_{wi}=1)=\operatorname{logit}^{-1}(\eta_{wi}),\\
     \text{ for binary outcomes:} \quad F_{fact}, F_c, G. 
\end{split}
\end{equation}
To estimate the effect of the $p_{me}$ and $p_{mo}$, we model the G-Eval as an outcome variable, which a proportion, with a zero-one inflated beta regression~\cite{ospina2010inflated} defined in \ref{zoib}, where $\gamma$ was an intercept only term and $\eta^{zoi}_{wi}$ shares the same structure as $\eta_{wi}$ in the Bernoulli models in \ref{model}.
\begin{equation}\label{zoib}
\begin{split}
     Y_{wi} \sim 
     \begin{cases}
         \alpha_i (1-\gamma) & =0\\
         \alpha_i \gamma & =1\\
         (1-\alpha_i)\text{Beta}(Y_{wi}, \mu_i, \phi)& (0,1)
     \end{cases}\\     
     \mu_i=\operatorname{logit}^{-1}(\eta_{wi}),\\ \alpha_i=\operatorname{logit}^{-1} (\eta^{zoi}_{wi}),\\ \gamma=\operatorname{logit}^{-1}(c), \phi=\exp(c_{\phi})\\
\end{split}
\end{equation}

The priors are as follows,
\begin{equation}\label{priors1}
\begin{split}
    Pr_{1a}: \beta_0\sim N(0,3), \beta_{p_t}\sim N(0,2),\\ \sigma_u \sim Exp(1)\\
    Pr_{1b}: \beta_0^{\mu}\sim N(0,3), \beta^{\mu}_{p_t}\sim N(0,2),\\ \sigma_u^{\mu} \sim Exp(1), \beta_0^{\alpha}\sim N(0,3),\\ \beta_{p_t}^{\alpha}\sim N(0,2),\sigma_u^{\alpha} \sim Exp(1),\\ \beta_0^{\gamma}\sim N(0,3)\\
\end{split}    
\end{equation}
\begin{equation}\label{priors2}
\begin{split}
    Pr_{2a}: \beta_0\sim N(0,3), \beta_{p_t}\sim Cauchy(0,1),\\ \sigma_u \sim Exp(1)\\
    Pr_{2b}: \beta_0^{\mu}\sim N(0,3), \beta_{p_t}^{\mu}\sim Cauchy(0,1),\\ \sigma_u^{\mu} \sim Exp(1), \beta_0^{\alpha}\sim N(0,3),\\ \beta_{p_t}^{\alpha}Cauchy(0,1), \sigma_u^{\alpha} \sim Exp(1),\\ \beta_0^{\gamma}\sim N(0,3)\\
\end{split}    
\end{equation}
The $\sigma_u$ is regularizing in \ref{priors1} and \ref{priors2} as some texts produce noticeably higher or lower rates in the outcomes, e.g., some texts tend to produce more contextually faithful diagrams. Due to the lack of prior information on hallucination rates, G-Eval, or quality effects on diagram code generation with ICL, $\beta_{p}$ in $Pr_{1a}, Pr_{1b}$ are rather diffuse. We compare it to $Pr_{2a}, Pr_{2b}$ with $Cauchy(0,1)$ to allow for larger effects while reducing false positives for small effects. 
\subsection{Experimental Setup}
The example demonstrations and generated diagrams were produced by selecting a set of educational texts licensed under CC BY 4.0, CC BY-SA 3.0, or with permission of copyright holders. The methods described in Subsections 3.2 and 4.1 were tested with `gpt-4o-2025-03-26' and `o3-2025-04-03' with the default temperature, where each model was used throughout all the steps, except for GPT-4o only for code repair\footnote{Bugs were extremely infrequent in generated dot code.}. For those diagrams evaluated automatically as described in Section 4.2, we used the GPT-5.6 family of models, Sol, Terra, Luna, and the pre-reasoning GPT-4.1 models: `gpt-4.1-2025-04-14' and `gpt-4.1-mini-2025-04-14'. For possible code repairs, we used GPT-5.6 Luna. Each was tested with the reasoning effort set to `high' and `none' for the GPT-5.6 models; the GPT-4.1 models do not have a reasoning effort. 

During prompt engineering, we used a set of short LLM-generated texts for tuning instructions. The IRR metrics were calculated with the R language implementation by \citet{RJ-2021-046} for Krippendorff's $\alpha$ and by \citet{kendallW} for Kendall's $W$. The Bayesian models were implemented in R with the package `brms' by \citet{JSSv100i05}. The G-Eval and Faithfulness scores were calculated in Python with `DeepEval' by \citet{Ip_Vongthongsri_deepeval_2026} with GPT-5.4. The BERTScore~\cite{Zhang*2020BERTScore:} was obtained with `roberta-large-mnli'~\cite{liu2019roberta}. 
\section{Results}

\subsection{Human Evaluation}

\begin{table*}[h!]
\caption{Human Evaluation Results for Logical Organization $C_1$, Connectivity $C_2$, Layout Aesthetic $C_3$. Q1, Q2, and Q3 stand for quartiles. Krippendorff's $\hat{\alpha}$ with $95\%$ CI and Kendall's $W$ with $p-$values are provided for IRR.}
    \label{tab:humaneval}
    \centering
    \begin{tabular}{lrrrrrrrrr}
    \toprule
     & \multicolumn{3}{c}{\textbf{$C_1$}}& \multicolumn{3}{c}{\textbf{$C_2$}}& \multicolumn{3}{c}{\textbf{$C_3$}}\\
     \midrule
 {IRR}& \multicolumn{3}{r}{$\hat{\alpha}=$ 0.83 (0.75, 0.89)}& \multicolumn{3}{r}{$\hat{\alpha}=$ 0.68 (0.44, 0.82)}& \multicolumn{3}{r}{$\hat{\alpha}=$ 0.60 (0.46, 0.71)}\\
 & \multicolumn{3}{r}{$W=$ 0.91 $(<0.01)$}& \multicolumn{3}{r}{$W=$ 0.79 $(<0.01)$}& \multicolumn{3}{r}{$W=$ 0.82 $(<0.01)$}\\
 \hline
    \end{tabular}
\end{table*}
\begin{table*}[ht!]
\caption{Hallucination occurrence and the ratio of $G$ (high quality diagrams).}
    \label{tab:human_results}
    \centering
\begin{tabular}{lllllll}
\toprule
&\multicolumn{3}{c}{GPT-4o}&\multicolumn{3}{c}{o3}\\
\cmidrule{2-4}\cmidrule{5-7}
&0-shot& RST1& RST2&0-shot& RST1& RST2\\
\midrule
$F_{fact}$ &0.2&  0.2&  0.04&0.12& 0.04& 0.04\\
$F_{c}$ &0.68&  0.52&  0.48&0.6& 0.36& 0.28\\
 $G$ &0.64& 0.6&0.4&0.76& 0.84&0.8\\
\bottomrule
\end{tabular}
\end{table*}
Table \ref{tab:humaneval} includes the IRR metrics: the Krippendorff’s $\hat{\alpha}$ show that $C_1$ evaluation is within nearly perfect agreement, while the corresponding estimates for $C_2$ and $C_3$ indicate substantial and moderate agreement respectively \cite{RJ-2021-046}; Kendall’s $W$ for $C_1,C_2,C_3$ suggest statistically significant agreement between the raters \cite{gibbons1993}. Lower agreement between raters in aesthetic evaluation of diagrams is documented in earlier research and stems from inherent subjectivity in human assessment~\citep{bergstrom2022evaluating}. Overall, the expert evaluation is most consistent for logical organization and acceptable for connectivity and layout quality. We cannot compare our IRR results to earlier research on AI-generated diagrams as \citet{mondal-etal-2024-scidoc2diagrammer} do not report IRR metrics. Our human evaluation discovered that the same concept can be accurately represented in an undefined number of diagram variations.

Table \ref{tab:human_results} displays the results of the expert evaluation for diagram quality and hallucination, which suggest that o3 outperforms GPT-4o across all the methods and metrics. There is a difference in performance between RST1 and RST2 with GPT-4o, where the latter offers a lower performance for $G$, which is not pronounced for o3; 0-shot generation outcomes are not affected by the model selection to the same degree.  Overall, the AI-generated diagrams scored high across all the criteria, except for RST2 in case of GPT-4o. All the methods and models produced content with factual hallucinations, with RST2 producing the least. RST1 and RST2 produced consistently lower rates of $F_c$. Faithfulness hallucination rates differ between the models: GPT-4o performs worse overall, especially for RST2; for o3, RST1 shows the best outcome. Our analysis also showed that the models cannot reliably identify and repair layout issues in rendered diagrams, often introducing additional problems: GPT-4o failed in up to $84\%$ of the RST1 diagrams and o3 in $52\%$ in the 0-shot and RST1 cases.
\subsection{Automated Evaluation}
\begin{figure}[ht]
    \centering
    \includegraphics[width=1\linewidth]{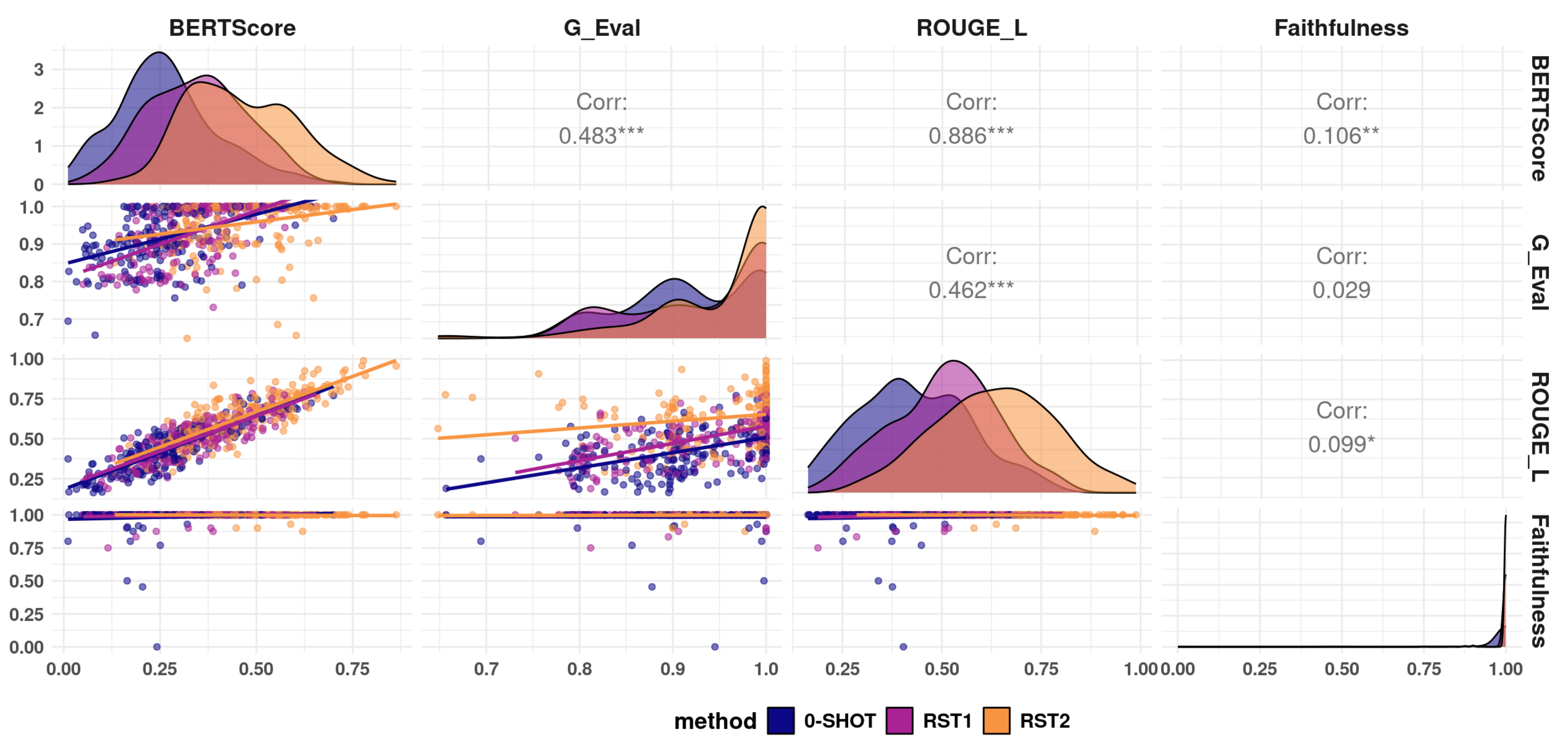}
    \caption{Distribution and correlation between BERTScore, G-Eval, ROUGE-L, Faithfulness metric. Asterisks denote statistical significance.}
    \label{fig:distributions}
\end{figure}
\begin{figure}[ht]
     \centering
     \includegraphics[width=1\linewidth]{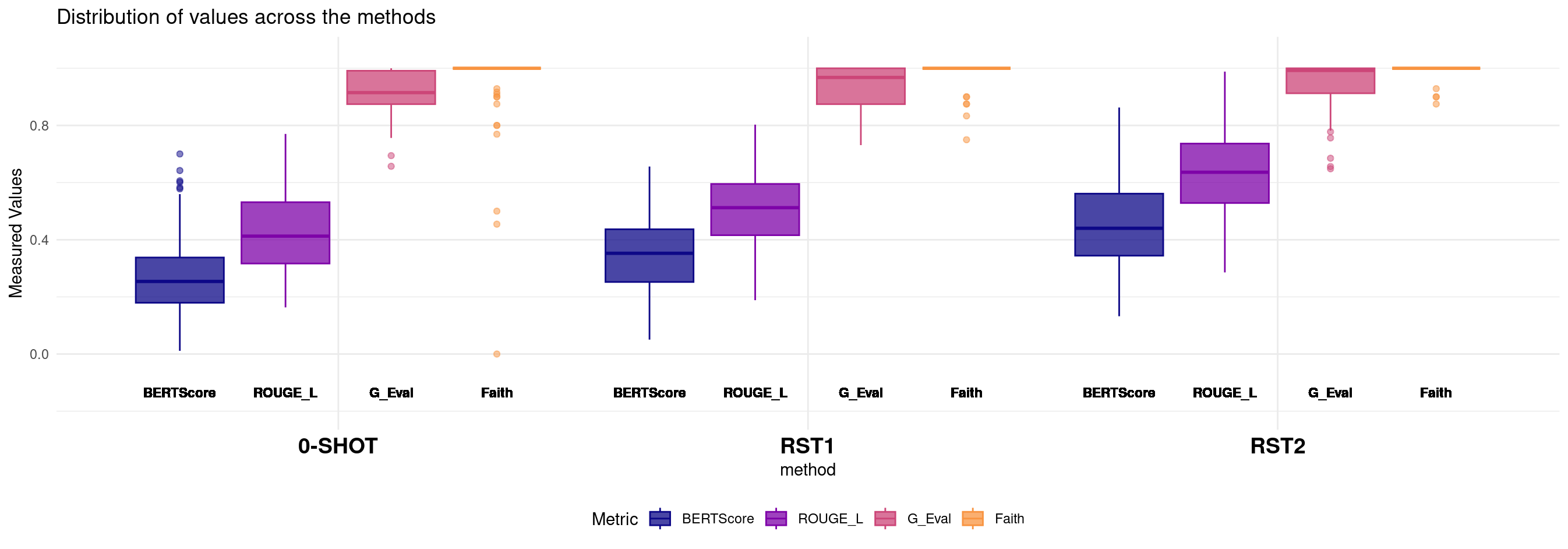}
     \caption{Distributions of BERTScore, ROUGE-l, G-Eval, Faithfulness Metric per each method.}
     \label{fig:box}
\end{figure}

 Figures and \ref{fig:distributions} and \ref{fig:box} show the distribution of the evaluated metrics for G-Eval, BERTScore F1, ROUGE-L F1, and Faithfulness metric. Across all the metrics, 0-shot has produces the lowest results in terms of the LLM-as-a-judge faithfulness and text similarity. The distributions of the BERTScore and ROUGE-L somewhat overlap, whereas the G-Eval is asymmetrical with no observations below $0.6$.The Faithfulness scores results are less certain as GPT-5.4 assigned most of the diagrams the perfect score for the proportion of truthful context claims, which made any fine-grained analysis impossible. Figure \ref{fig:distributions} contains correlation coefficients between the metrics: BERTScore and ROUGE-L are highly correlated with each other and are individually moderately correlated with G-Eval ($p< 0.001$), wheres the Faithfulness metric is weakly correlated with the other metrics.

\begin{table*}[ht!]
\caption{Mean score results across the model configurations} 
\centering
\begin{tabular}{lllrlll}
  \toprule
model & reasoning & method & G-Eval&Faithfulness&BERTS &ROUGE-L\\ 
  \midrule
GPT-4.1 & - & 0-SHOT & 0.89 &1.00  &0.25  &0.37 \\ 
  GPT-4.1 & - & RST1 & 0.94 &0.98  &0.30  &0.46 \\ 
  GPT-4.1 & - & RST2 & \textbf{0.96} &1.00  &\textbf{0.46 } &\textbf{0.65 }\\ \hline
  GPT-4.1-mini & - & 0-SHOT & 0.89 &0.97  &0.24  &0.40 \\ 
  GPT-4.1-mini & - & RST1 & 0.93 &0.99  &0.35  &0.51 \\ 
  GPT-4.1-mini & - & RST2 & \textbf{0.96 }&1.00  &\textbf{0.54}  &\textbf{0.75} \\ \hline
  GPT-5.6-luna & high & 0-SHOT & 0.93 &0.99  &0.29  &0.46 \\ 
  GPT-5.6-luna & high & RST1 & 0.95 &1.00  &0.38  &0.55 \\ 
  GPT-5.6-luna & high & RST2 & \textbf{0.96} &0.99  &\textbf{0.54 } &\textbf{0.72} \\ \hline
  GPT-5.6-luna & none & 0-SHOT & 0.93 &0.98  &0.26  &0.44 \\ 
  GPT-5.6-luna & none & RST1 & 0.94 &1.00  &0.35  &0.53 \\ 
  GPT-5.6-luna & none & RST2 & \textbf{0.95} &1.00  &\textbf{0.53 } &\textbf{0.73 }\\ \hline
  GPT-5.6-sol & high & 0-SHOT & 0.93 &0.98  &0.31  &0.46 \\ 
  GPT-5.6-sol & high & RST1 & 0.93 &1.00  &0.36  &0.50 \\ 
  GPT-5.6-sol & high & RST2 & \textbf{0.96} &1.00  &\textbf{0.40  }&\textbf{0.54 }\\ \hline
  GPT-5.6-sol & none & 0-SHOT & 0.92 &0.96  &0.25  &0.42 \\ 
  GPT-5.6-sol & none & RST1 & 0.92 &1.00  &0.34  &0.48 \\ 
  GPT-5.6-sol & none & RST2 & \textbf{0.95} &1.00  &\textbf{0.38 } &\textbf{0.53} \\ \hline
  GPT-5.6-terra & high & 0-SHOT & 0.92 &1.00  &0.28  &0.44 \\ 
  GPT-5.6-terra & high & RST1 & 0.93 &1.00  &0.34  &0.52 \\ 
  GPT-5.6-terra & high & RST2 & \textbf{0.94} &1.00  &\textbf{0.40 } &\textbf{0.57 }\\ \hline
  GPT-5.6-terra & none & 0-SHOT & 0.94 &1.00  &0.27  &0.45 \\ 
  GPT-5.6-terra & none & RST1 & 0.94 &1.00  &0.35  &0.52 \\ 
  GPT-5.6-terra & none & RST2 & \textbf{0.95} &1.00  &\textbf{0.41 } &\textbf{0.56} \\ 
   \bottomrule
\end{tabular}
\label{desc_mean}
\end{table*}

Table \ref{desc_mean} shows the mean values across the model configurations and methods, where RST1 and RST2 display higher mean values for both LLM-as-a-Judge G-Eval and Faithfulness and BERTScore and ROUGE-L. The RST2 diagrams largely have higher level of similarity and faithfulness proportions to their source contexts than both the 0-shot and RST1. The results also show that the ICL diagrams do not appear to benefit from larger models and higher reasoning settings in terms of faithfulness and similarity.

\begin{figure}[h!]
     \centering
     \includegraphics[width=1\linewidth]{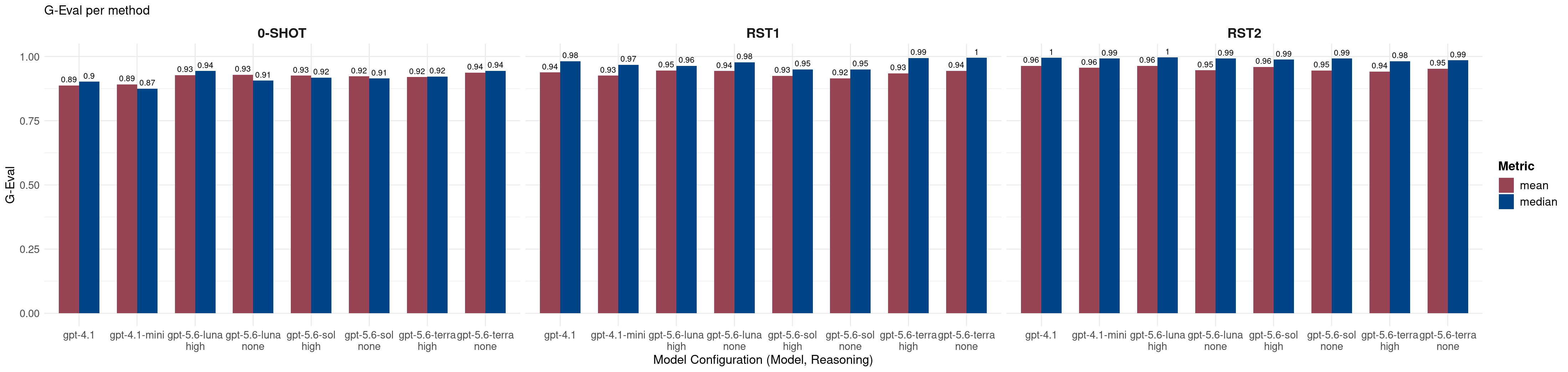}
     \caption{Mean and median of G-Eval scores across the model configurations per each method.}
     \label{fig:eval}
\end{figure}
\begin{figure}[h!]
     \centering
     \includegraphics[width=1\linewidth]{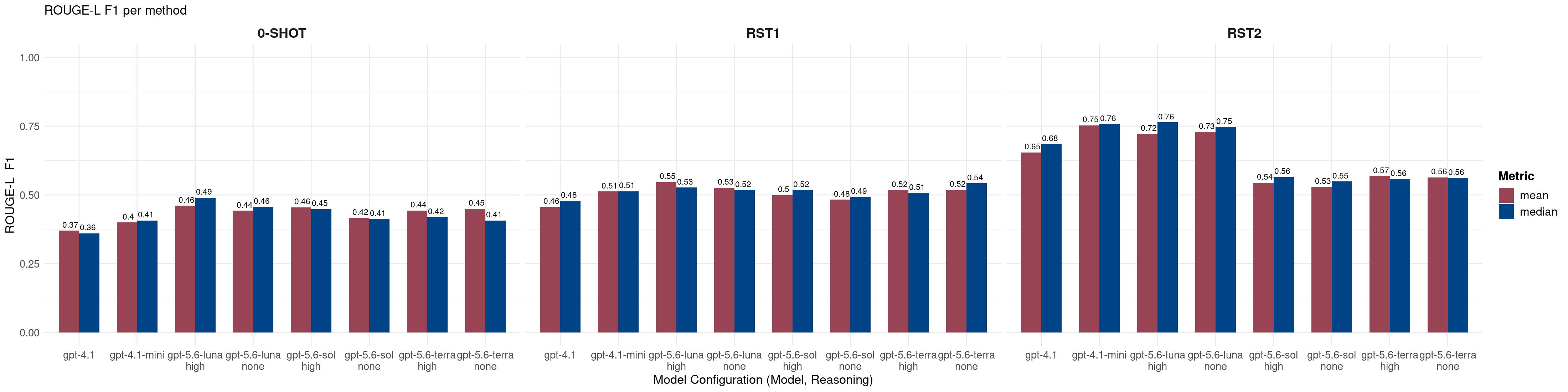}
     \caption{Mean and median of ROUGE-L F1 scores across the model configurations per each method.}
     \label{fig:rougel}
\end{figure}
\begin{figure}[h!]
     \centering
     \includegraphics[width=1\linewidth]{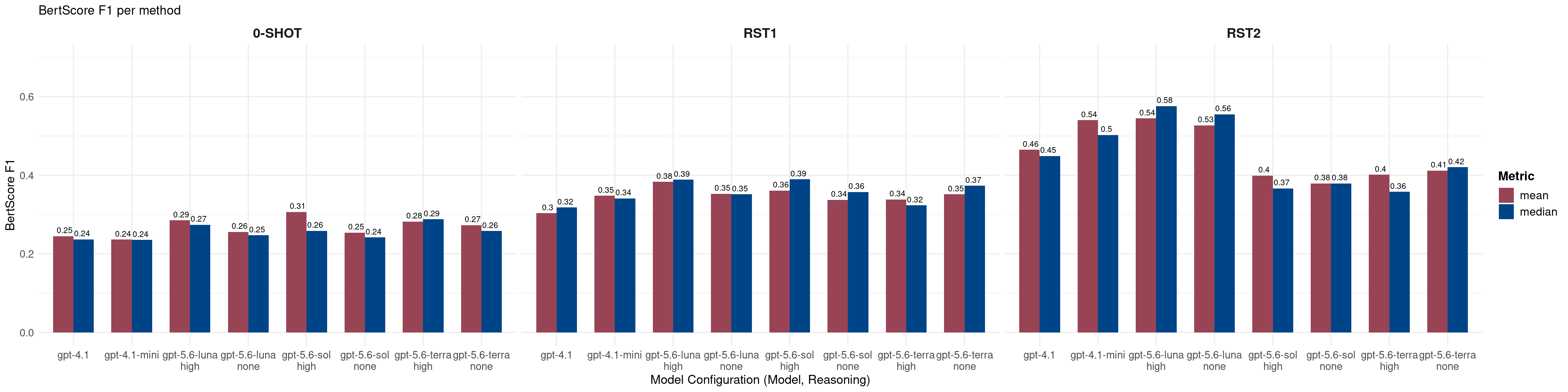}
     \caption{Mean and median of BERTScore (F1) across the model configurations per each method.}
     \label{fig:bert}
\end{figure}
Figure \ref{fig:eval} displays the mean and medium G-Eval across all the model configurations for each method (for detailed results see Appendix H). ICL-based methods produce higher mean and median G-Eval scores across all the models. For the 0-shot generation, the GPT-4.1 models produced least faithful diagrams, with GPT-4.1-mini sample mean = $0.89$ and median = $0.87$, while RST2 with GPT-4.1-mini mean = $0.96$ and median = $0.99$. The results show when the GPT-5.6 models are configured with no reasoning effort, their performance for G-Eval scores remains largely similar.

Figures \ref{fig:rougel} and \ref{fig:bert} show overall low ROUGE-L and BERTScore mean and median scores for the 0-shot diagrams; Table \ref{desc_mean} contains the mean scores for the configurations, where RST2 produces considerably higher BERTScore and ROUGE-L F1 averages, especially for the smaller tested models, GPT-4.1-mini and GPT-5.6 Luna: the highest ROUGE-L scores of $0.72-0.73$ with RST2 compared to $0.40-0.46$ with 0-shot, and BERTScore of $\approx0.54$ for RST2 and $0.40-0.46$ for 0-shot. For RST2, the smaller models produce diagrams with much higher text similarity to their contexts, where GPT-5.6 Terra and Sol result in BERTScore $\approx0.4$ and ROUGE-L $0.53-0.57$. 

The results of the automated assessment of the diagram quality $G$ defined in \ref{g_outcome} have been unreliable as GPT-5.6 Sol overinflated assessment scores, which resulted in its low performance for identifying diagrams of low quality (F1-score $= 0.53$, see the full performance statistics in Appendix H). \citet{mondal-etal-2024-scidoc2diagrammer} also report models assigning higher scores in layout assessment than human annotators (up to a $1$ point discrepancy on a 5-scale evaluation) but do not provide any additional information on reliability of their automated layout evaluation, which does not allow us to compare performance specifically for flagging low quality diagrams. 

We assessed the reliability of GPT-5.6 Luna performance for reconstructing source texts from the given RST analyses, which we used for G-Eval, BERTScore, and ROUGE-L assessment for the diagrams generated with the RST2 approach. The model was instructed to extract EDUs and organize them according to the relational structure in the given analysis. We assessed the reliability with BERTScore (full details in Appendix H), with the average F1 $=0.94$, by comparing the reconstructed texts to the original source texts that were analyzed. 
\subsection{Effect Sizes} 

\begin{figure}[ht!]
     \centering
     \includegraphics[width=1\linewidth]{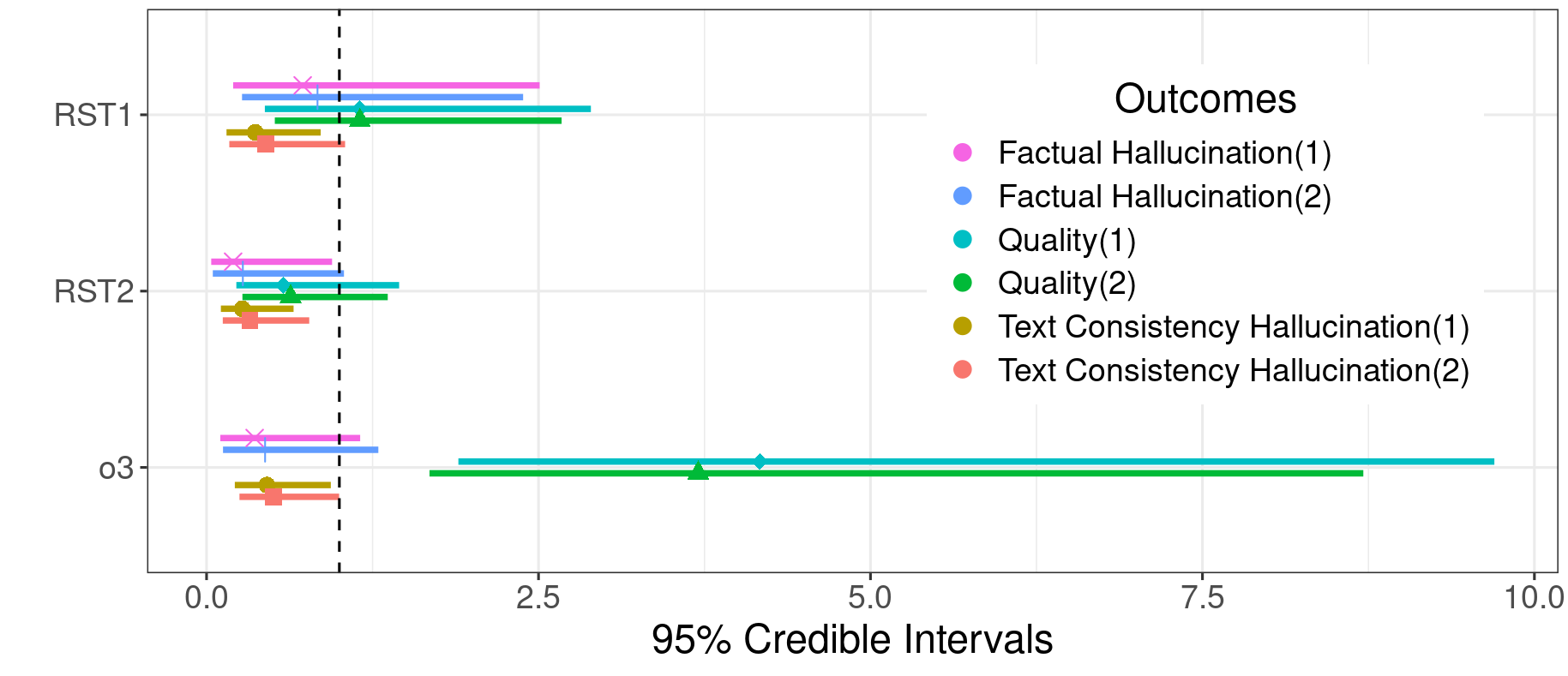}
     \caption{Results of Bayesian Data Analysis: Odds Ratios for Hallucination and Quality Outcomes. $(1)$ stands for posterior estimates obtained with $Pr_1a$, $(2)$ for $Pr_2a$.}
     \label{fig:bayes}
\end{figure}

\begin{figure}[ht!]
     \centering
\includegraphics[width=1\linewidth]{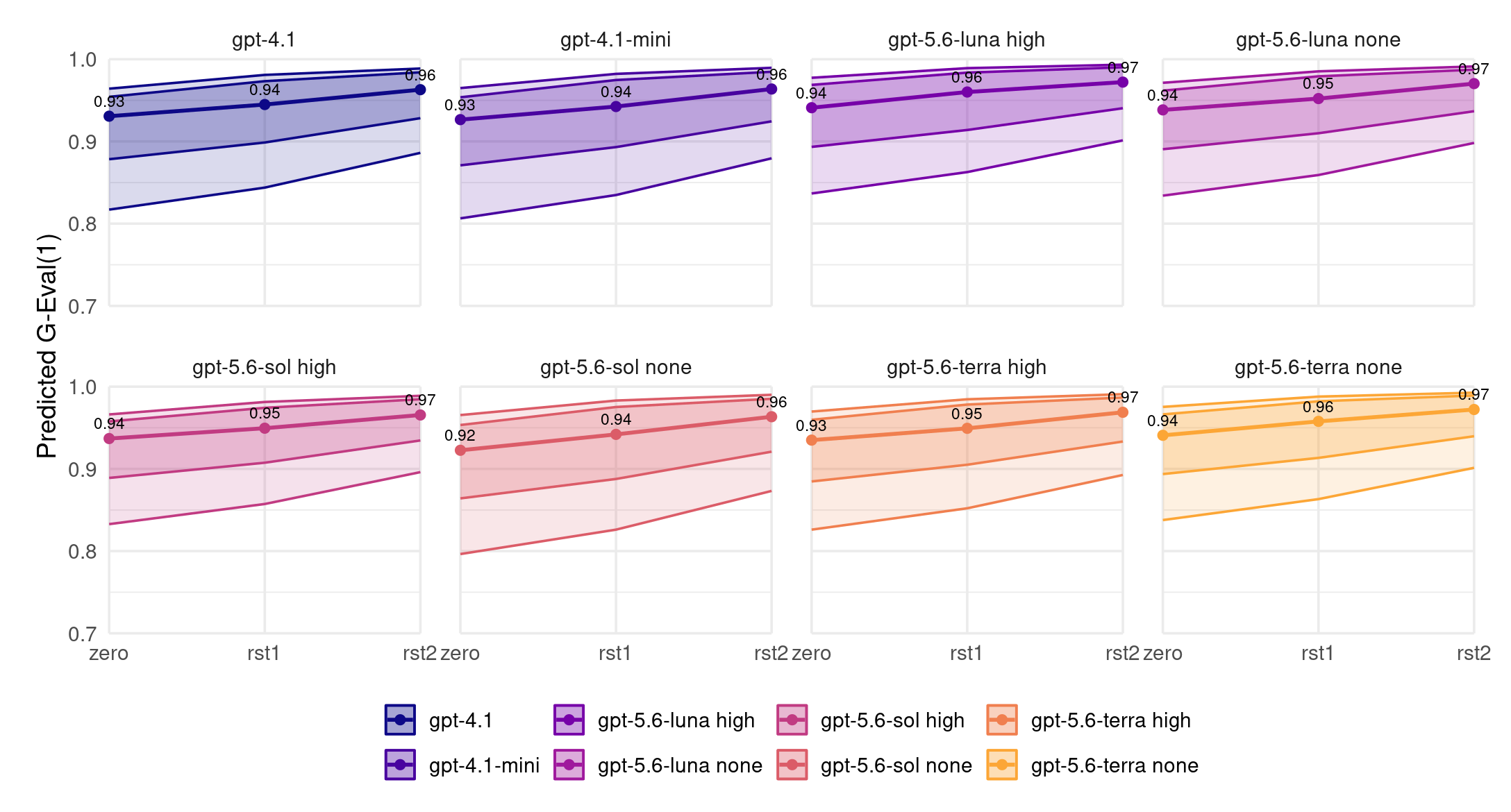}
     \caption{Predicted G-Eval scores for the model configurations and the methods with $Pr_{1b}$.}
    \label{fig:bayes_aut1}
\end{figure}
\begin{figure}[ht!]
     \centering
\includegraphics[width=1\linewidth]{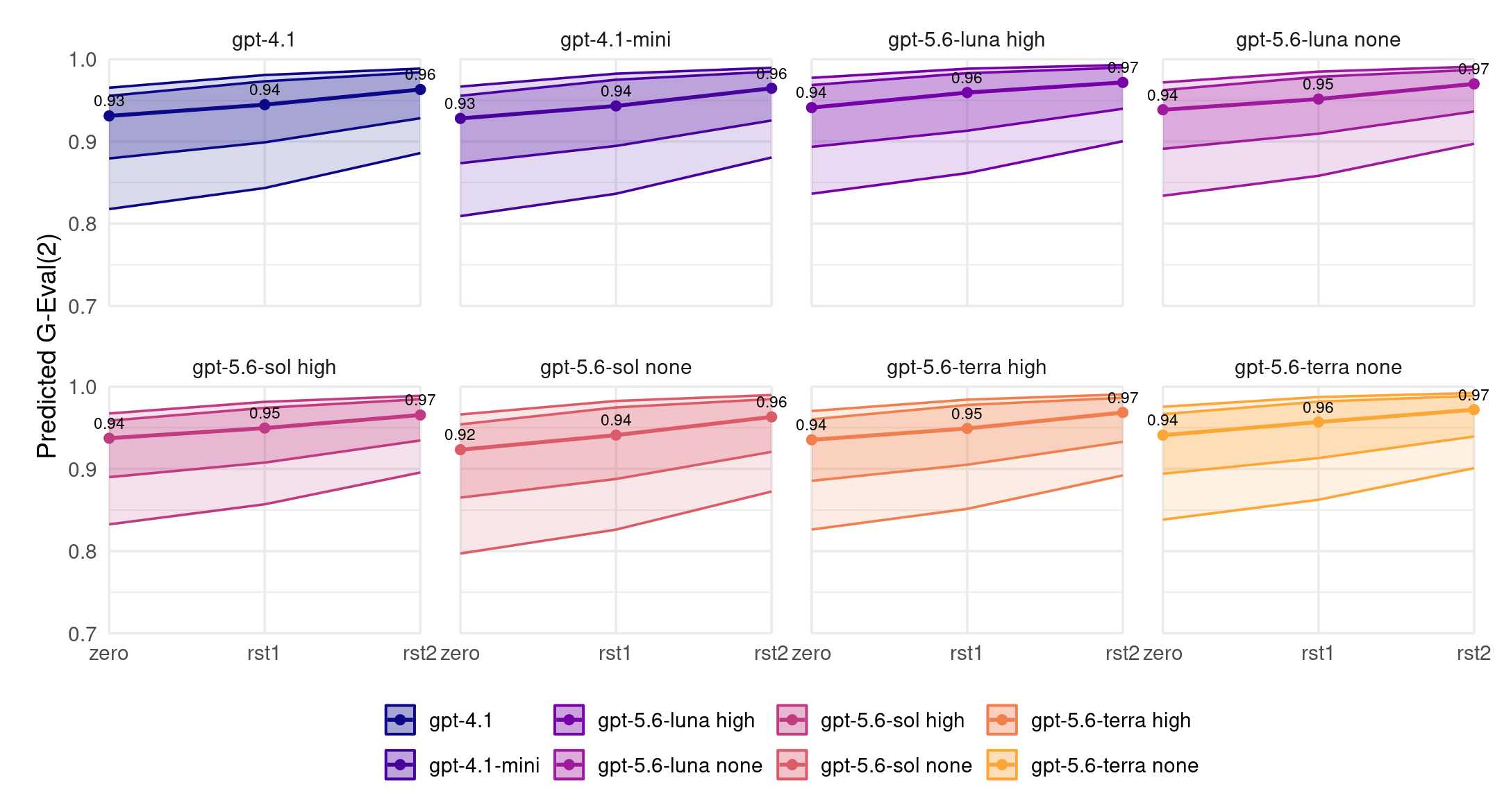}
     \caption{Predicted G-Eval scores for the model configurations and the methods with $Pr_{2b}$.}
    \label{fig:bayes_aut2}
\end{figure}

Fig. \ref{fig:bayes} displays the effect sizes of the methods and models on the $F_{fact}, F_c$ hallucination and quality outcomes in the human evaluation. We report the full results of the posterior estimates in Appendix D. Prior sensitivity comparison shows that the $G$ estimates remain mostly unchanged with the priors (\ref{priors1},\ref{priors2}); however, the credible intervals for the quality outcome indicate high uncertainty about the magnitude of the effects. There is no strong evidence of difference between the RST1, RST2, and the baseline for the quality outcome $G$, whereas with o3 the odds of obtaining a good quality diagram are higher. The data in Tables \ref{tab:humaneval}, however, show that the quality for RST2 varies based on the model choice, so there is a possible interaction that we did not account for in the regression fit due to the sample size being too small for a larger number of parameters. 

There is significant evidence text faithfulness hallucination $F_c$ is decreased with RST2 OR 0.27 (0.11, 0.66) $Pr_{1a}$; OR 0.33 (0.12, 0.77) $Pr_{2a}$), but the credible intervals for o3 and RST1 are uncertain based on the prior sensitivity analysis with the $Cauchy(0,1)$ prior for coefficients restraining smaller effects. Similarly, the effects on factual hallucination $F_{fact}$, although appear to be reduced with o3 and the ICL, are not conclusively supported by the posterior credible intervals crossing the threshold of 0. 

RST2 appears to have a measurable effect both on the value of G-Eval $(0,1)$ and whether a score lands exactly on a 1. Since our observations lacked any data points below $0.65$, the $\gamma$ component in Eq. \ref{zoib} was modeled as $\sim1$ setting conditional one-inflation (`coi') probability to be a constant. In the analysis, 0-shot and GPT-4.1 were set as the baselines. Neither of the 7 model configuration (GPT-4.1-mini and GPT-5.6 Luna, Terra, Sol with both no and high reasoning) show any significant effect on the G-Eval faithfulness scores compared to GPT-4.1.  The results of the prior sensitivity indicate the credible intervals of the posterior estimates largely overlap (see Appendix E). For $\beta_{rst2}^{\mu}$, the estimate is $0.53 (0.36, 0.71)$ for $Pr_{1b}$ and $0.52(0.35, 0.70)$ for $Pr_{2b}$ respectively; the $\beta_{rst2}^{\alpha}$ estimates are $1.61 (1.02, 2.20)$ for $Pr_{1b}$ and $1.51 (0.91, 2.13)$ for $Pr_{2b}$. For RST1, the $\beta_{rst1}^{\mu}$ lower interval is close to $0$; however, the $\beta_{rst1}^{\alpha}$ estimate is $1.13 (0.55, 1.73)$ for $Pr_{1b}$ and $1.03 (0.46, 1.63)$ for $Pr_{2b}$. RST2 produces OR $4.99 (2.77, 9.06)$ with $Pr_{1b}$ and OR $4.54 (2.49, 8.43)$ with $Pr_{2b}$ for zero-one inflation; RST1 results in OR $3.10 (1.73, 5.65)$ with $Pr_{1b}$ and OR $2.81 (1.58, 5.13)$ with $Pr_{2b}$ respectively. Figures \ref{fig:bayes_aut1} and \ref{fig:bayes_aut2} display the predicted median G-Eval scores; RST2 outperforms both RST1 and 0-shot in terms of the predicted G-Eval median outcomes by 3-4\% increase depending on a model configuration. 
\section{Discussion}
We investigated how in-context examples affect various aspects of diagram code generation. We compared two variants of ICL formats: RST1 requiring a model to perform a conventional task of diagram code generation from an ordinary text and RST2 asking it to generate diagram code from a discourse analysis. Our results suggest that diagram code generation with RST1 and RST2 functions differently as the measured quality outcomes are model-dependent. As earlier research showed that LLMs fail to perform tasks that fall outside of their pretraining task types and a model's ability to generalize on OOD tasks depends on their pretraining task diversity \citep{wang2025can,pmlr-v267-goddard25a}, our results suggest that o3's ability to perform the discourse-to-diagram mapping --- while GPT-4o largely failed on it --- is possibly due to the task itself being outside the pretraining types. As ICL performance relies on overcoming the distributional difference between a given prompt itself and a model's pretraining prior by in-context examples allowing the model to infer latent concepts from the prompt~\cite{xie2022an}, RST1, having produced a higher proportion of good diagrams with both models, appears to provide a sufficiently strong signal for performing the task.

Our results of the automated G-Eval faithfulness evaluation with 8 model configurations (600 diagrams) supports our human evaluation findings. RST2 increases the predicted median G-Eval score by 3-4\% and also shows a $\approx$ 4.5-5 fold improvement in the odds of the G-Eval hitting exactly 1 in comparison to 0-shot generation, whereas RST1 offers $\approx$ 3 fold increase in the odds. Our results suggest that there is no evidence of a significant difference between the tested models and their reasoning effects in terms of the G-Eval faithfulness scores. Additionally, we found the DeepEval Faithfulness assessment uncertain as GPT-5.4 assigned the perfect score to almost all the 600 diagrams. The obtained G-Eval scores are moderately correlated with the ROUGE-L and BERTScore numbers, while the Faithfulness scores are weakly correlated, which is likely due to a different underlying scoring mechanism.

We also attempted LLM-as-a-judge assessment of diagram logical organization, connectivity, and layout with human-written criteria performed by GPT-5.6 Sol with high reasoning, which resulted in unreliable results with the model not being able to reliably detect diagrams of poor quality.~\citet{mondal-etal-2024-scidoc2diagrammer} reported LLMs diverging from human assessment by up to 1 point on their 5-point scale, which is in line with our findings. Our human evaluation also showed that neither o3 or GPT-4o were able to consistently refine diagram layout.

The ROUGE-L and BERTScore scores show that RST2 largely produces diagrams with tangibly more similar texts than both RST1 and 0-shot. We observe that the highest mean ROUGE-L F1 scores are achieved with the smaller tested models, GPT-4.1-mini and GPT-5.6 Luna (both no and high reasoning), meaning that the smaller models appear to reproduce given contexts in generated DOT when given a novel ICL format. 

Our novel finding lies in RST2 improving context fidelity, suggesting that the task format within the diagram generation prompt affects LLMs' output. This effect is also present for RST1, but its magnitude is smaller and its significance is uncertain.  0-shot generation reflects the diagram generation with the posterior being dominated by the prior pretraining, as it lacks any specific signal present in ICL. While it produces visibly coherent diagrams, they suffer from higher rates of infidelity to provided contexts. Overall, based on our findings, there is no evidence that the ICL affects the superficial quality of diagram presentation in terms of layout and logical structure; however, it significantly improves faithfulness to input context by reducing the odds of context faithfulness hallucination by nearly $70\%$. \citet{kang-etal-2025-unfamiliar} suggest manipulating AI hallucination through fine-tuning with unfamiliar examples. Our findings indicate that a similar approach could be beneficial for ICL for diagram generation, with RST2 significantly improving faithfulness measured with both G-Eval and text inconsistency hallucination occurrences.
\section{Conclusion}
Our study proposed a new framework for in-context examples for diagram code generation based on discourse structure and evaluated how the inclusion of the ICL demonstrations of two different formats affects various aspects of diagram quality, namely logical organization, connectivity, aesthetics, and hallucination occurrence. Our findings reveal that zero-shot instruction-based generation was more likely to produce superficially fluent but unfaithful outputs. We found that the ICL affects diagram context faithfulness, with the novel format task significantly improving fidelity. In future studies, our empirical experiments should be expanded to other context-grounded structured output task research to investigate if our findings are consistent with other tasks.
\section*{Acknowledgments}
This research was supported by the Research Council of Finland (Academy Research Fellow grant number 356114). 

We would like to express our gratitude to Prof. Tuomo Hiippala for providing us with early feedback on our work and relevant background literature, Dr. Artturi Tilanterä for their feedback on the generated diagrams and the student survey, and all the authors of the materials we used in our work.

\appendix
\section{Prompts}\label{prompts}
$R_1$ is adapted from \citet{stede2017annotation}

You are a linguist analyzing text according to the Rhetorical Structure Theory. The theory states that a text can be represented as a tree where each leaf is an elementary discourse unit (EDU) and nodes display relations between EDUs. Relations can be formed between EDUs or larger spans recursively. Primary units, which contain the core information of a relation, are nuclei, whereas secondary units are called satellites. In the RST annotation, EDUs are the smallest units of discourse and RST trees have EDUs as leaves. In the following paragraphs, you will be given a step by step guide on how to perform an analysis.

Step 1. SEGMENTATION
Segment text. Generally, discourse segments are clauses and sentences. All discourse segments to contain a verb. Whenever a discourse boundary is inserted, the two newly created segments must each contain a verb. You need to segment coordinated clauses and coordinated verbal phrases, adjunct clauses with either finite or non-finite verbs, and non-restrictive relative clauses (marked by commas, parentheses or other typographical features). Here are some extra guidelines on EDUs:
1.1 COMPLEMENT CLAUSES:
[I wouldn't be far off] [if I said this is one of his greatest performances.]
Complement clauses do not constitute EDUs. These include subject and object clauses, and some objects of nouns and other parts of speech. In this example, there are 2 EDUs, the first one is the main one, and the second one is a subordinate conditional clause. The direct object clause, a complement of the verb, is not an EDU. 
1.2 RELATIVE CLAUSES 
The best candidates for discourse status are non-restrictive relative clauses, i.e., those set off from the rest of the clause by commas, dashes or similar typography. 
1.3 ATTRIBUTION 
Do not segment complement clauses of reporting verbs. Ignore direct speech and quotes, and include any material in quotes as part of the main clause.
1.4 PREPOSITIONAL ADJUNCTS 
Some prepositional adjuncts are good candidates for discourse status, as they are very close to clausal adjuncts. However, an EDU should contain a verb, which is why a prepositional phrase without one can't be considered an EDU. For example, in  Both students and faculty pay the same amount for childcare, regardless of income.  "regardless of income" isn't an EDU; but in It is hard to see past his megastar status, [regardless of how good the performance is.] "regardless of how good the performance is" is an EDU.

Step 2.
Decide on the hierarchical structure of the text: Which adjacent units are to be connected to each other in what order and what is the resulting tree structure that covers the complete text?

Step 3.
When joining two adjacent units into a larger one, decide on the relation to be applied. Decide whether one of the EDUs is more important than the other, or whether both are of equal weight. In case you choose a multinuclear relation, more than two EDUs might belong together. Relations are being defined by explaining (a) the role of the two units that are being adjoined and (b) the effect that the author wants to achieve by applying the relation to the units. The following format is used in the definitions: N: nucleus, S: satellite, N/S: the function of the nucleus/satellite combination, R: reader, W: writer. Here are the definitions of various relations that you need to use in your analysis: 
3.1 PRIMARY PRAGMATIC RELATIONS 
Background: N/S: Understanding S makes it easier for R to understand the content of N; without the background information in S, it would be difficult to comprehend N. In a text, S mostly but not always precedes N. A Background S at the beginng of the text often serves to introduce the topic of the text.
Antithesis: N: W regards the content of N as more important; it is the antithesis that W is identifying with. S: In comparison to N, W regards the content of S as less important. S is considered to be the thesis which the W is not identifying with.
Concession: N/S: W concedes S and implicitly confirms that S and N are usually not compatible; in the current instance, however, they are compatible, and N is being emphasized.
Evidence: N/S: Understanding S makes it easier for R to accept N, or to share the particular viewpoint of W. 
Reason:  N/S: Understanding S makes it easier for R to accept N, or to share the particular viewpoint of W. Reason is more specific than Evidence. The different lies on whether S is being presented by W as "objective" (Evidence) or also constitutes a subjective statement itself (Reason).
Justify: N/S: Understanding S makes it easier for R to accept N, or to share the particular viewpoint of W.
Evaluation:  N/S: S evaluates N or N evaluates S.
Motivation: N/S: S presents a reason for performing the action described in N.
Enablement:  N/S: Comprehending S makes it easier for R to perform the action described in N.
3.2 PRIMARY SEMANTIC RELATIONS 
Circumstance: N/S: S characterizes a framework in which N is to be interpreted, such as its temporal or locative position. Typical connectives: as; when; while; ... for a temporal frame.
Condition: N/S: The realization of N depends on the realization of S. Typical connectives: if .. then; in case; ...
Otherwise:  N/S: The realization of N impedes the realization of S. Typical connectives: otherwise; ...
Unless:  N/S: S determines the realization of N: N is only being realized if S is not being realized.Typical connectives: unless; ...
Elaboration:  N/S: S provides details or more information on the state of affairs described in N (but not on a single entity mentioned in N; see E-Elaboration below). N precedes S in the text. Typical relations between N and S are set::element, whole::part, abstraction::instance, procedure::step.  Typical connectives: in particular; for example; ...
E-Elaboration: N/S: S provides details or more information on a single entity mentioned in N. N precedes S in the text.
Interpretation: N/S: S shifts the content of N to a different conceptual frame. This does not imply an evaluation of N (or the evaluation is of only secondary importance). N precedes S in the text. Typical connectives: thus; ...
Means:  N/S: S provides information that makes the realization/execution of N more probable or simple (e.g., an instrument). Typical connectives: thus; ... Example: [In August, Berliners always enjoy travelling to Lichtenrade.]N [To that end, they usually take the S25 train.]S
Cause:  N/S: The state/event in N is being caused by the state/event in S. Typical connectives: because; since; therefore; ...
Result: N/S: The state/event in S is being caused by the state/event in N. Typical connectives: because; since; therefore; ... This relation is parallel to Cause. Deciding between the two depends solely on judging the relative importance of the segments for the text.
Purpose: N/S: S is being realized through the realization/execution of N. Typical connectives: in order to; to; ...  There is a causal relationship in a wide sense. The difference to the relations Cause/Result is that with Purpose, S is clearly marked as hypothetical/unrealized, and represents the intention or goal of the acting person.
Solutionhood:  N/S: The content of N can be regarded as a solution to the problem in S. N usually precedes S in the text. Example: [With the anti-smoker regulations being passed, many pubs will be caught in a trap.]S [They should start looking into possibilities for having separate rooms.]N
3.3 TEXTUAL RELATIONS 
Preparation:  N/S: S precedes N in the text. S orients R toward the topic of N. This relation is to be used when S does not serve any stronger purpose than setting the topic for N, or when it consists of an introductory formula. S should contain only minimal information on its own. 
Restatement:  N/S: N precedes S in the text. S repeats the information given in N, using a different wording. N and S are of roughly equal size.  Typical connectives: in other words; ...   Example: [The mayor gave all the information to the councillors,]N [kind of filling them in completely.]S  
Summary:  N/S: S succeeds N in the text and repeats the information given in N, but in a shorter form.
3.4 MULTINUCLEAR RELATIONS    
Contrast: N: Exactly two nuclei. Both are of equal importance for W's purposes.The contents are comparable yet not identical. They differ in aspects that  are important to W.  Typical connectives: on the other hand; yet; but; ... Example: [My first car was small.]N [The second was already a sizable limousine.]N
Sequence:  N: The nuclei describe states of affairs that occur in a particular temporal order. Typical connectives: then; before; afterwards; ... The states of affairs can be presented in their actual temporal order ("afterwards") or in the opposite one ("before that").
List: N: The nuclei provide information that can be recognized as related, enumerating. They all contribute to the text function in the same way. Example: What I did yesterday: [Cook dinner,]N [look after the kids,]N [clean the bathromm.]N 
Conjunction:  N: The nuclei provide information that can be recognized as related, enumerating. They all contribute to the text function in the same way, and they are linked by coordinating conjunctions.  Typical connectives: and; or; ... The functions of List and Conjunction are identical. When the surface condition for Conjunction holds, this relation is to be used.
Joint: N: The nuclei provide different kinds of information, which are not of the same type; yet they are not in a clearly identifiable semantic or pragmatic relation, nor do they form an enumeration. Still, there is a coherent link, as they contribute in similar ways to the overall text function. Typical connectives: Additive connectives such as in addition; also. Joint is to be used when a multinuclear relation is needed (from the text-global perspective) but none of the specific relations are applicable.
When all pairs of neighbouring EDUs have been checked, continue by considering the larger units. A connective can join longer units than a single EDU, and relations between EDUs and/or larger units can also be unsignalled. In marking the relations between larger segments, it is advisable to proceed in a bottom-up fashion: Conjoin EDUs and/or neighbouring larger segments, and successively construct the tree moving upward.

Step 4. 
Decide on the nucleus/satellite status of the linked units.   
The last three steps are not performed separately but are closely tied to one another. At the end of the annotation process, the complete text has to be covered by the tree structure, without any gaps (EDUs that are not participating in the analysis). At any point, adjacent units are being related to one another such that no crossing edges originate in the tree. (In other words, the tree is "projective".)                  
Another important property of the tree is that no node has more than one parent node, which means that any unit of the text can play only one role in the rhetorical structure. It thus cannot function as a satellite in two distinct relations, being linked to different nuclei. On the other hand, it is possible that a single nucleus has multiple satellites. The final sentence amounts to the central statement of the text and therefore constitutes the central nucleus of the text: If you start at the top (root) node of the tree and move to that leaf, along the path you encounter only nucleus links.
Be concise. Output EDUs and an RST tree as lists.

\textbf{Prompt $R_2$}

System message: "You will be given a discourse analysis of a text and asked to find a similar text based on Rhetorical Structure Theory from a set of 4 texts. Your choice should be based on the types of discourse relations present in the text."  + $A$

User message: "You will be given a piece of analyzed text as input. Your output should contain the id of the chosen text from the 4 analyses. Text analysis:" + $a_i$

\textbf{Prompt $R_{3rst1}$}

System message: "You are an educator creating Graphviz diagrams from educational materials. You are given an example diagram. Pay attention to the dot syntax. Follow general design principles for diagrams: avoid line crossings and bends; keep nodes unified in size, shape, and color; minimize width; when applicable, follow flowchart conventions. Here is the example:" + $E$

User message: "You will be given a piece of text as input. Your output should contain a piece of dot diagram code for the diagram generation. Add a disclaimer to the diagram's label stating that it is generated with an AI model. Text:" + $s_i$

\textbf{Prompt $R_{3rst2}$}

System message: "You are an educator creating Graphviz diagrams from educational materials using the Rhetorical Structure Theory analysis to aid you. You will be given an example containing an analyzed text and a diagram.
Read a given RST analysis and map its structure into individual nodes and edges, considering how discourse relations are visualized in the example. Do not visualize the RST tree itself or name the relations between the EDUs. Do not mention RST in diagrams. Follow general design principles for diagrams: avoid line crossings and bends; keep nodes unified in size, shape, and color; minimize width; when applicable, follow flowchart conventions. Here is the example: " + $E$

User message: "You will be given a piece of text as input. Your output should contain a piece of dot diagram code for the diagram generation. Add a disclaimer to the diagram's label stating that it is generated with an AI model. Text:" + $a_i$

\textbf{Prompt $R_{30}$}

System message:"You are given a text, and your task is to produce a diagram in the dot Graphviz syntax from the given text. Follow the design principles for diagrams: avoid line crossings and bends; keep nodes unified in size, shape, and color; minimize width; when applicable, follow flowchart conventions."

User message: "You will be given a piece of text as input. Your output should contain a piece of dot diagram code for the diagram generation. Add a disclaimer to the diagram's label stating that it is generated with an AI model. Text:" + $s_i$

\textbf{Prompt $R_4$}

Input Text: "Check if the given diagram has any following issues: elements obscuring each other; non-uniform size of nodes; line crossings and bends; asymmetry; excessively long text lines or edges; text overflowing boxes. If it is a flowchart, make sure it follows flowchart conventions (e.g., diamond blocks for conditionals). Ensure the label has a disclaimer stating that it is generated with an AI model. Here's the dot source of the diagram: " + $C$ +
"Output a short (less than 100 words) explanation for your improvement and your improved dot code."

\textbf{Prompt $R_5$}

System message: "You are given a piece of faulty code in the dot syntax and an error message. Correct the code by fixing the error. Do not introduce any other changes."

User message: "Your output should contain dot diagram code. Add a disclaimer to the diagram's label stating it is generated with an AI model if it is not already there. Diagram code:" + $C$ +"Error:" + $e$

\section{Pre-made Example Diagrams and Rubric}
\label{examples}

\textbf{Text 1 $o_1$} adapted from \citet{sorva}:

You might wonder what on earth 'REPL' stands for? Well, here's some short and sweet explanation: 'REPL' is short for read–evaluate–print loop. Those words convey the basic idea:

read: You can type in bits of Scala that the REPL receives as input, or "reads".

evaluate: The REPL runs your code as soon as it receives it. For instance, when given an arithmetic expression, the REPL performs the given calculation to produce a result.

print: The REPL reports the results of evaluation onscreen.

loop: This interaction between the user and the REPL keeps repeating as long as the user likes.

\textbf{Analysis 1 $a_1$}:

Disclaimer: This text is generated with the o3 language model.
The model uses the following paper to generate its output: Stede, Manfred, Maite Taboada, and Debopam Das. 'Annotation guidelines for rhetorical structure.' Manuscript. University of Potsdam and Simon Fraser University (2017).
EDUs:
1. You might wonder what on earth 'REPL' stands for?
2. Well, here's some short and sweet explanation:
3. 'REPL' is short for read-evaluate-print loop.
4. Those words convey the basic idea:
5. read: You can type in bits of Scala that the REPL receives as input, or 'reads'.
6. evaluate: The REPL runs your code
7. as soon as it receives it.
8. For instance, when given an arithmetic expression,
9. the REPL performs the given calculation to produce a result.
10. print: The REPL reports the results of evaluation onscreen.
11. loop: This interaction between the user and the REPL keeps repeating
12. as long as the user likes.

RST tree (mononuclear relations are [Relation-Name, Nucleus, Satellite];
multinuclear relations are [Relation-Name, $Nucleus_1$, $Nucleus_2$, ...]):

["Elaboration",
   ["Preparation",
      3,
      ["Preparation",
         2,
         1
      ]
   ],
   ["Elaboration",
      4,
      ["List",
         5,
         ["Elaboration",
            ["Circumstance", 6, 7],   
            ["Circumstance", 9, 8]   
         ],
         10,
         ["Circumstance", 11, 12]
      ]
   ]
]

\begin{figure}[h!]
    \centering
    \includegraphics[width=1\linewidth]{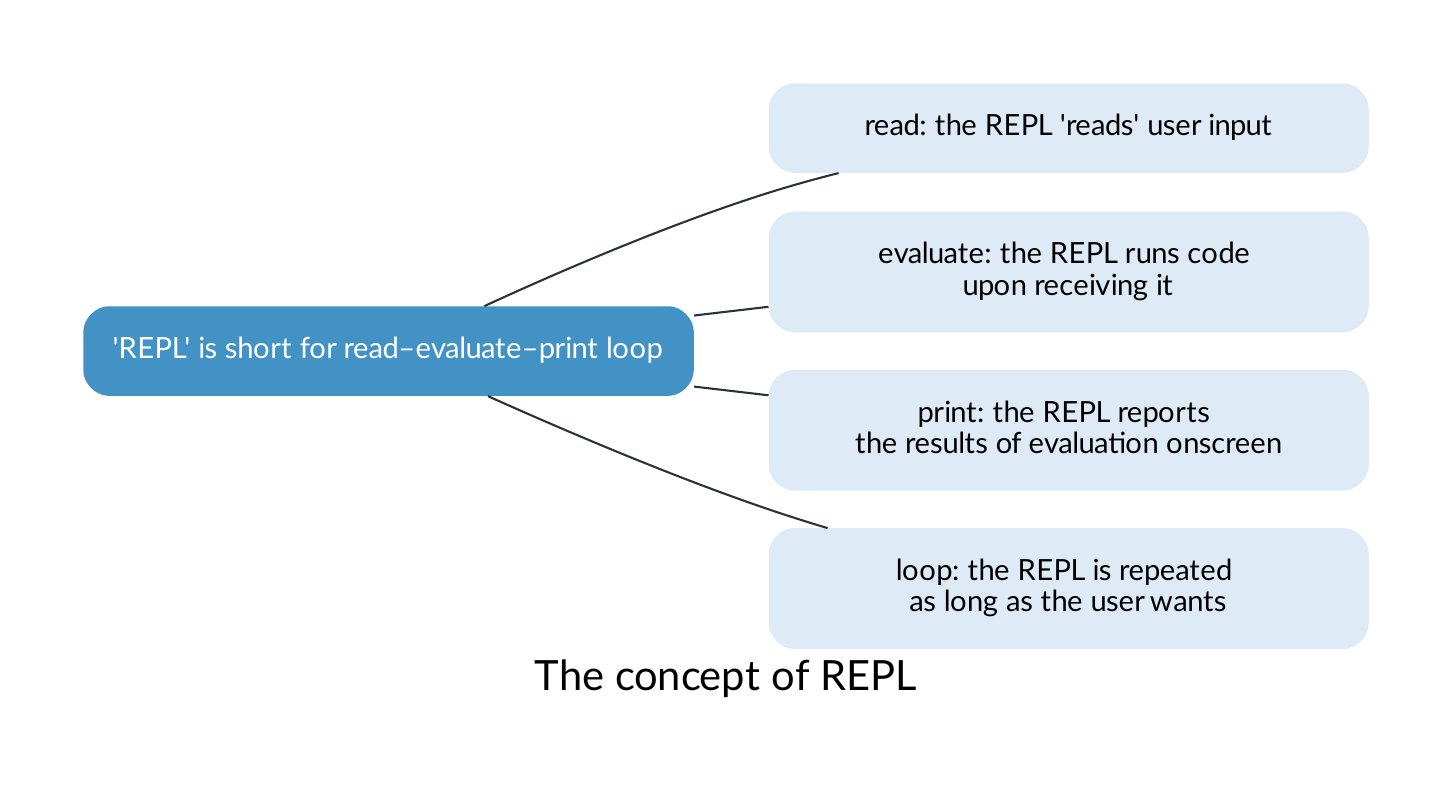}
    \caption{\textbf{Diagram 1 for Text 1}}
    \label{fig:repl}
\end{figure}

\textbf{Text 2 $o_2$} adapted from \citet{hellas}:

So how do packets travel in the TCP/IP protocol? The process is easier than it seems. First, a router receives a packet, then it checks the header of the packet for the IP-address of the recipient. Next, the router looks up the IP-address from its routing table and chooses the best match for passing the packet forward. If the router is not connected to the recipient, it passes the packet forward again. If the router is connected to the recipient, it forwards the packet to the recipient, which means that the packet has reached its destination. The packet is checked, and the recipient responds with a message to the sender, indicating that the packet has been received.

\textbf{Analysis 2 $a_2$}:

Disclaimer: This text is generated with the o3 language model.
The model uses the following paper to generate its output: Stede, Manfred, Maite Taboada, and Debopam Das. 'Annotation guidelines for rhetorical structure.' Manuscript. University of Potsdam and Simon Fraser University (2017).
EDUs:
1. So how do packets travel in the TCP/IP protocol?
2. The process is easier than it seems.
3. First, a router receives a packet,
4. then it checks the header of the packet for the IP-address of the recipient.
5. Next, the router looks up the IP-address from its routing table
6. and chooses the best match for passing the packet forward.
7. If the router is not connected to the recipient,
8. it passes the packet forward again.
9. If the router is connected to the recipient,
10. it forwards the packet to the recipient,
11. which means that the packet has reached its destination.
12. The packet is checked,
13. and the recipient responds with a message to the sender,
14. indicating that the packet has been received.

RST tree (nested lists, Relation - roles - children):
['Solutionhood', ['Satellite', 1], ['Nucleus',['Evaluation', ['Satellite', 2],['Nucleus', ['Sequence','Sequence', 3, 4], ['Sequence', 5, 6], ['Contrast',['Condition',['Satellite', 7], ['Nucleus', 8]],['Condition',['Satellite', 9],['Nucleus',['Result', ['Nucleus', 10],['Satellite', 11]]]]],['Sequence',12, ['E-Elaboration', ['Nucleus', 13],['Satellite', 14]]]]]]]]

\begin{figure}[h!]
    \centering
    \includegraphics[width=1\linewidth]{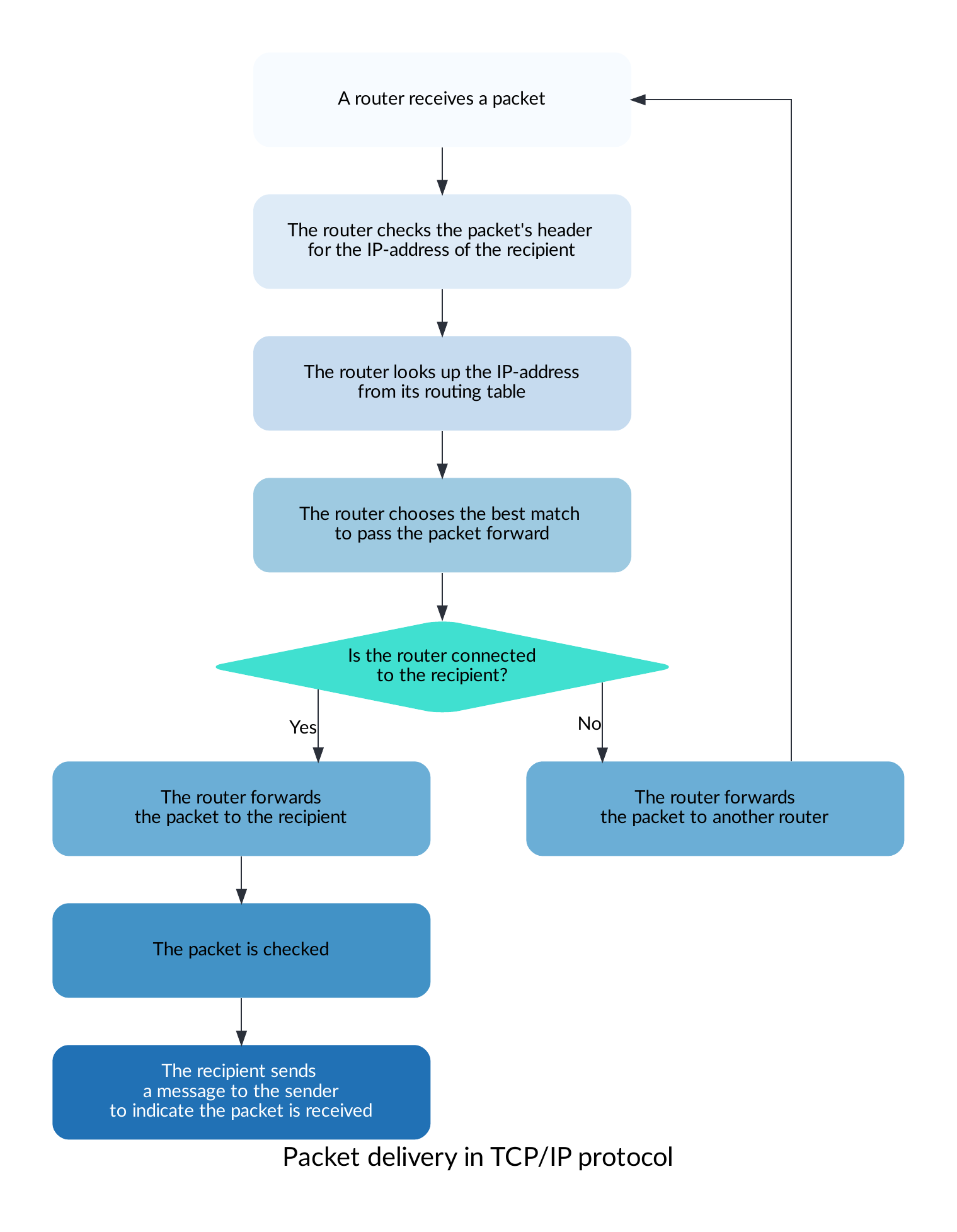}
    \caption{\textbf{Diagram 2 for Text 2}}
    \label{fig:tcpip}
\end{figure}

\textbf{Text 3 $o_3$} adapted from \citet{sorva}:

The assignment "Pick 30 fully random numbers between 0 and 9." is easier said than done. One practical option is to use a "good enough" mathematical algorithm for computing a sequence of numbers. For example:

1.Start by picking some number N, say N=20.

2.Take the Nth decimal of $\pi$. Let that be the first "random number".

3.Determine the next "random number" by taking the another digit from $\pi$ at 2*N.

4.Follow with 3*N, 4*N, and so on.

\textbf{Analysis 3 $a_3$}:

Disclaimer: This text is generated with the o3 language model.
The model uses the following paper to generate its output: Stede, Manfred, Maite Taboada, and Debopam Das. 'Annotation guidelines for rhetorical structure.' Manuscript. University of Potsdam and Simon Fraser University (2017).
Motivation
--- Satellite: EDU1
--- Nucleus
    --- Enablement
        --- Nucleus: EDU2
        --- Satellite
            --- Sequence
                --- Nucleus: EDU3
                --- Nucleus: EDU4
                --- Nucleus: EDU5
                --- Nucleus: EDU6
                --- Nucleus: EDU7

EDUs:
EDU1: The assignment "Pick 30 fully random numbers between 0 and 9." is easier said than done.
EDU2: One practical option is to use a "good enough" mathematical algorithm for computing a sequence of numbers.
EDU3: Start by picking some number N, say N=20.
EDU4: Take the Nth decimal of $\pi$.
EDU5: Let that be the first "random number".
EDU6: Determine the next "random number" by taking the another digit from $\pi$ at 2*N.
EDU7: Follow with 3*N, 4*N, and so on.

\begin{figure}[h!]
    \centering
    \includegraphics[width=1\linewidth]{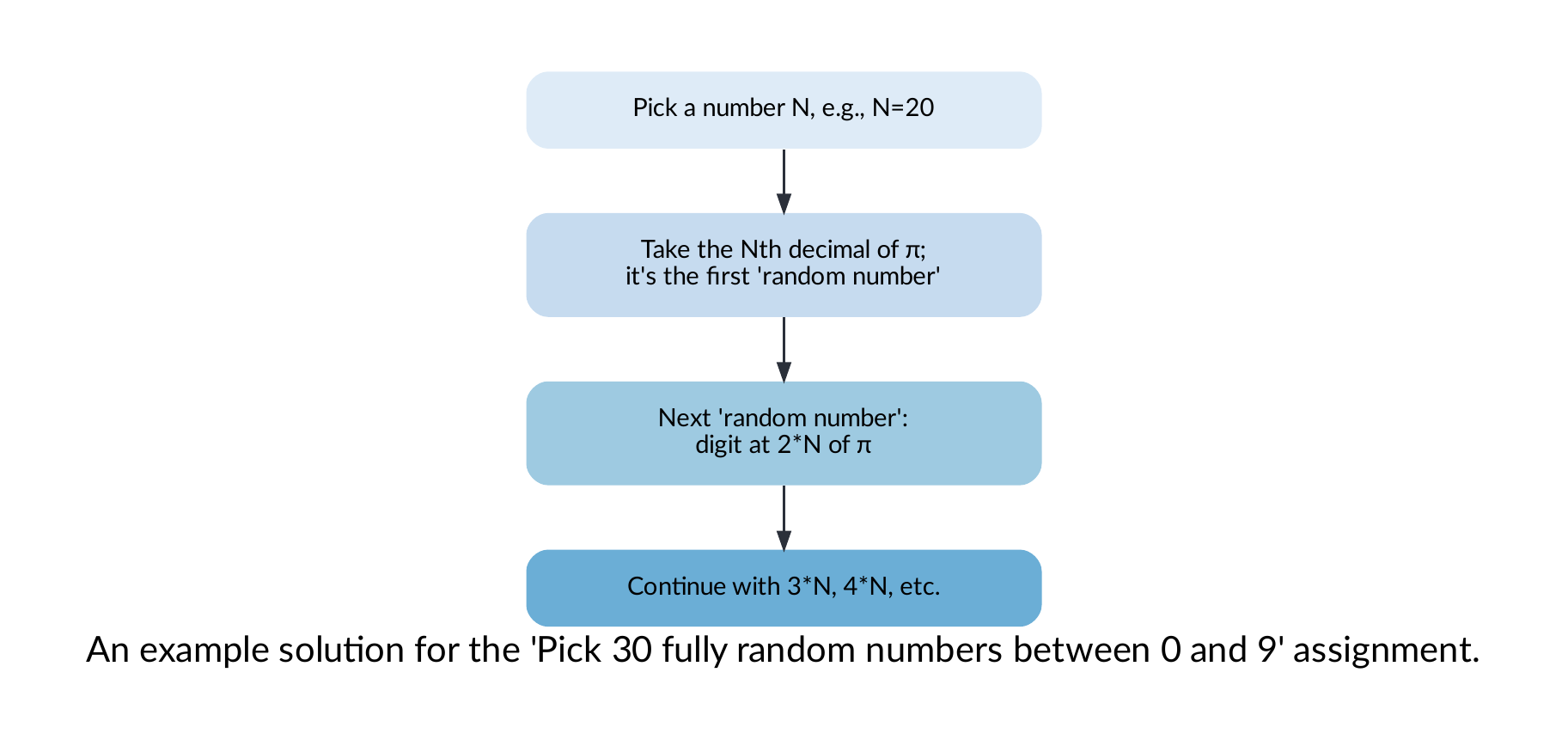}
    \caption{\textbf{Diagram 3 for Text 3}}
    \label{fig:number}
\end{figure}

\textbf{Text 4 $o_4$} adapted from \citet{suomela2025}:

There are two types variables: shared and private. Any variable that is declared outside a parallel region is shared, while any declared inside is private. In case of a shared variable, there is only one copy of it, and all threads refer to the same variable. Care is needed whenever you refer to such a variable. On the other hand, in case of any private variable, each thread has its own copy of it. Such variables are always safe to use. If a shared variable is read-only, you can safely refer to it from multiple threads inside the parallel region. However, if any thread ever writes to a shared variable, then proper coordination is needed to ensure that no other thread is simultaneously reading or writing to it.

\textbf{Analysis 4 $a_4$}:

Disclaimer: This text is generated with the o3 language model.
The model uses the following paper to generate its output: Stede, Manfred, Maite Taboada, and Debopam Das. 'Annotation guidelines for rhetorical structure.' Manuscript. University of Potsdam and Simon Fraser University (2017).
EDUs:
1. There are two types variables: shared and private.
2. Any variable that is declared outside a parallel region is shared,
3. while any declared inside is private.
4. In case of a shared variable, there is only one copy of it,
5. and all threads refer to the same variable.
6. Care is needed
7. whenever you refer to such a variable.
8. On the other hand, in case of any private variable, each thread has its own copy of it.
9. Such variables are always safe to use.
10. If a shared variable is read-only,
11. you can safely refer to it from multiple threads inside the parallel region.
12. if any thread ever writes to a shared variable,
13. then proper coordination is needed
14. to ensure that no other thread is simultaneously reading or writing to it.

RST tree (nested brackets, relation names in CAPS, N/S labels, EDU numbers in parentheses):

JOINT [nuclei]
 ---ELABORATION
      --- Nucleus (1)
      --- Satellite
            --- CONTRAST [nuclei]
                  --- (2)
                  --- (3)
 --- CONTRAST [nuclei]
      --- Cause
           --- Nucleus
                --- CIRCUMSTANCE
                      --- Nucleus (6)
                      --- Satellite (7)
           --- Satellite
                 --- CONJUNCTION [nuclei]
                       --- (4)
                       --- (5)
      --- Cause
            --- Nucleus (9)
            --- Satellite (8)
 --- CONTRAST [nuclei]
       --- CONDITION
            --- Satellite (10)
            --- Nucleus (11)
       --- CONDITION
             --- Satellite (12)
             --- PURPOSE
                   --- Nucleus (13)
                   --- Satellite (14)

\begin{figure}[ht!]
    \centering
    \includegraphics[width=1\linewidth]{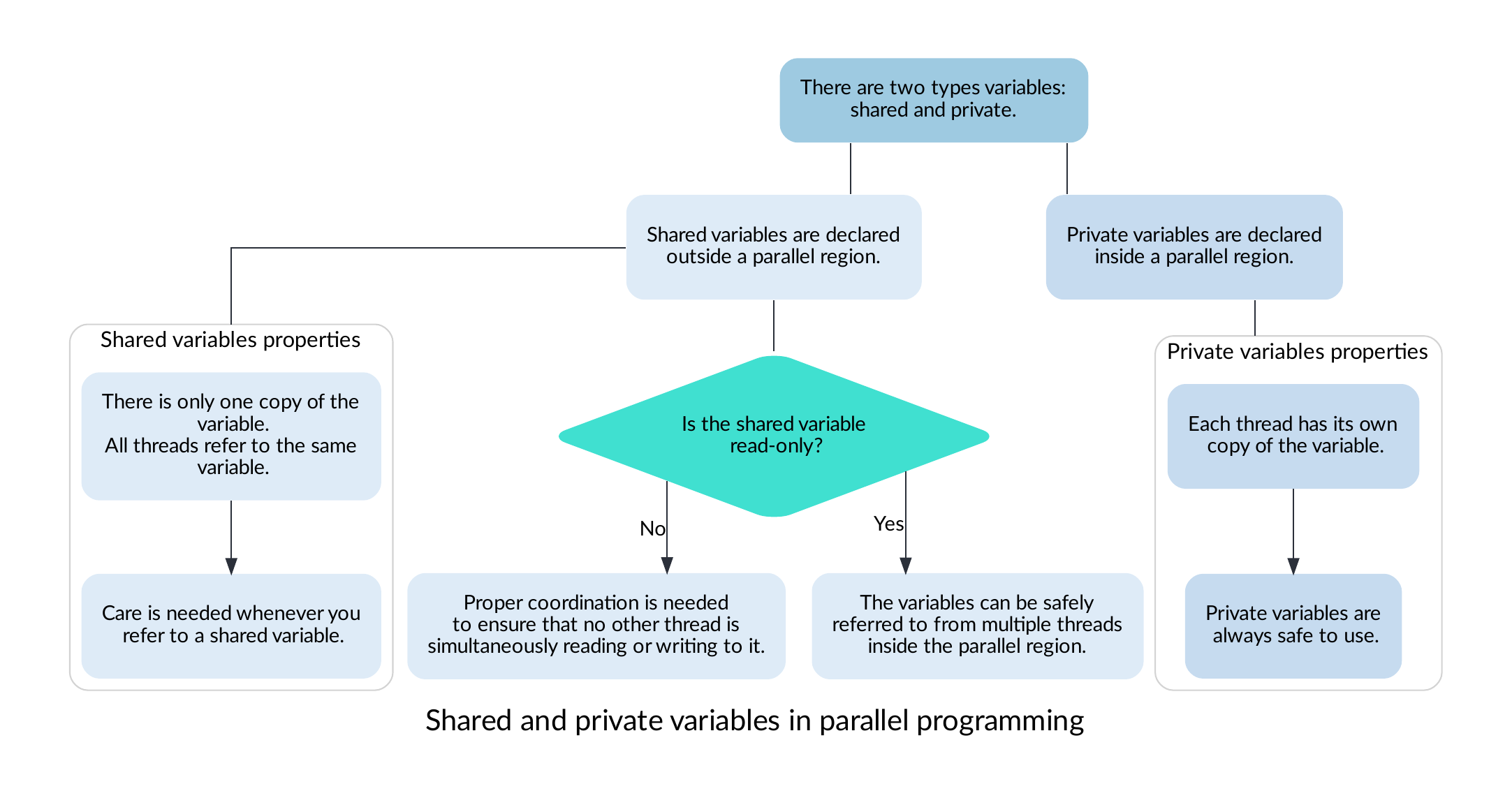}
    \caption{\textbf{Diagram 4 for Text 4}}
    \label{fig:var}
\end{figure}
\clearpage
\section{Diagram Evaluation Rubric}
\begin{strip}
\centering
\begin{tabular}{lp{0.8\linewidth}}
 \multicolumn{2}{c}{\textbf{Connectivity criteria}}\\
     \toprule
         1& The diagram's elements are fully disconnected or do not form a unified whole. Even with a lot effort the reader cannot follow it. \\\hline
         2& The reader can follow the diagram with considerable effort. The diagram contains several orphan nodes that should be part of a sequence or group, for example:
         \begin{itemize}
             \item If it is a flowchart, it has a break leaving several nodes in a sequence disconnected; 
             \item The diagram is missing an edge connecting a parent node to its children, OR many elements are connected in a random, non-uniform fashion.
         \end{itemize}\\\hline
         3& The diagram contains an orphan node that should be part of a sequence or group, e.g., if it is a flowchart, it might have a break leaving two or more nodes in a sequence disconnected; OR the diagram has some of its elements connected in a random, non-uniform fashion. 
The reader has some trouble following the diagram. \\\hline
         4& The diagram does not contain orphan nodes that should be part of a sequence or group. If it is a flowchart, it should have no breaks leaving two or more nodes in a sequence disconnected. The diagram's elements are connected in a uniform fashion. There might a minor issue with the diagram that does not affect the reader's comprehension, e.g., an unnecessary enclosure or a dashed edge instead of a solid one.\\\hline
         5& The diagram does not contain orphan nodes that should be part of a sequence or group. If it is a flowchart, it should have no breaks leaving two or more nodes in a sequence disconnected. The diagram's elements are connected in a uniform fashion. There are no issues detected with the connectivity of the diagram, e.g., no displaced or missing edges.\\
         \bottomrule
    \end{tabular}
\end{strip}

\begin{strip}
    \centering
    \begin{tabular}{lp{0.8\linewidth}}
 \multicolumn{2}{c}{\textbf{Layout Aesthetic}}\\
    \toprule
    & Each criterion is equal in weight, the grade depends on the number of issues present.\\\hline
         $1 (\leq5)$& \multirow{4}{=}{The diagram has the following issues: 1) line crossings and/or excessive number of bends, 2) elements overlapping or obscuring each other, 3) elements impossible to comprehend due to their color, size, or shape (e.g., a dark text color on the dark background). 4) The diagram is asymmetrical. 5) The lines are not aligned horizontally or vertically. 6) The diagram is too wide to fit the reader's display. 7) The diagram's layout is dishomogeneous --- for example, the nodes are distributed randomly, they can be of different sizes and randomly colored (a variation due to, for instance, varied text length within nodes or gradient colors for flowcharts is permitted).}\\
         $2(=4)$& \\
         $3(=3)$& \\
         $4(3\leq)$& \\
 &\\
 &\\
 &\\
 &\\
 &\\
 &\\
 &\\\hline
         $5$& The diagram fulfills these criteria: 1) line crossings and/or excessive number of bends are avoided, 2) no elements overlapping or obscuring each other, 3) no elements that are impossible to comprehend due to their color, size, or shape (e.g., a dark text color on a dark background). 4) The diagram is symmetrical. 5) The lines are aligned horizontally or vertically. 6) The diagram is not too wide to fit the reader's display. 7) The diagram's appearance is homogeneous -- for example, the nodes are distributed uniformly, they are mostly of similar sizes and colors (a variation due to, for instance, varied text length within nodes or gradient colors for flowcharts is permitted).\\
         \bottomrule
    \end{tabular}
\end{strip}
\begin{strip}
    \begin{tabular}{lp{0.9\linewidth}}
 \multicolumn{2}{c}{\textbf{Logical Organization Criteria}}\\
    \toprule
         & Each criterion is followed by its weight. An example for when the criteria differ a lot: a 5 for a), a 3 for b), and a 5 for c) -> the grade: 5*0.6+3*0.3+5*0.1 = 4.4, rounded to a 4. When rounding, x $\leq$ .5 is rounded up. \\\hline
         1& a) The diagram's flow and structure do not make sense, e.g., a sequence of actions is depicted as a tree or a set of completely disjoint nodes. The reader cannot follow the diagram. 0.6
b) There is a number of language issues, and the reader cannot comprehend the text. 0.3
c) Widely accepted diagram conventions are not honored making it impossible to understand. 0.1\\\hline
         2& a) The diagram's flow and structure barely make sense, e.g., a sequence of actions is depicted as a list or a flowchart whose edges and nodes do not form a sensible sequence.   0.6
b) There is a number of language issues, and the reader can barely comprehend the text.  0.3
c) The diagram breaks multiple flowchart conventions when they are applicable, e.g., a third edge coming from a conditional block when inappropriate, one or more nodes are of a wrong shape (a square when a diamond would be more appropriate), undirected edges are used to depict sequences, it might contain closed cycles or loops when unneeded, labels might not correspond to edges or nodes. 0.1\\\hline
         3& a) The diagram's flow and structure are more or less logical, e.g., a flowchart depicts a somewhat sensible sequence of events from the beginning to the end, a tree might partially depict a hierarchical structure, a list shows a group of somewhat similar and related elements. 0.6
b) There is a number of language issues, but the reader can comprehend the text without much effort. 0.3
c) The diagram breaks one flowchart convention when they are applicable, e.g., a third edge coming from a conditional block when inappropriate, one or more nodes are of a wrong shape (a square when a diamond would be more appropriate), undirected edges are used to depict sequences. 0.1\\\hline
         4& a) The diagram's flow and structure are mostly logical, e.g., a flowchart depicts a sensible sequence of events from the beginning to the end, a tree depicts a hierarchical structure, a list shows a group of similar and related elements, but there might be small issues present, e.g., a logical flowchart that is not perfectly fluent. The diagram requires a small improvement, e.g., adding an additional label. 0.6
b) The language is mostly clear and free of grammar mistakes. There might be an insignificant grammar mistake that does not prevent the reader from understanding the diagram. 0.3
c) The diagram does not break flowchart conventions when they are applicable, e.g., a third edge coming from a conditional block when inappropriate. 0.1\\\hline
         5& a) The diagram's flow and structure are logical, e.g., a flowchart depicts a sensible sequence of events from the beginning to the end, a tree depicts a hierarchical structure, a list shows a group of similar and related elements. The diagram does not require any improvements to its structure and organization and can be easily read as is. 0.6
b) The language is clear and free of grammar mistakes and easy to comprehend. There might be an occasional awkward phrasing or word choice. 0.3
c) The diagram does not break flowchart conventions when they are applicable, e.g., a third edge coming from a conditional block when inappropriate. 0.1\\
\bottomrule
    \end{tabular}
\end{strip}
\mbox{}
\clearpage
\section{Bayesian Data Analysis: Human Evaluation}
The linear predictor and the family for each specific outcome is described in (\ref{model}). We use weakly informative priors as constrains in (\ref{priors1}, \ref{priors2}) to allow for substantial effect sizes except for the $\sigma_u$ for source text level intercepts. Certain source texts (mostly short and simple) produce very little hallucinated instances and better quality scores and introduce sample level quasi-separation. The regularizing prior of $Exp(1)$ for a varying intercept prevents the models from over-fitting in this case.

\begin{figure*}[ht!]
    \centering
     \begin{subfigure}[b]{0.49\textwidth}
         \centering
         \includegraphics[width=\textwidth]{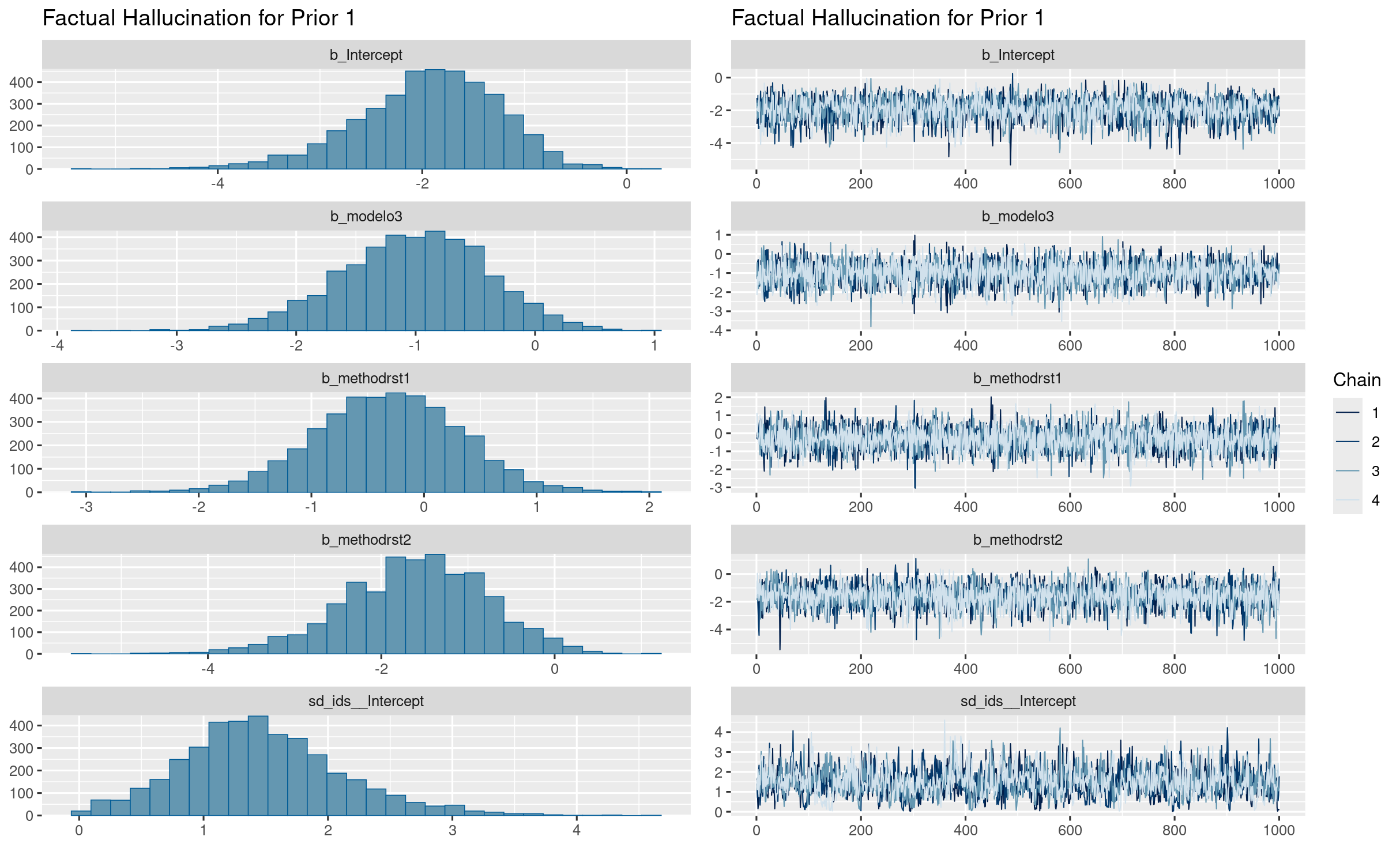}
         \caption{Factual Hallucination $F_{fact}$ with $Pr_1$}
         \label{fig:postfact1}
     \end{subfigure}
     \hfill
     \begin{subfigure}[b]{0.49\textwidth}
         \centering
         \includegraphics[width=\textwidth]{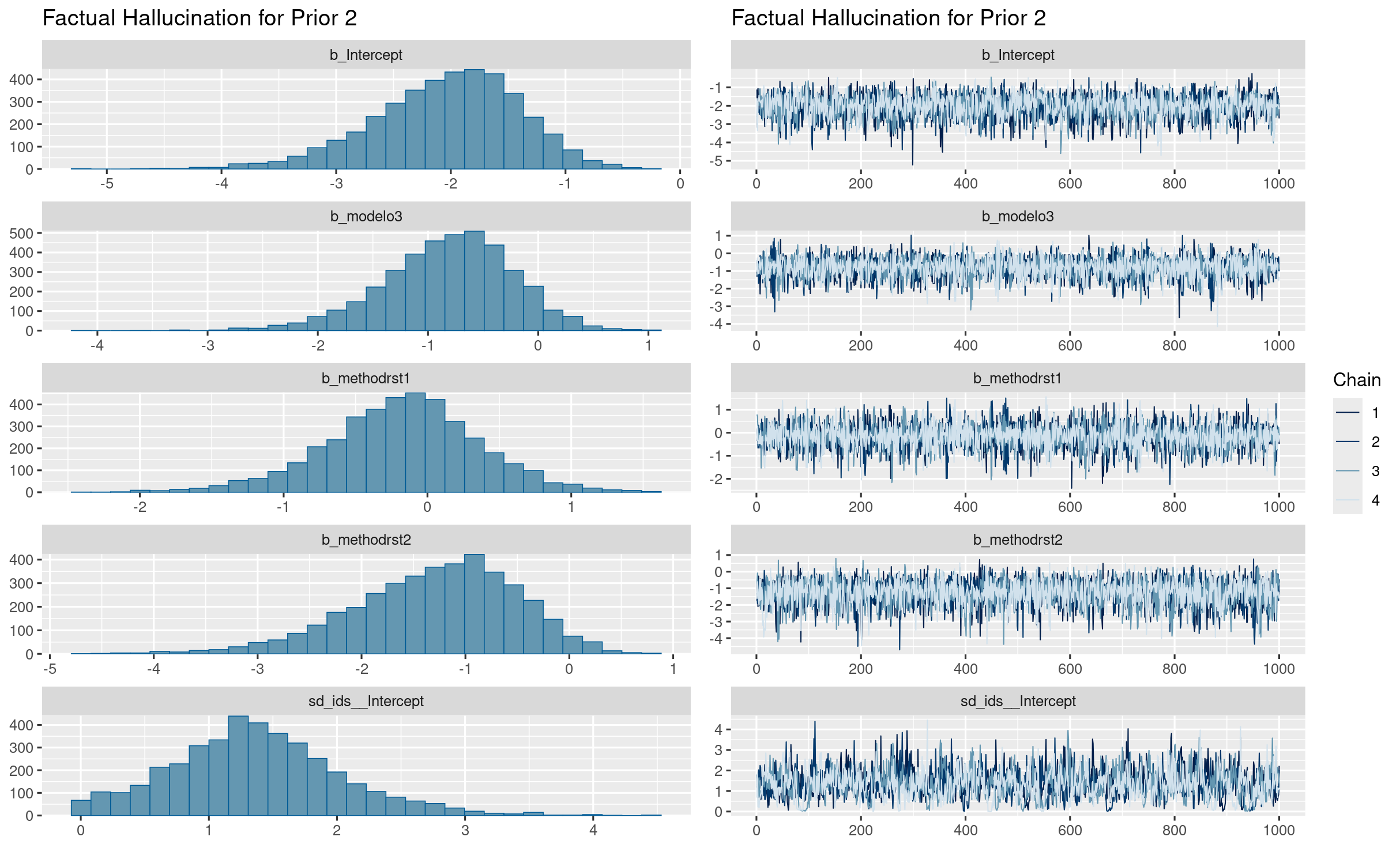}
         \caption{Factual Hallucination $F_{fact}$ with $Pr_2$}
         \label{fig:postfact2}
     \end{subfigure}
     \hfill
     \begin{subfigure}[b]{0.49\textwidth}
         \centering
         \includegraphics[width=\textwidth]{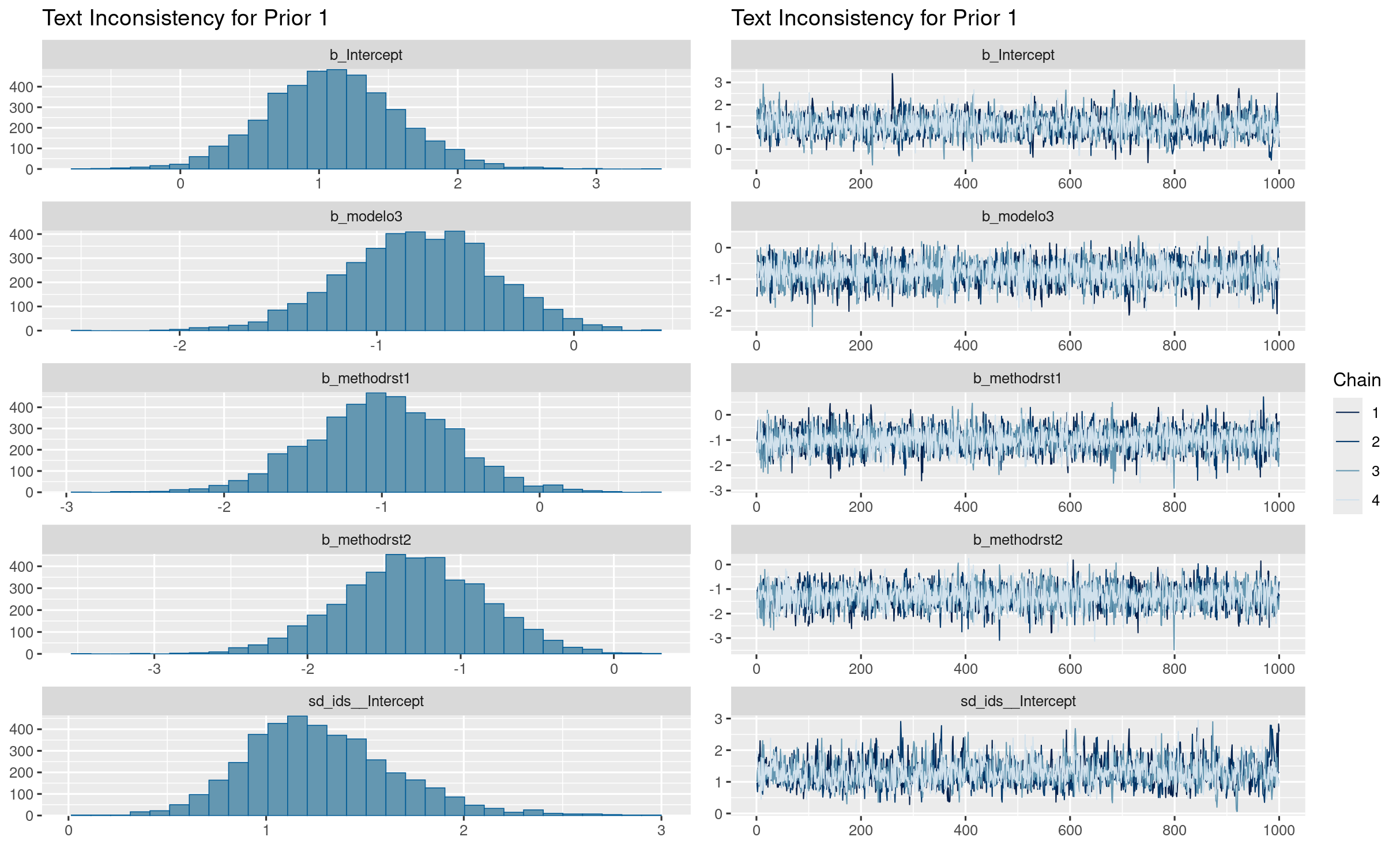}
         \caption{Text Inconsistency $F_c$ with $Pr_1$}
         \label{fig:posttext1}
     \end{subfigure}
     \begin{subfigure}[b]{0.49\textwidth}
         \centering
         \includegraphics[width=\textwidth]{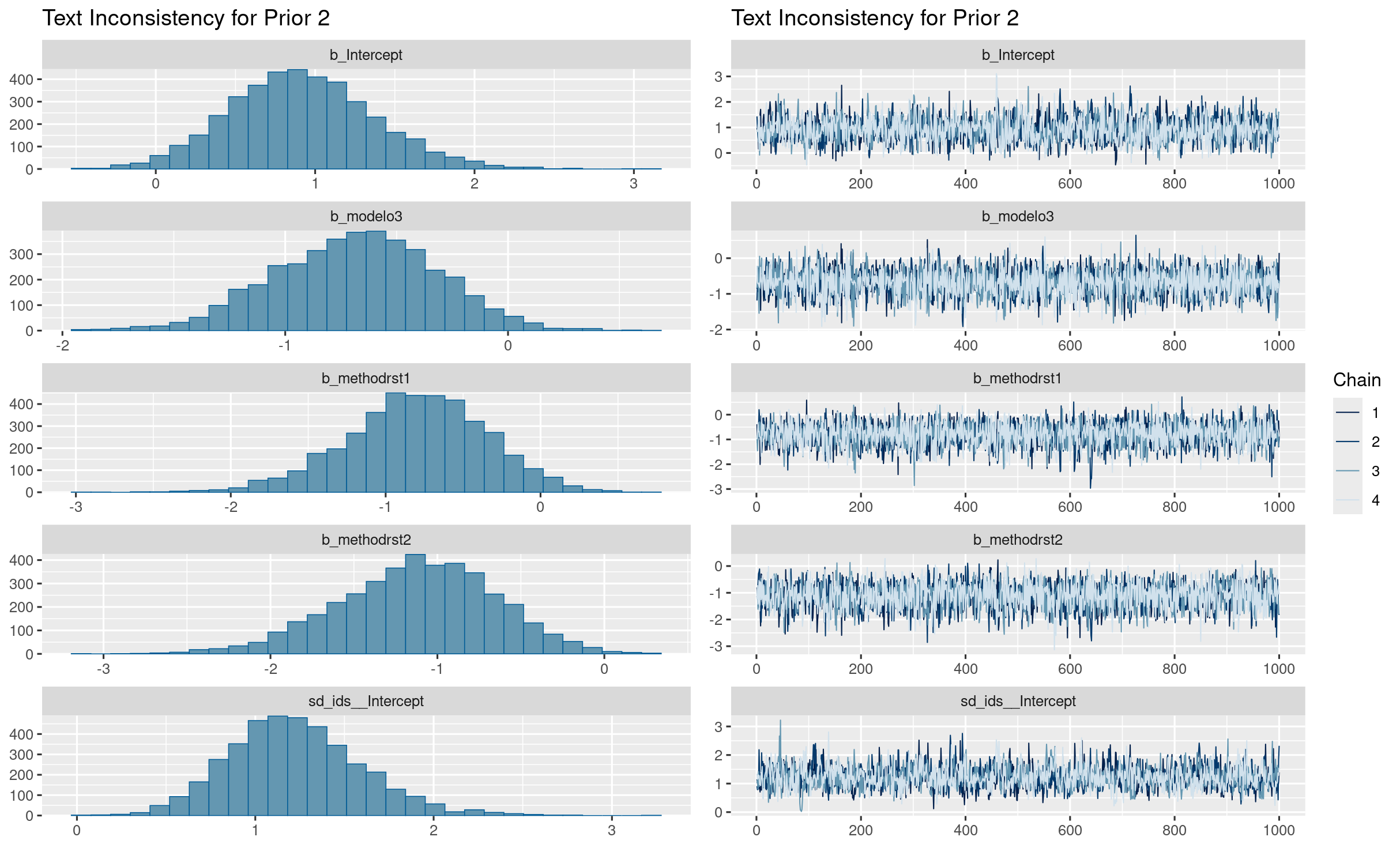}
         \caption{Text Inconsistency $F_c$ with $Pr_2$}
         \label{fig:posttext2}
     \end{subfigure}
        \begin{subfigure}[b]{0.49\textwidth}
         \centering
         \includegraphics[width=\textwidth]{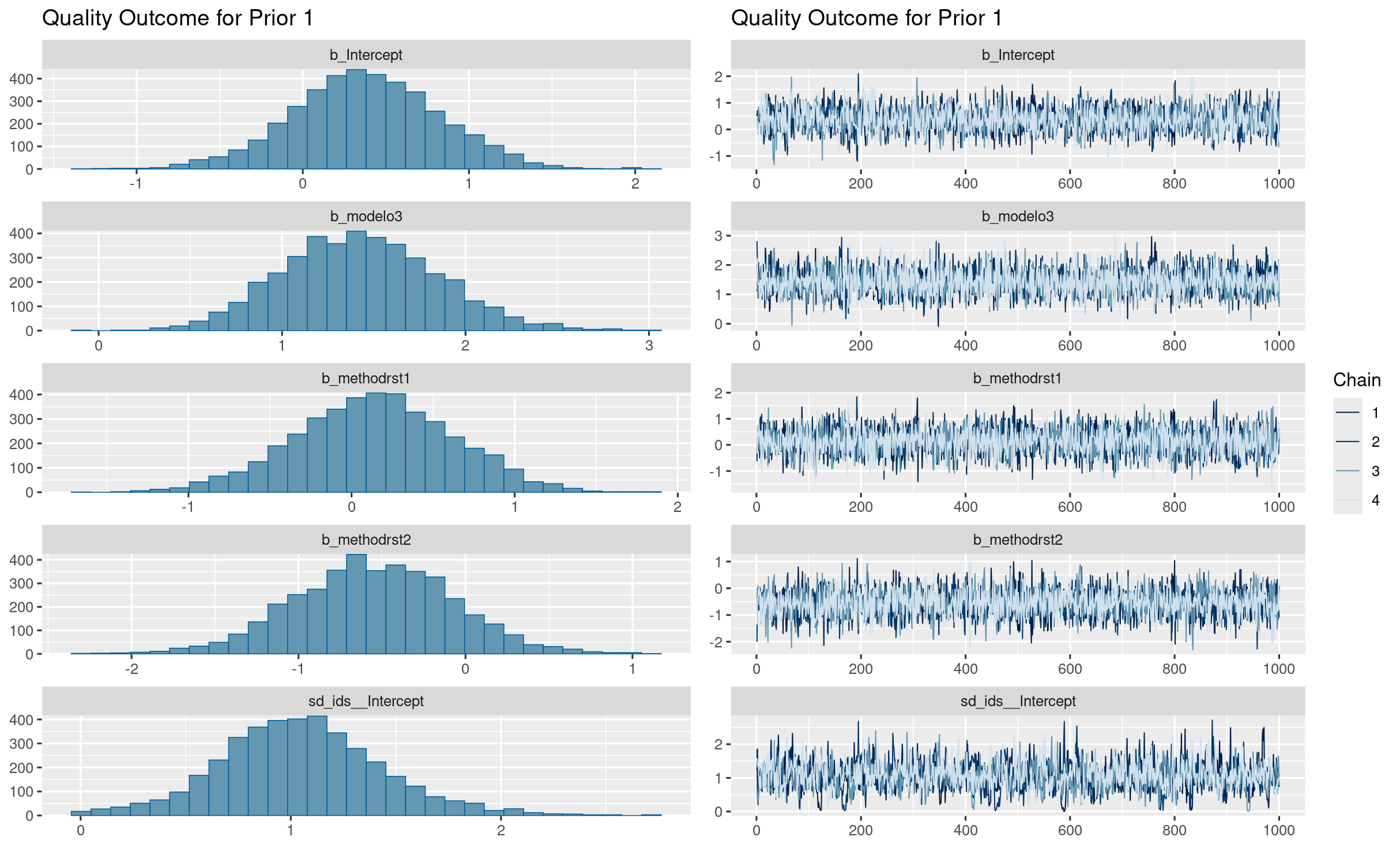}
         \caption{Quality Outcome $G$ with $Pr_1$}
         \label{fig:postquality1}
     \end{subfigure}
     \begin{subfigure}[b]{0.49\textwidth}
         \centering
         \includegraphics[width=\textwidth]{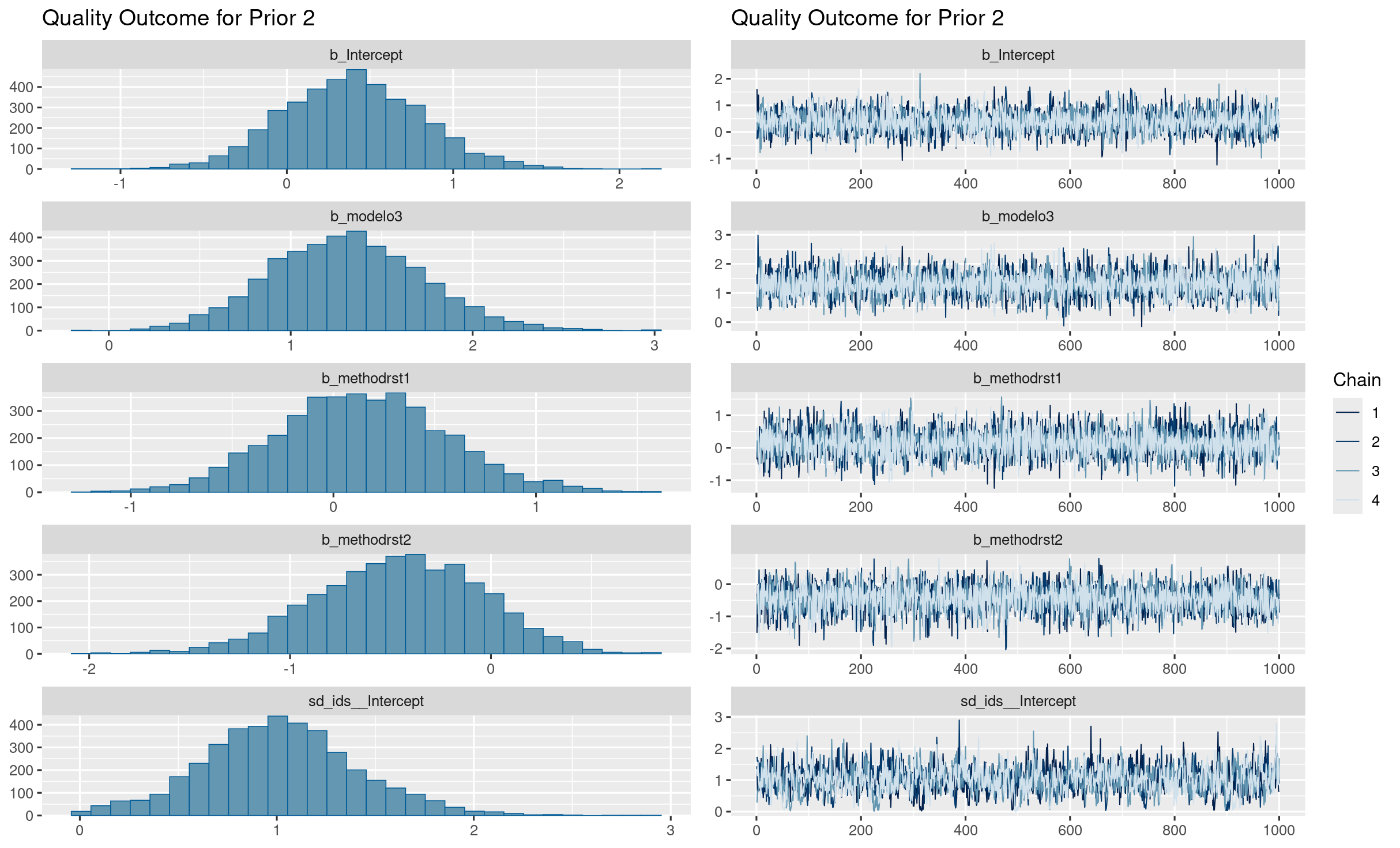}
         \caption{Quality Outcome $G$ for $Pr_2$}
         \label{fig:postquality2}
     \end{subfigure}
        \caption{Posterior density and trace plots.}
        \label{fig:post_check}
\end{figure*}

Table \ref{tab:diagnostics} show the LOO estimate of the expected log pointwise predictive density, the standard error, the number of effective parameters, and the number of Pareto $\hat{k}$ > 0.7 \cite{vehtari2024pareto, vehtari2021rank}. The convergence details are provided in Table \ref{tab:convergence}, we used the default 4 chains with 4000 total post-warmup draws. All $\hat{R}$ $\leq$ 1.01 and minimum bulk effective sample size > 400 signal that the chains have mixed \citep{vehtari2017practical}. There was 2 divergent transitions 
 in the fit for the factual hallucination so we increased the \textit{adapt\_delta} to 0.95 \citep{betancourt2017delta}. We provide the trace plots in Fig. \ref{fig:post_check}.
\begin{figure}[ht!]
    \centering
     \begin{subfigure}[b]{0.31\textwidth}
         \centering
         \includegraphics[width=\textwidth]{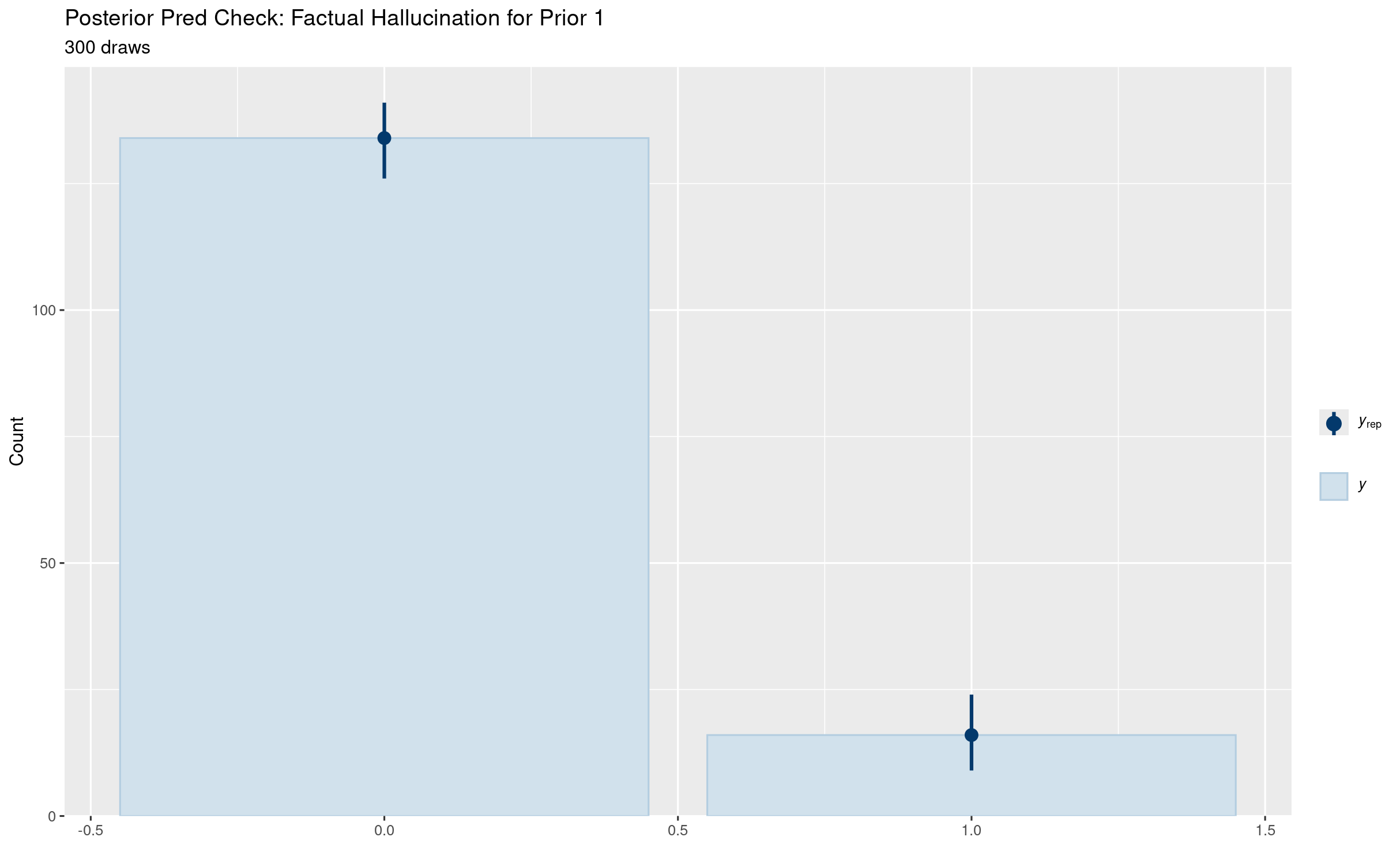}
         \caption{$F_{fact}$ with $Pr_1$}
         \label{fig:ppfact1}
     \end{subfigure}
     \begin{subfigure}[b]{0.31\textwidth}
         \centering
         \includegraphics[width=\textwidth]{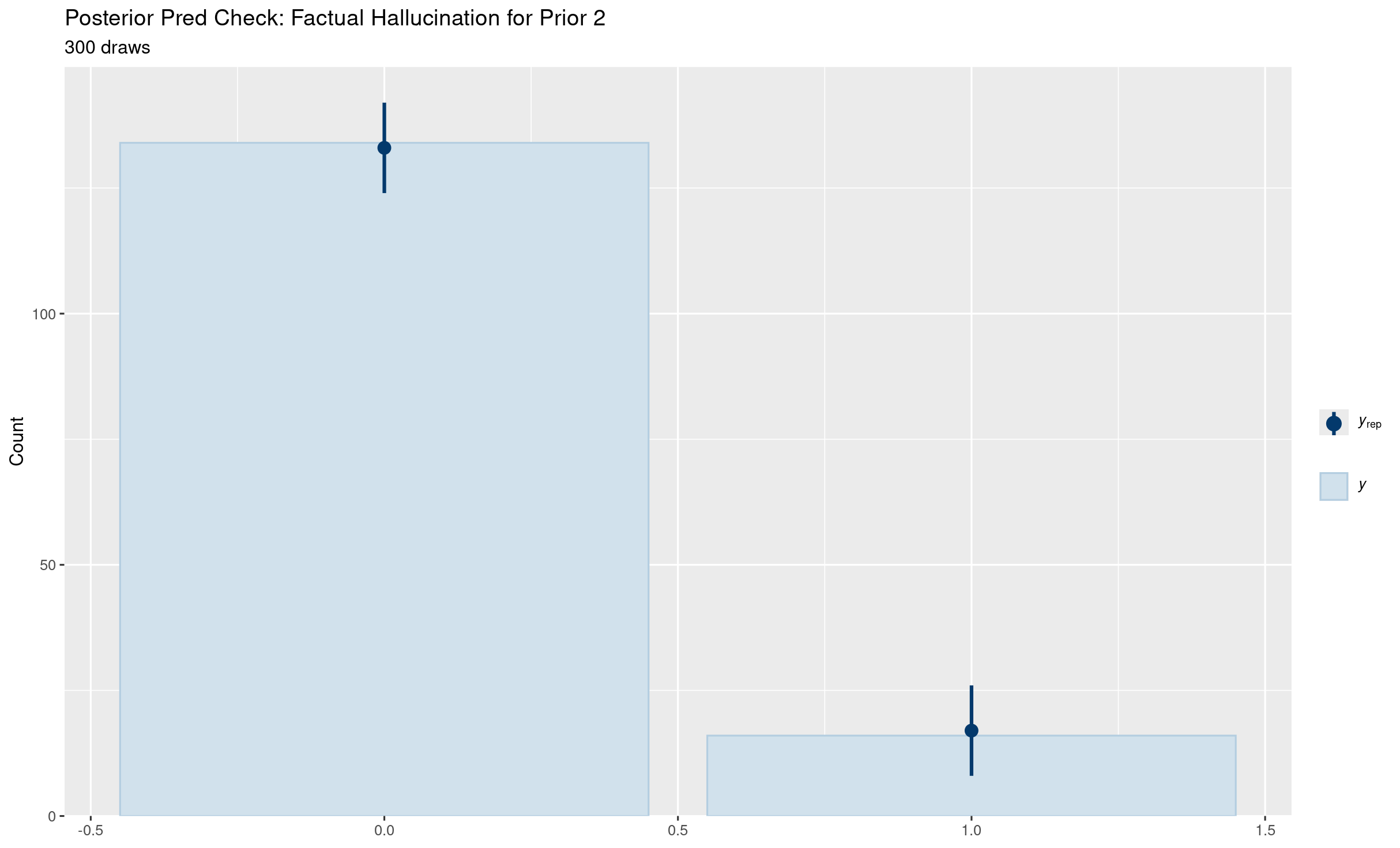}
         \caption{$F_{fact}$ with $Pr_2$}
         \label{fig:ppfact2}
     \end{subfigure}
     \begin{subfigure}[b]{0.31\textwidth}
         \centering
         \includegraphics[width=\textwidth]{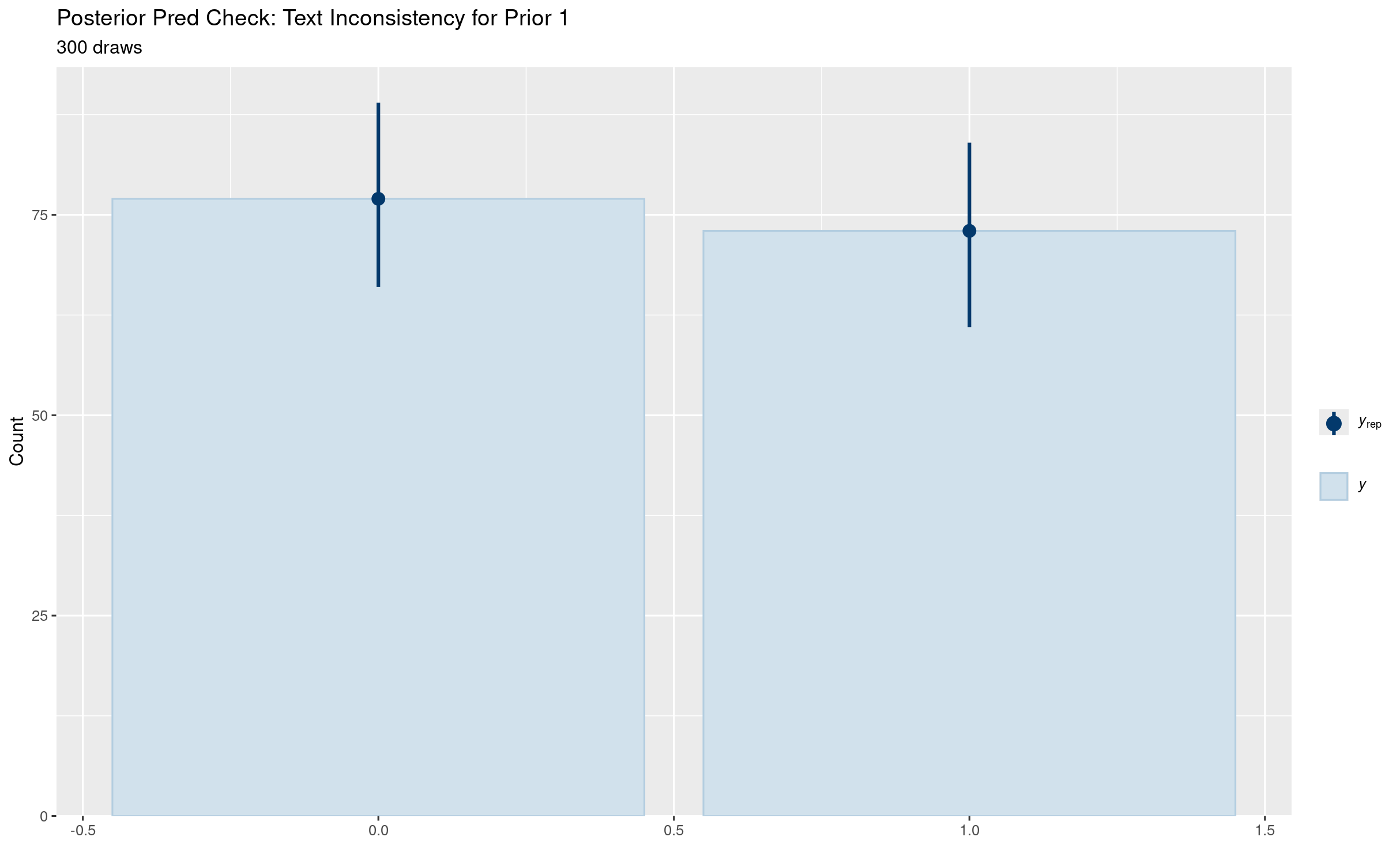}
         \caption{$F_c$ with $Pr_1$}
         \label{fig:pptext1}
     \end{subfigure}
     \begin{subfigure}[b]{0.31\textwidth}
         \centering
         \includegraphics[width=\textwidth]{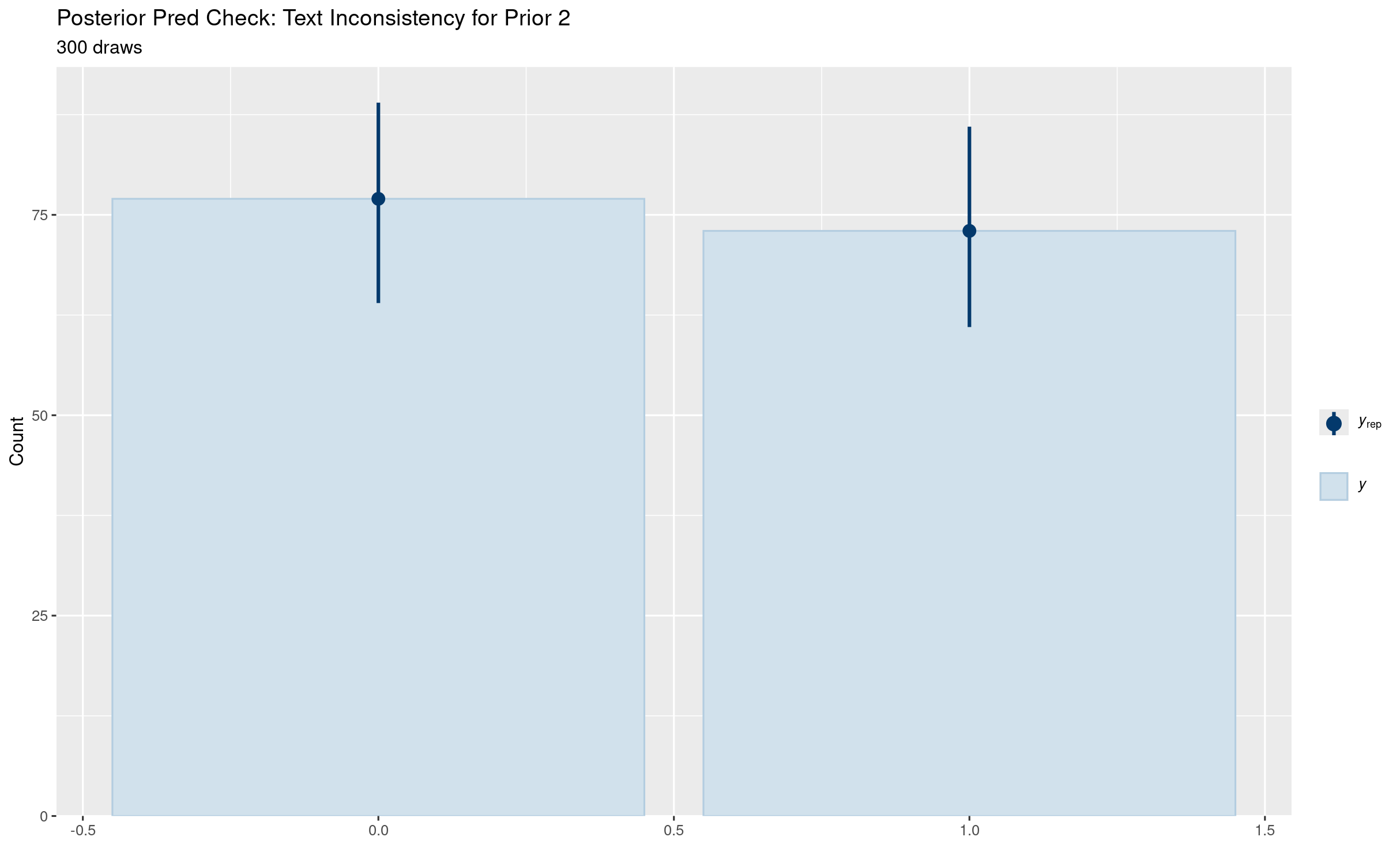}
         \caption{$F_c$ with $Pr_2$}
         \label{fig:pptext2}
     \end{subfigure}
        \begin{subfigure}[b]{0.31\textwidth}
         \centering
         \includegraphics[width=\textwidth]{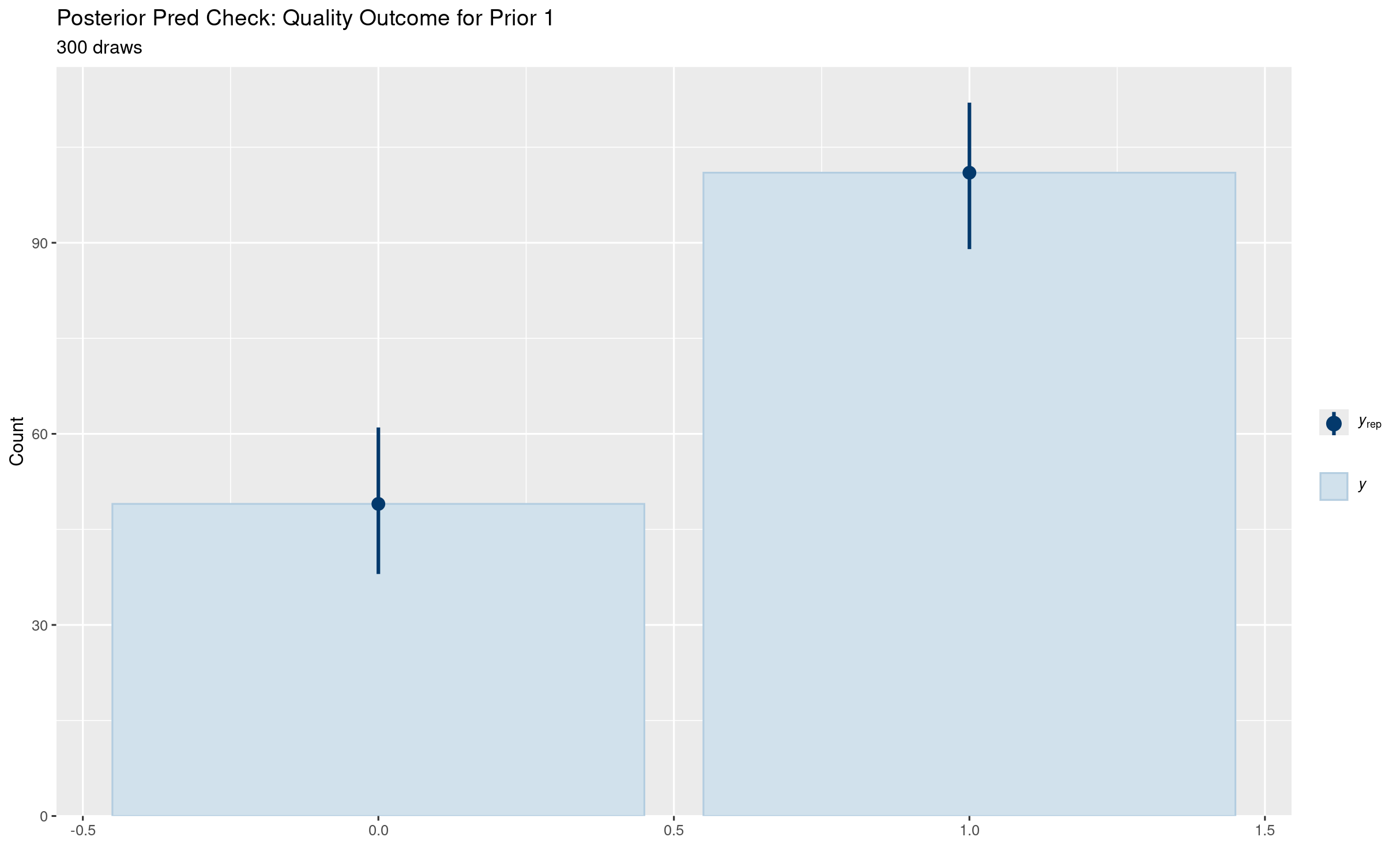}
         \caption{$G$ with $Pr_1$}
         \label{fig:ppquality1}
     \end{subfigure}
     \begin{subfigure}[b]{0.31\textwidth}
         \centering
         \includegraphics[width=\textwidth]{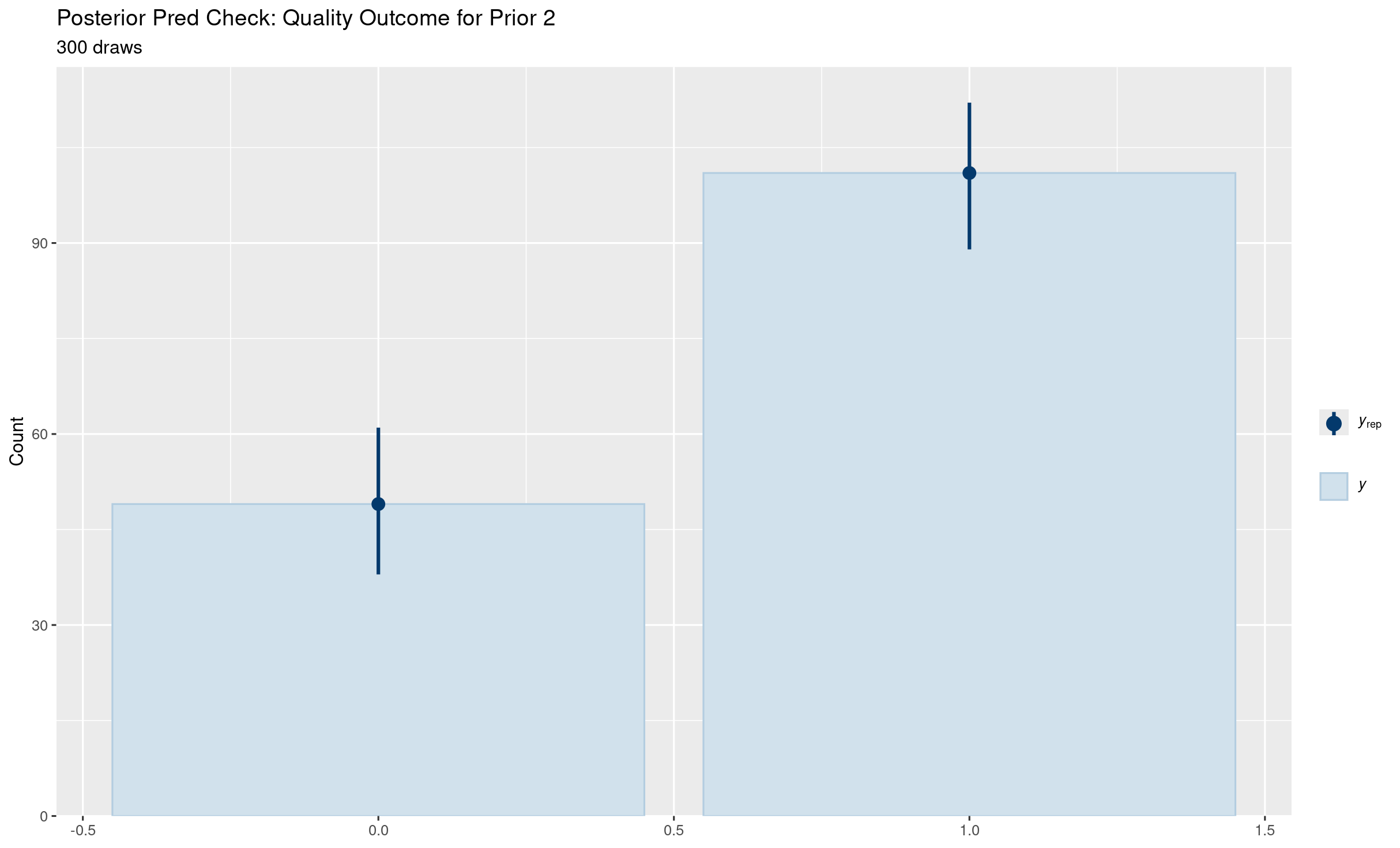}
         \caption{$G$ for $Pr_2$}
         \label{fig:ppquality2}
     \end{subfigure}
        \caption{Posterior predictive checks for factual hallucination $F_{fact}$, context inconsistency $F_c$, quality outcome $G$.}
        \label{fig:pp_check}
\end{figure}

 We report the posterior estimates in Table \ref{tab:posterior} and posterior density plots in Fig. \ref{fig:post_check}. The posterior predictive checks visualized in Fig. \ref{fig:pp_check} reflect the uncertainty of the posterior predictive distributions with the intervals for the replicated $y_{rep}$ counts being wide; however, the observed counts appear to fall within the range of the replicated values. Given the sample size, we interpret the posterior predictive plots as a consistency checks.

\begin{table}[h!]
        \caption{Convergence diagnostics for the Brms fits. $F_{fact}$ stands for factual hallucination, $F_{c}$ --- context faithfulness, $G$---quality outcome. (1) is a fit with $Pr_1$, (2) --- $Pr_2$}
    \label{tab:convergence}
    \begin{tabular}{llll}
    \toprule
Fit & max $\hat{R}$ & min bulk ESS & divergent C \\
\midrule
$F_c$ (1) & 1.01 & 1015.31 & 0 \\
$F_c$ (2) & 1.01 & 955.96 & 0 \\
$F_{fact}$(1) & 1.00 & 872.10 & 0 \\
$F_{fact}$(2) & 1.01 & 684.24 & 0 \\
$G$(1) & 1.00 & 933.78 & 0 \\
$G$(2) & 1.00 & 878.40 & 0 \\
\bottomrule
    \end{tabular}
\end{table}

\begin{table*}[h!]
    \caption{Diagnostics for the Brms fits. $F_{fact}$ stands for factual hallucination, $F_{c}$ --- context faithfulness, $G$---quality outcome. (1) is a fit with $Pr_1$, (2) --- $Pr_2$}
    \label{tab:diagnostics}
    \begin{tabular}{llllll}
    \toprule
Fit & $ELPD_{LOO}$ & $SE_{ELPD}$ & $P_{LOO}$ & $R^2_{bayes}$ & Pareto $\hat{k}$ \\\midrule
$F_{fact}$(1) & -47.07 & 7.60 & 13.11 & 0.21 & 0 \\
$F_{fact}$(2) & -47.29 & 7.48 & 12.35 & 0.19 & 0 \\
$F_c$ (1)  & -93.50 & 5.27 & 18.75 & 0.27 & 0 \\
$F_c$ (2)  & -93.89 & 4.83 & 18.27 & 0.25 & 0 \\
$G$(1) & -87.79 & 5.99 & 16.43 & 0.23 & 0 \\
$G$(2) & -87.58 & 5.71 & 15.65 & 0.22 & 0 \\
\bottomrule
    \end{tabular}
\end{table*}

\begin{table*}[h!]
\caption{Posterior Estimates for the Brms fits. $F_{fact}$ stands for factual hallucination, $F_{c}$ --- context faithfulness, $G$---quality outcome. (1) is a fit with $Pr_1$, (2) --- $Pr_2$}
\label{tab:posterior}
\begin{tabular}{lllllll}
\toprule
\centering
Effect & Term & Estimate & std.error & conf.low & conf.high & Fit \\
\midrule
fixed & (Intercept) & -1.94 & 0.69 & -3.46 & -0.75 & $F_{fact}$(1) \\
fixed & o3 & -1.02 & 0.61 & -2.27 & 0.15 & $F_{fact}$(1) \\
fixed & RST1 & -0.32 & 0.65 & -1.61 & 0.92 & $F_{fact}$(1) \\
fixed & RST2 & -1.61 & 0.83 & -3.35 & -0.06 & $F_{fact}$(1) \\
varying & sd(Intercept) & 1.47 & 0.65 & 0.28 & 2.95 & $F_{fact}$(1) \\
\midrule
fixed & (Intercept) & -2.04 & 0.64 & -3.43 & -0.94 & $F_{fact}$(2) \\
fixed & o3 & -0.82 & 0.59 & -2.09 & 0.26 & $F_{fact}$(2) \\
fixed & RST1 & -0.18 & 0.54 & -1.32 & 0.87 & $F_{fact}$(2) \\
fixed & RST2 & -1.30 & 0.80 & -3.06 & 0.03 & $F_{fact}$(2) \\
varying & sd(Intercept) & 1.37 & 0.67 & 0.12 & 2.81 & $F_{fact}$(2) \\
\midrule
fixed & (Intercept) & 1.10 & 0.48 & 0.18 & 2.04 & $F_c$ (1) \\
fixed & o3 & -0.79 & 0.38 & -1.55 & -0.07 & $F_c$ (1) \\
fixed & RST1 & -1.00 & 0.45 & -1.90 & -0.15 & $F_c$ (1) \\
fixed & RST2 & -1.31 & 0.46 & -2.23 & -0.42 & $F_c$ (1) \\
varying & sd(Intercept) & 1.28 & 0.38 & 0.62 & 2.12 & $F_c$ (1) \\
\midrule
fixed & (Intercept) & 0.92 & 0.46 & 0.07 & 1.87 & $F_c$ (2) \\
fixed & o3 & -0.68 & 0.36 & -1.40 & 0.00 & $F_c$ (2) \\
fixed & RST1 & -0.81 & 0.46 & -1.76 & 0.04 & $F_c$ (2) \\
fixed & RST2 & -1.12 & 0.47 & -2.10 & -0.26 & $F_c$ (2) \\
varying & sd(Intercept) & 1.22 & 0.37 & 0.56 & 2.01 & $F_c$ (2) \\
\midrule
fixed & (Intercept) & 0.38 & 0.44 & -0.51 & 1.26 & $G$(1) \\
fixed & o3 & 1.43 & 0.42 & 0.64 & 2.27 & $G$(1) \\
fixed & RST1 & 0.14 & 0.49 & -0.82 & 1.06 & $G$(1) \\
fixed & RST2 & -0.55 & 0.48 & -1.49 & 0.37 & $G$(1) \\
varying & sd(Intercept) & 1.05 & 0.39 & 0.27 & 1.89 & $G$(1) \\
\midrule
fixed & (Intercept) & 0.39 & 0.42 & -0.40 & 1.24 & $G$(2) \\
fixed & o3 & 1.31 & 0.42 & 0.52 & 2.16 & $G$(2) \\
fixed & RST1 & 0.14 & 0.42 & -0.67 & 0.98 & $G$(2) \\
fixed & RST2 & -0.46 & 0.42 & -1.31 & 0.31 & $G$(2) \\
varying & sd(Intercept) & 1.02 & 0.40 & 0.22 & 1.84 & $G$(2) \\
\bottomrule
\end{tabular}
\end{table*}
\clearpage
\section{Bayesian Analysis: Automated Evaluation}

\begin{figure}[h]
    \centering
     \begin{subfigure}[b]{0.45\textwidth}
         \centering
         \includegraphics[width=\textwidth]{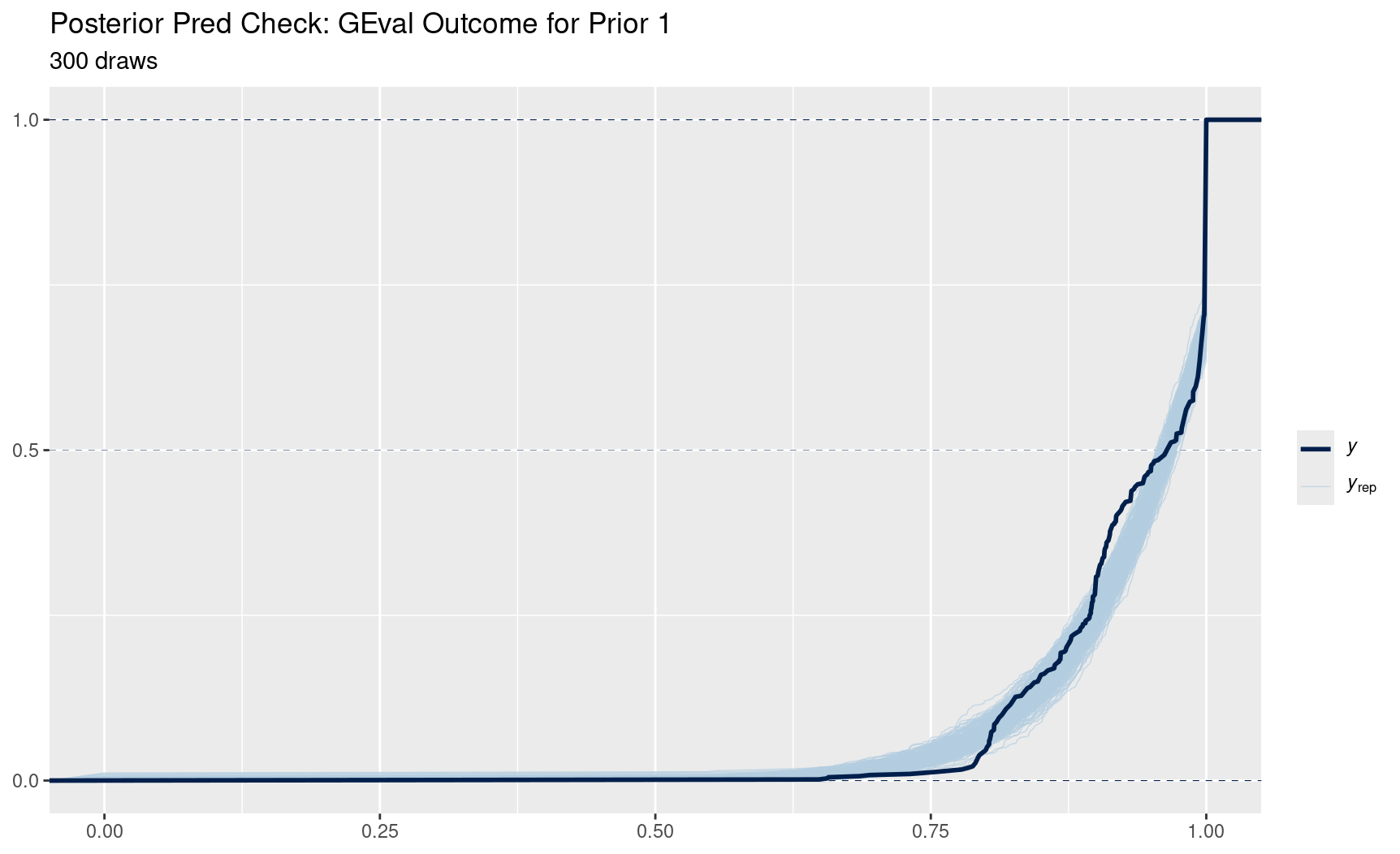}
         \caption{Posterior G-Eval with $Pr_{1b}$}
         \label{fig:autqua1}
     \end{subfigure}
     \hfill
     \begin{subfigure}[b]{0.45\textwidth}
         \centering
         \includegraphics[width=\textwidth]{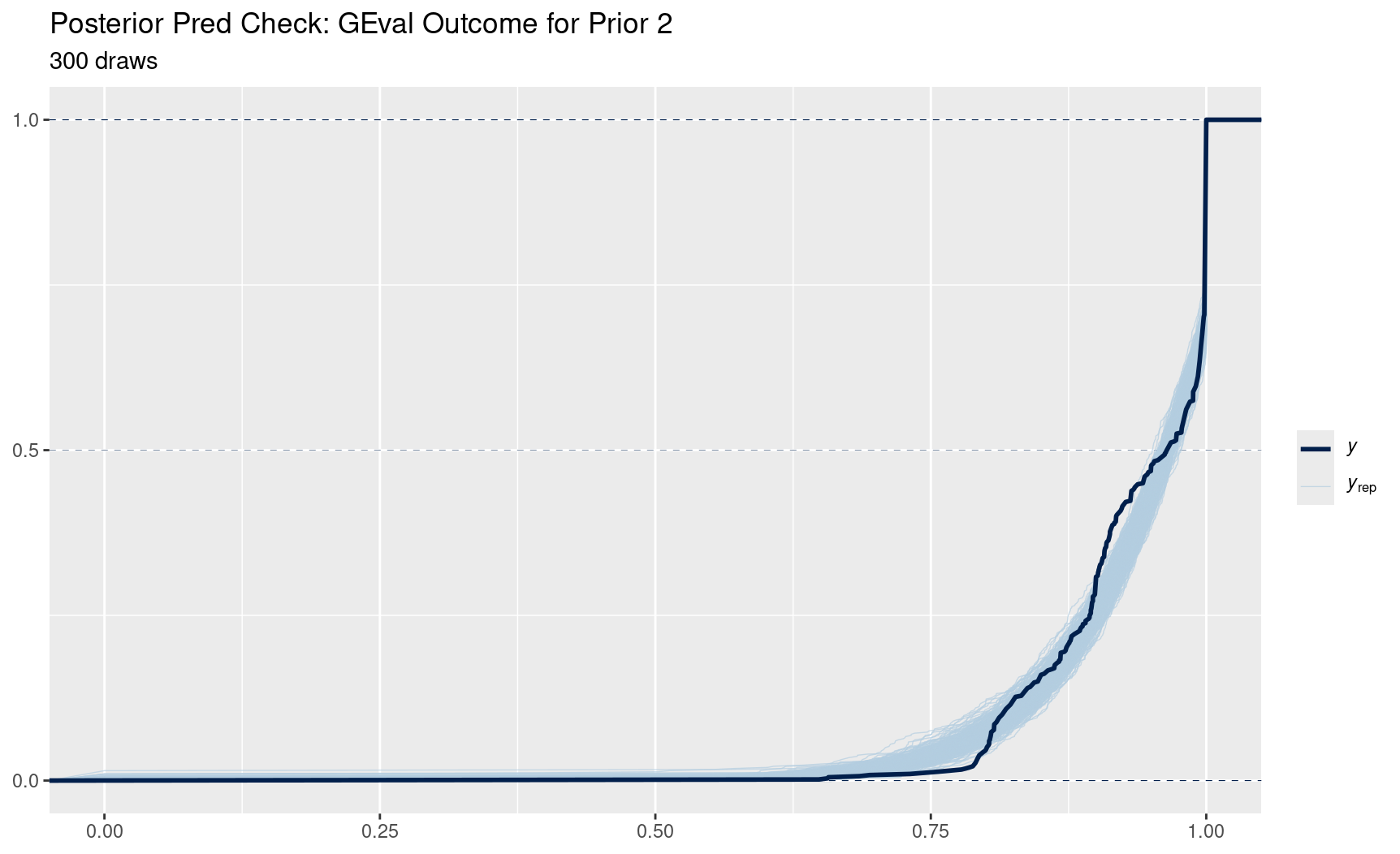}
         \caption{Posterior G-Eval with $Pr_{2b}$}
         \label{fig:autqua2}
     \end{subfigure}
     \caption{Posterior predictive checks for G-Eval.} 
     \label{geevalpp}
\end{figure}

\begin{figure*}[h!]
    \centering
     \begin{subfigure}[b]{0.4\textwidth}
         \centering
         \includegraphics[width=\textwidth]{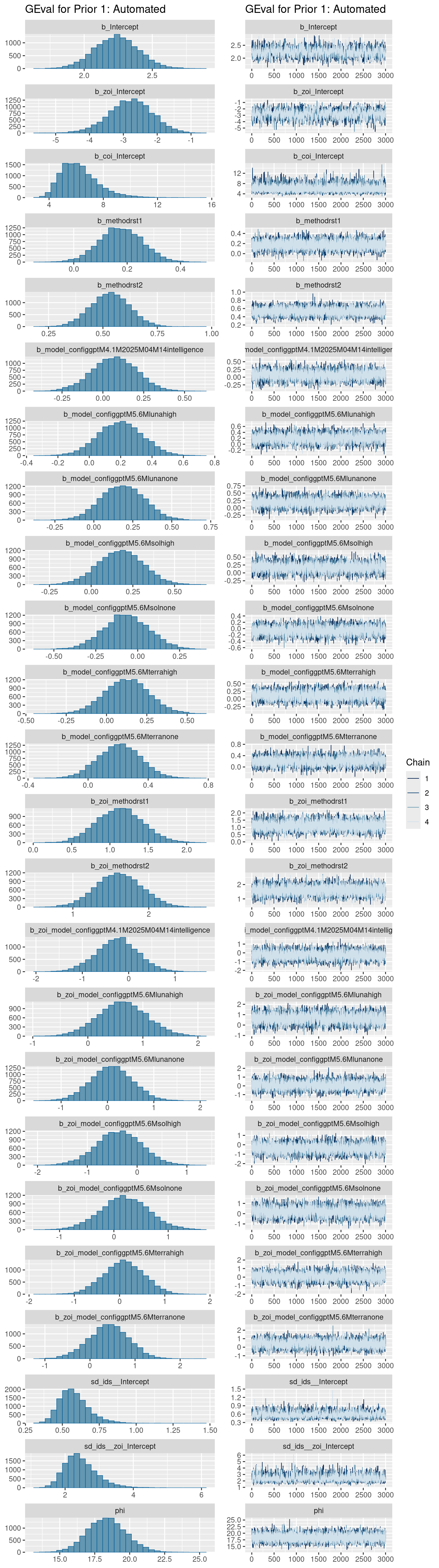}
         \caption{G-Eval with $Pr_{1b}$}
         \label{fig:geval1}
     \end{subfigure}
     \hfill
     \begin{subfigure}[b]{0.4\textwidth}
         \centering
         \includegraphics[width=\textwidth]{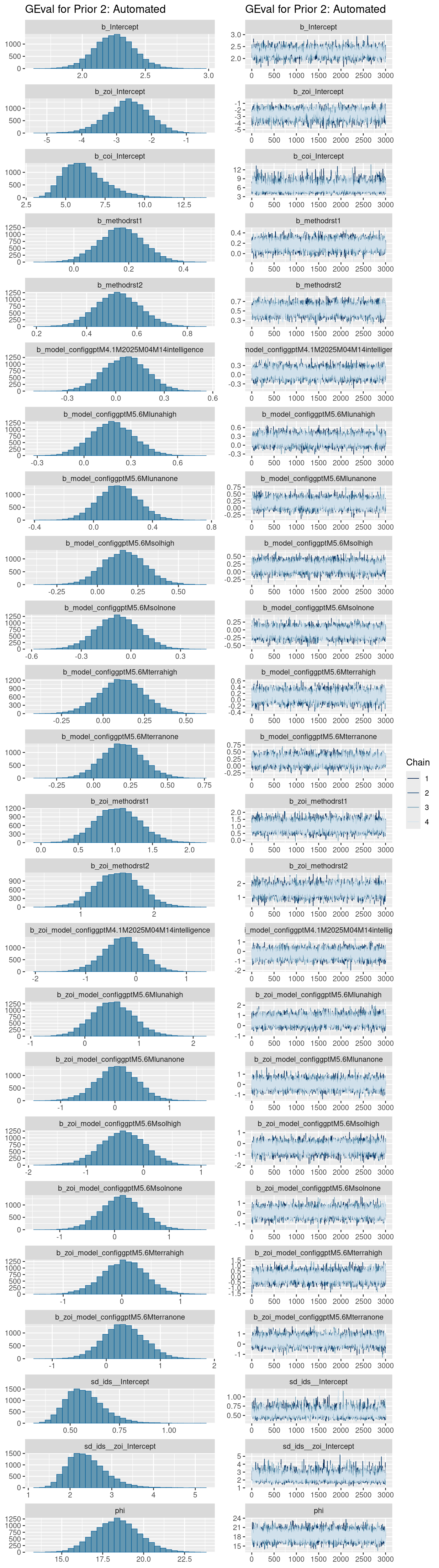}
         \caption{G-Eval with $Pr_{2b}$}
         \label{fig:geval2}
     \end{subfigure}
     \caption{Posterior density and trace plots for G-Eval.}
\end{figure*}

Table \ref{tab:convergence_aut} show the LOO estimate of the expected log pointwise predictive density, the standard error, the number of effective parameters, and the number of Pareto $\hat{k}$ > 0.7 \cite{vehtari2024pareto, vehtari2021rank}. We used the 4 chains with 12,000 total post-warmup draws. All $\hat{R}$ < 1.00 and minimum bulk effective sample size > 400 signal that the chains have mixed \citep{vehtari2017practical}. There were no divergent transitions and Pareto $\hat{k}$ > 0.7. We provide the trace plots of the brms fits in Figures \ref{fig:geval1} and \ref{fig:geval2} and posterior predictive checks plots in Figure \ref{geevalpp}. 

\begin{table}[h]
        \caption{Convergence diagnostics for the Brms fits. G-Eval, (1) is a fit with $Pr_{1b}$, (2) --- $Pr_{2b}$}
    \label{tab:convergence_aut}
    \begin{tabular}{lcc}
    \toprule
 & G-Eval(1) & G-Eval(2) \\
\midrule
$\hat{R}$ & 1.00 & 1.00 \\
min ESS & 1403.77 & 2525.27 \\
div C & 0 & 0 \\
$ELPD_{LOO}$ & 433.57 & 434.79 \\
$SE_{ELPD}$ & 31.23 & 31.15 \\
$P_{LOO}$ & 59.22 & 57.63 \\
$R^2_{bayes}$ & 0.47 & 0.47 \\
Pareto $\hat{k}$ & 0 & 0 \\
\bottomrule
    \end{tabular}
\end{table}

\begin{table*}[h!]
\caption{Posterior Estimates for the Brms fits for automated assessment.G-Eval---quality outcome. (1) is a fit with $Pr_{1b}$}
\label{tab:posteriorgeval1}
\begin{tabular}{llllll}
\toprule
\centering
Effect & Term & Estimate & std.error & conf.low & conf.high \\
\midrule
fixed & (Intercept) & 2.24 & 0.16 & 1.92 & 2.55 \\
fixed & zoi (Intercept) & -2.79 & 0.65 & -4.11 & -1.56 \\
fixed & coi (Intercept) & 6.24 & 1.38 & 4.14 & 9.53 \\
fixed & RST1 & 0.18 & 0.08 & 0.02 & 0.33 \\
fixed & RST2 & 0.53 & 0.09 & 0.36 & 0.71 \\
fixed & GPT-4.1 & 0.07 & 0.14 & -0.20 & 0.33 \\
fixed & GPT-5.6 Luna High & 0.20 & 0.14 & -0.08 & 0.47 \\
fixed & GPT-5.6 Luna None & 0.19 & 0.14 & -0.08 & 0.46 \\
fixed & GPT-5.6 Sol High & 0.18 & 0.14 & -0.09 & 0.44 \\
fixed & GPT-5.6 Sol None & -0.07 & 0.13 & -0.33 & 0.20 \\
fixed & GPT-5.6 Terra High & 0.13 & 0.14 & -0.14 & 0.40 \\
fixed & GPT-5.6 Terra None & 0.21 & 0.14 & -0.06 & 0.48 \\
fixed & zoi RST1& 1.13 & 0.30 & 0.55 & 1.73 \\
fixed & zoi RST2& 1.61 & 0.30 & 1.02 & 2.20 \\
fixed & zoi GPT-4.1 & -0.25 & 0.45 & -1.13 & 0.63 \\
fixed & zoi GPT-5.6 Luna High & 0.63 & 0.45 & -0.24 & 1.51 \\
fixed & zoi GPT-5.6 Luna None & 0.09 & 0.45 & -0.78 & 0.96 \\
fixed & zoi GPT-5.6 Sol High & -0.36 & 0.45 & -1.28 & 0.52 \\
fixed & zoi GPT-5.6 Sol None & 0.20 & 0.44 & -0.67 & 1.08 \\
fixed & zoi GPT-5.6 Terra High & 0.09 & 0.45 & -0.78 & 0.96 \\
fixed & zoi GPT-5.6 Terra None & 0.41 & 0.44 & -0.48 & 1.26 \\
varying & sd (Intercept) & 0.58 & 0.10 & 0.42 & 0.81 \\
varying & sd zoi (Intercept) & 2.46 & 0.48 & 1.69 & 3.57 \\
\bottomrule
\end{tabular}
\end{table*}

\begin{table*}[h!]
\caption{Posterior Estimates for the Brms fits for automated assessment.G-Eval---quality outcome. (1) is a fit with $Pr_{2b}$}
\label{tab:posteriorgeval2}
\begin{tabular}{llllll}
\toprule
\centering
Effect & Term & Estimate & std.error & conf.low & conf.high \\
\midrule
fixed & (Intercept) & 2.26 & 0.16 & 1.95 & 2.57 \\
fixed & zoi (Intercept) & -2.68 & 0.61 & -3.95 & -1.53 \\
fixed & coi (Intercept) & 6.23 & 1.37 & 4.12 & 9.43 \\
fixed & RST1 & 0.17 & 0.08 & 0.01 & 0.33 \\
fixed & RST2 & 0.52 & 0.09 & 0.35 & 0.70 \\
fixed & GPT-4.1 & 0.06 & 0.13 & -0.20 & 0.32 \\
fixed & GPT-5.6 Luna High & 0.18 & 0.14 & -0.09 & 0.45 \\
fixed & GPT-5.6 Luna None & 0.17 & 0.14 & -0.10 & 0.44 \\
fixed & GPT-5.6 Sol High & 0.16 & 0.13 & -0.09 & 0.42 \\
fixed & GPT-5.6 Sol None & -0.08 & 0.13 & -0.34 & 0.18 \\
fixed & GPT-5.6 Terra High & 0.11 & 0.14 & -0.15 & 0.38 \\
fixed & GPT-5.6 Terra None & 0.20 & 0.14 & -0.07 & 0.47 \\
fixed & zoi RST1& 1.03 & 0.30 & 0.46 & 1.63 \\
fixed & zoi RST2& 1.51 & 0.31 & 0.91 & 2.13 \\
fixed & zoi GPT-4.1 & -0.26 & 0.38 & -1.03 & 0.48 \\
fixed & zoi GPT-5.6 Luna High & 0.52 & 0.38 & -0.21 & 1.30 \\
fixed & zoi GPT-5.6 Luna None & 0.04 & 0.38 & -0.72 & 0.77 \\
fixed & zoi GPT-5.6 Sol High & -0.36 & 0.39 & -1.15 & 0.39 \\
fixed & zoi GPT-5.6 Sol None & 0.13 & 0.38 & -0.60 & 0.87 \\
fixed & zoi GPT-5.6 Terra High & 0.03 & 0.37 & -0.71 & 0.74 \\
fixed & zoi GPT-5.6 Terra None & 0.33 & 0.38 & -0.41 & 1.08 \\
varying & sd (Intercept) & 0.57 & 0.10 & 0.41 & 0.79 \\
varying & sd zoi (Intercept) & 2.41 & 0.46 & 1.67 & 3.45 \\
\bottomrule
\end{tabular}
\end{table*}
\clearpage
\section{Pipeline Design}
\begin{figure}[h!]
    \centering
    \includegraphics[width=1\linewidth]{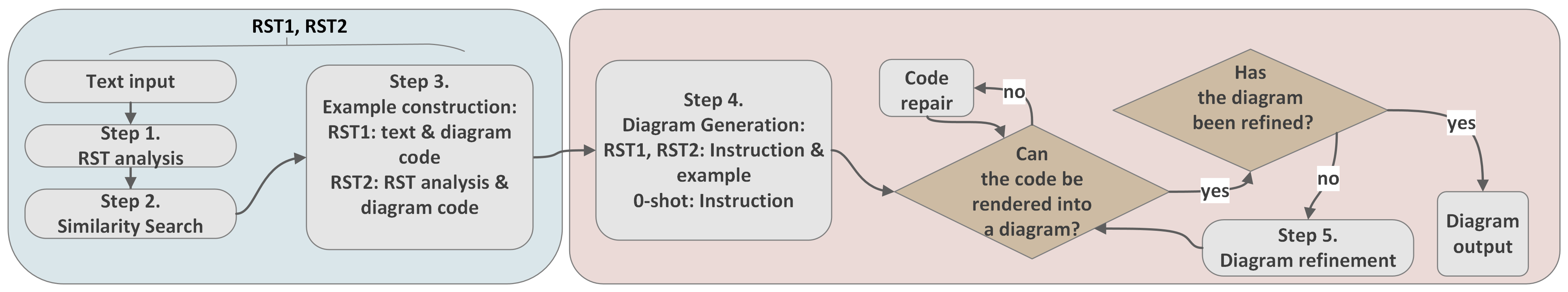}
    \caption{Diagram Generation Pipelines: RST1 and RST2 include Steps 1:5, 0-shot: Steps 4:5.}
    \label{fig:pipeline}
\end{figure}

\begin{algorithm}[hbt!]
\caption{In-Context Example Selection}\label{alg1}
\begin{algorithmic}
\Require Source text $s_i$; Premade examples $D_x = [(o_z,a_z,d_z)]_{z=1}^M, M=4$; Example text analyses $A=[a_1,...,a_4]$; Prompt instructions $R_1$ (RST analysis), $R_2$ (similarity search)
\Ensure ICL Examples $E_{x1}, E_{x2}$, RST analysis $a_i$
\item[Step 1] RST analysis
\Function{ANALYSIS}{$R_1; s_i$} 
\State $M_{sys}\gets R_1; M_{q} \gets s_i$
\Comment{The system message is $M_{sys}$, the user query is $M_q$}
\State \Return $a_i \gets$analysisLLM$(M_{sys}, M_q)$
\EndFunction
\item[Step 2] Similarity Search
\Function{SIMILARITY}{$R_2, A, a_i$}
\State $M_{sys}\gets R_2 \oplus A; M_{q} \gets a_i$ 
\State \Return $z \gets \argmax_{q \in [4]}\operatorname{simLLM}(M_{sys}; M_{q})$ 
\EndFunction
\item[Step 3] Example Construction
\Function{EXAMPLE}{$z, D_x$}
\State \Return $E_{x1} \gets \{d_z, o_z\}$, $E_{x2} \gets \{d_z, a_z\}$ 
\Comment{$o_z$ is a text, $a_z$ is an RST analysis}
\EndFunction
\end{algorithmic}
\end{algorithm}
\begin{algorithm}[hbt!]
\caption{Diagram Generation}\label{alg2}
\begin{algorithmic}[1]
\Require Source text $s_i$; ICL Examples $E_{x1}, E_{x2}$; RST analysis $a_i$; Prompt instructions for diagram code generation $R_{3rst1}$, $R_{3rst2}$, $R_{30}$
\Ensure Dot code $C_{i}$
\Function{GENERATION}{$R_{3rst1}$, $R_{3rst2}$, $R_{30}$, $R_5$, $a_i$, $s_i$, $E_{x1}, E_{x2}$}
\If{task is RST1}
    \State $M_{sys} \gets R_{3rst1} \oplus E_{x1}; M_{q} \gets s_i$
\ElsIf{task is RST2}
    \State $M_{sys} \gets R_{3rst2} \oplus E_{x2}; M_{q} \gets a_i$
\ElsIf{task is 0-shot}
    \State $M_{sys} \gets R_{30}; M_{q} \gets s_i$
\EndIf
\State \Return $C_{i}$ $\gets$ $generateLLM(M_{sys},M_{q})$
\EndFunction
\end{algorithmic}
\end{algorithm}
\begin{algorithm}[hbt!]
\caption{Diagram Code Render and Repair}\label{alg3}
\begin{algorithmic}[1]
\Require Diagram dot code $C$; Prompt instructions for dot code repair $R_5$
\Ensure Diagram dot code $C$; Image $I$
\Function{IMAGE}{$R_5$, $C$}
\State $attempts \gets 0$
\While{$attempts\leq5$}
    \State $I \gets renderGraphViz(C$) 
    \Comment{Render $C$ into an image $I$}
    \If{$I\neq\perp$}
        \State \Return $C, I$
    \ElsIf{$attempts<5$}
        \State $e \gets catchError(C);$ 
        \Comment{Catch error $e$ if $C$ cannot be rendered}
        \State $M_{sys}\gets R_5, M_q\gets C\oplus e$ 
        \Comment{pass $e$ and $C$ into the user message}
        \State $C_{c} \gets repaircodeLLM(M_{sys},M_q)$; 
        \Comment{$C_c$ is repaired code}
        \State $C \gets C_{c}$ \
        \State $attempts \gets attempts +1$
    \Else
        \State \Return $\perp$
    \EndIf
\EndWhile
\EndFunction
\end{algorithmic}
\end{algorithm}
\begin{algorithm}[hbt!]
\caption{Diagram Refinement}\label{alg4}
\begin{algorithmic}[1]
\Require Dot code $C_{i}$; Image $I_i$; Prompt instructions for diagram refinement $R_4$ 
\Ensure Dot code $C_{refi}$
\Function{REFINEMENT}{$R_4$, $C_{i}$, $I_i$}
\State $M_{sys} \gets R_4$; $M_q \gets $ $C_{i} \oplus I_i$
\State \Return $C_{refi}, T_{explain}$ $\gets$ $refineLLM(M_{sys},M_{q})$
\EndFunction
\end{algorithmic}
\end{algorithm}
\clearpage
\section{Hallucination Evaluation}
\begin{table}[ht!]
    \caption{Hallucination analysis description. X means incompliance, $I_i$ is a first diagram version, $I_{refi}$ is its refined version.}
    \label{tab:halluc_descr}
    \centering
    \begin{tabular}{lcccl}
    \hline
         \textbf{H}&  \textbf{Where}&  \textbf{Metric}& \textbf{Type} &Description\\\hline
         $F_{fact}$&  $I_{refi}$&  $\{0,1\}$& factual  &Factual fabrication or contradiction\\
          \multirow{2}{*}{$F_{ae}$}&  \multirow{2}{*}{$I_{refi}$}&  \multirow{2}{*}{$\{0,1\}$}&  X  &Layout issues described in \ref{c3}\\& & &with $R_3 \vee R_4$ & the model should avoid and correct\\
         \multirow{2}{*}{$F_{log}$} &  \multirow{2}{*}{$I_{refi}$}&  \multirow{2}{*}{$\{0,1\}$}& logical  &\multirow{2}{*}{Internal logical inconsistency}\\& & &inconsistency  &\\
         \multirow{2}{*}{$F_c$} &  \multirow{2}{*}{$I_{refi}$}&  \multirow{2}{*}{$\{0,1\}$}& X  &\multirow{2}{*}{Deviation from a given context}\\& & &with $s_i$ or  $a_i$ &\\
         \multirow{2}{*}{$F_{ref}$} &  \multirow{2}{*}{$I_{i},I_{refi}$}&  \multirow{2}{*}{$\{0,1\}$}& X &Layout issues described in \ref{c3}\\& & &with $R_4$ &not fixed or new introduced\\
         \multirow{2}{*}{$F_{inh}$}&  \multirow{2}{*}{$I_{refi}$}&  \multirow{2}{*}{$\{0,1\}$}& passed from  &\multirow{2}{*}{Hallucination passed from earlier outputs}\\& & &$F_{rst} \vee F_{sim}$ &\\
         $F_{rst}$&  $a_i$&  $\{0,1\}$& X $R_1$ &Hallucination in RST\\
  $F_{sim}$& $z$& $\{0,1\}$&X $R_2$ &Hallucination in similarity\\
  \bottomrule
         \end{tabular}
\end{table}
\begin{table}[ht!]
\caption{Hallucination occurrence and the ratio of $G$ (high quality diagrams).}
    \label{tab:hallucination}
\begin{tabular}{lllllll}
\toprule
&\multicolumn{3}{c}{GPT-4o}&\multicolumn{3}{c}{o3}\\
\cmidrule{2-4}\cmidrule{5-7}
&0-shot& RST1& RST2&0-shot& RST1& RST2\\
\midrule
$F_{fact}$ &0.2&  0.2&  0.04&0.12& 0.04& 0.04\\
$F_{ae}$ &0.6&  0.72&  0.8&0.44& 0.44& 0.36\\
$F_{c}$ &0.68&  0.52&  0.48&0.6& 0.36& 0.28\\
$F_{log}$ &0.28& 0.28&  0.48&0.2& 0& 0.12\\
 $F_{inh}$ && 0.28&0.52&-& 0.08&0.24\\
 $F_{rst}, F_{sim}$& -& \multicolumn{2}{c}{1,0.24}& -& \multicolumn{2}{c}{0.64, 0.12}\\
 $F_{ref}$ &0.72& 0.84& 0.72&0.52&  0.52& 0.48\\
 $G$ &0.64& 0.6&0.4&0.76& 0.84&0.8\\
\bottomrule
\end{tabular}
\end{table}
\clearpage
\section{Automated Evaluation}
\begin{table}[ht]
\centering\caption{BERTScore for GPT-5.6 Luna Performance for RST analysis to original text reconstruction}
\begin{tabular}{lrrr}
\toprule
 & Precision & Recall & F1 Score\\
\midrule

Mean & 0.9503 & 0.9393 & 0.9447 \\
SD & 0.0603 & 0.0838 & 0.0685 \\
Min & 0.7748 & 0.6484 & 0.7432 \\
25\% & 0.9313 & 0.9054 & 0.9030 \\
50\% & 0.9764 & 0.9752 & 0.9784 \\
75\% & 1.0000 & 1.0000 & 1.0000 \\
Max & 1.0000 & 1.0000 & 1.0000 \\
\bottomrule
\end{tabular}
\end{table}
\begin{table}[ht!]
\caption{G-Eval descriptive results across the model configurations} 
\centering
\begin{tabular}{lllrrrrrrr}
  \toprule
model & reasoning & method & min & q1 & q3 & SD & mean & q2 & max \\ 
  \midrule
gpt-4.1 &  & 0-SHOT & 0.66 & 0.84 & 0.92 & 0.08 & 0.89 & 0.90 & 1.00 \\ 
  gpt-4.1 &  & RST1 & 0.80 & 0.90 & 1.00 & 0.07 & 0.94 & 0.98 & 1.00 \\ 
  gpt-4.1 &  & RST2 & 0.78 & 0.92 & 1.00 & 0.06 & 0.96 & 1.00 & 1.00 \\ 
  gpt-4.1-mini &  & 0-SHOT & 0.69 & 0.83 & 0.98 & 0.09 & 0.89 & 0.87 & 1.00 \\ 
  gpt-4.1-mini &  & RST1 & 0.79 & 0.87 & 1.00 & 0.08 & 0.93 & 0.97 & 1.00 \\ 
  gpt-4.1-mini &  & RST2 & 0.76 & 0.92 & 1.00 & 0.06 & 0.96 & 0.99 & 1.00 \\ 
  gpt-5.6-luna & high & 0-SHOT & 0.79 & 0.89 & 1.00 & 0.07 & 0.93 & 0.94 & 1.00 \\ 
  gpt-5.6-luna & high & RST1 & 0.81 & 0.91 & 1.00 & 0.06 & 0.95 & 0.96 & 1.00 \\ 
  gpt-5.6-luna & high & RST2 & 0.66 & 0.97 & 1.00 & 0.08 & 0.96 & 1.00 & 1.00 \\ 
  gpt-5.6-luna & none & 0-SHOT & 0.81 & 0.88 & 1.00 & 0.06 & 0.93 & 0.91 & 1.00 \\ 
  gpt-5.6-luna & none & RST1 & 0.79 & 0.90 & 1.00 & 0.06 & 0.94 & 0.98 & 1.00 \\ 
  gpt-5.6-luna & none & RST2 & 0.69 & 0.90 & 1.00 & 0.08 & 0.95 & 0.99 & 1.00 \\ 
  gpt-5.6-sol & high & 0-SHOT & 0.79 & 0.90 & 0.99 & 0.07 & 0.93 & 0.92 & 1.00 \\ 
  gpt-5.6-sol & high & RST1 & 0.79 & 0.85 & 1.00 & 0.08 & 0.93 & 0.95 & 1.00 \\ 
  gpt-5.6-sol & high & RST2 & 0.84 & 0.91 & 1.00 & 0.05 & 0.96 & 0.99 & 1.00 \\ 
  gpt-5.6-sol & none & 0-SHOT & 0.79 & 0.90 & 1.00 & 0.06 & 0.92 & 0.91 & 1.00 \\ 
  gpt-5.6-sol & none & RST1 & 0.73 & 0.83 & 1.00 & 0.09 & 0.92 & 0.95 & 1.00 \\ 
  gpt-5.6-sol & none & RST2 & 0.80 & 0.89 & 1.00 & 0.07 & 0.95 & 0.99 & 1.00 \\ 
  gpt-5.6-terra & high & 0-SHOT & 0.76 & 0.88 & 0.98 & 0.07 & 0.92 & 0.92 & 1.00 \\ 
  gpt-5.6-terra & high & RST1 & 0.78 & 0.87 & 1.00 & 0.08 & 0.93 & 0.99 & 1.00 \\ 
  gpt-5.6-terra & high & RST2 & 0.65 & 0.91 & 1.00 & 0.08 & 0.94 & 0.98 & 1.00 \\ 
  gpt-5.6-terra & none & 0-SHOT & 0.80 & 0.90 & 1.00 & 0.06 & 0.94 & 0.94 & 1.00 \\ 
  gpt-5.6-terra & none & RST1 & 0.80 & 0.90 & 1.00 & 0.07 & 0.94 & 1.00 & 1.00 \\ 
  gpt-5.6-terra & none & RST2 & 0.82 & 0.90 & 1.00 & 0.06 & 0.95 & 0.99 & 1.00 \\ 
   \bottomrule
\end{tabular}
\label{desc_aut}
\end{table}

\begin{table*}[ht]
\centering\caption{Faithfulness descriptive results across the model configurations} 
\begin{tabular}{lllrrrrrrr}
  \hline
model & reasoning & method & min & q1 & q3 & SD & mean & median & max \\ 
  \hline
gpt-4.1 &  & 0-SHOT & 0.92 & 1.00 & 1.00 & 0.02 & 1.00 & 1.00 & 1.00 \\ 
  gpt-4.1 &  & RST1 & 0.75 & 1.00 & 1.00 & 0.06 & 0.98 & 1.00 & 1.00 \\ 
  gpt-4.1 &  & RST2 & 1.00 & 1.00 & 1.00 & 0.00 & 1.00 & 1.00 & 1.00 \\ 
  gpt-4.1-mini &  & 0-SHOT & 0.50 & 1.00 & 1.00 & 0.11 & 0.97 & 1.00 & 1.00 \\ 
  gpt-4.1-mini &  & RST1 & 0.83 & 1.00 & 1.00 & 0.03 & 0.99 & 1.00 & 1.00 \\ 
  gpt-4.1-mini &  & RST2 & 0.90 & 1.00 & 1.00 & 0.02 & 1.00 & 1.00 & 1.00 \\ 
  gpt-5.6-luna & high & 0-SHOT & 0.88 & 1.00 & 1.00 & 0.03 & 0.99 & 1.00 & 1.00 \\ 
  gpt-5.6-luna & high & RST1 & 1.00 & 1.00 & 1.00 & 0.00 & 1.00 & 1.00 & 1.00 \\ 
  gpt-5.6-luna & high & RST2 & 0.88 & 1.00 & 1.00 & 0.03 & 0.99 & 1.00 & 1.00 \\ 
  gpt-5.6-luna & none & 0-SHOT & 0.45 & 1.00 & 1.00 & 0.11 & 0.98 & 1.00 & 1.00 \\ 
  gpt-5.6-luna & none & RST1 & 0.90 & 1.00 & 1.00 & 0.02 & 1.00 & 1.00 & 1.00 \\ 
  gpt-5.6-luna & none & RST2 & 0.90 & 1.00 & 1.00 & 0.02 & 1.00 & 1.00 & 1.00 \\ 
  gpt-5.6-sol & high & 0-SHOT & 0.77 & 1.00 & 1.00 & 0.06 & 0.98 & 1.00 & 1.00 \\ 
  gpt-5.6-sol & high & RST1 & 1.00 & 1.00 & 1.00 & 0.00 & 1.00 & 1.00 & 1.00 \\ 
  gpt-5.6-sol & high & RST2 & 1.00 & 1.00 & 1.00 & 0.00 & 1.00 & 1.00 & 1.00 \\ 
  gpt-5.6-sol & none & 0-SHOT & 0.00 & 1.00 & 1.00 & 0.20 & 0.96 & 1.00 & 1.00 \\ 
  gpt-5.6-sol & none & RST1 & 1.00 & 1.00 & 1.00 & 0.00 & 1.00 & 1.00 & 1.00 \\ 
  gpt-5.6-sol & none & RST2 & 1.00 & 1.00 & 1.00 & 0.00 & 1.00 & 1.00 & 1.00 \\ 
  gpt-5.6-terra & high & 0-SHOT & 0.90 & 1.00 & 1.00 & 0.02 & 1.00 & 1.00 & 1.00 \\ 
  gpt-5.6-terra & high & RST1 & 0.90 & 1.00 & 1.00 & 0.02 & 1.00 & 1.00 & 1.00 \\ 
  gpt-5.6-terra & high & RST2 & 1.00 & 1.00 & 1.00 & 0.00 & 1.00 & 1.00 & 1.00 \\ 
  gpt-5.6-terra & none & 0-SHOT & 1.00 & 1.00 & 1.00 & 0.00 & 1.00 & 1.00 & 1.00 \\ 
  gpt-5.6-terra & none & RST1 & 1.00 & 1.00 & 1.00 & 0.00 & 1.00 & 1.00 & 1.00 \\ 
  gpt-5.6-terra & none & RST2 & 1.00 & 1.00 & 1.00 & 0.00 & 1.00 & 1.00 & 1.00 \\ 
   \hline
\end{tabular}
\end{table*}

\begin{table*}[ht]
\centering\caption{BERTScore F1 descriptive results across the model configurations} 
\begin{tabular}{lllrrrrrrr}
  \toprule
model & reasoning & method & min & q1 & q3 & SD & mean & median & max \\ 
  \midrule
gpt-4.1 &  & 0-SHOT & 0.03 & 0.19 & 0.31 & 0.11 & 0.25 & 0.24 & 0.51 \\ 
  gpt-4.1 &  & RST1 & 0.10 & 0.24 & 0.38 & 0.11 & 0.30 & 0.32 & 0.50 \\ 
  gpt-4.1 &  & RST2 & 0.31 & 0.37 & 0.55 & 0.11 & 0.46 & 0.45 & 0.71 \\ 
  gpt-4.1-mini &  & 0-SHOT & 0.01 & 0.16 & 0.30 & 0.13 & 0.24 & 0.24 & 0.52 \\ 
  gpt-4.1-mini &  & RST1 & 0.07 & 0.28 & 0.40 & 0.12 & 0.35 & 0.34 & 0.57 \\ 
  gpt-4.1-mini &  & RST2 & 0.30 & 0.40 & 0.65 & 0.16 & 0.54 & 0.50 & 0.86 \\ 
  gpt-5.6-luna & high & 0-SHOT & 0.06 & 0.18 & 0.33 & 0.14 & 0.29 & 0.27 & 0.64 \\ 
  gpt-5.6-luna & high & RST1 & 0.11 & 0.31 & 0.49 & 0.16 & 0.38 & 0.39 & 0.66 \\ 
  gpt-5.6-luna & high & RST2 & 0.27 & 0.47 & 0.63 & 0.14 & 0.54 & 0.58 & 0.78 \\ 
  gpt-5.6-luna & none & 0-SHOT & 0.06 & 0.17 & 0.30 & 0.12 & 0.26 & 0.25 & 0.52 \\ 
  gpt-5.6-luna & none & RST1 & 0.14 & 0.27 & 0.41 & 0.11 & 0.35 & 0.35 & 0.58 \\ 
  gpt-5.6-luna & none & RST2 & 0.31 & 0.40 & 0.63 & 0.13 & 0.53 & 0.56 & 0.78 \\ 
  gpt-5.6-sol & high & 0-SHOT & 0.06 & 0.23 & 0.40 & 0.15 & 0.31 & 0.26 & 0.70 \\ 
  gpt-5.6-sol & high & RST1 & 0.11 & 0.27 & 0.46 & 0.12 & 0.36 & 0.39 & 0.57 \\ 
  gpt-5.6-sol & high & RST2 & 0.17 & 0.32 & 0.53 & 0.11 & 0.40 & 0.37 & 0.57 \\ 
  gpt-5.6-sol & none & 0-SHOT & 0.05 & 0.18 & 0.34 & 0.13 & 0.25 & 0.24 & 0.61 \\ 
  gpt-5.6-sol & none & RST1 & 0.05 & 0.25 & 0.42 & 0.13 & 0.34 & 0.36 & 0.58 \\ 
  gpt-5.6-sol & none & RST2 & 0.13 & 0.29 & 0.48 & 0.13 & 0.38 & 0.38 & 0.61 \\ 
  gpt-5.6-terra & high & 0-SHOT & 0.05 & 0.20 & 0.36 & 0.14 & 0.28 & 0.29 & 0.58 \\ 
  gpt-5.6-terra & high & RST1 & 0.12 & 0.23 & 0.45 & 0.13 & 0.34 & 0.32 & 0.57 \\ 
  gpt-5.6-terra & high & RST2 & 0.28 & 0.32 & 0.47 & 0.10 & 0.40 & 0.36 & 0.60 \\ 
  gpt-5.6-terra & none & 0-SHOT & 0.06 & 0.20 & 0.36 & 0.13 & 0.27 & 0.26 & 0.58 \\ 
  gpt-5.6-terra & none & RST1 & 0.16 & 0.24 & 0.44 & 0.11 & 0.35 & 0.37 & 0.56 \\ 
  gpt-5.6-terra & none & RST2 & 0.26 & 0.32 & 0.46 & 0.11 & 0.41 & 0.42 & 0.64 \\ 
   \bottomrule
\end{tabular}
\end{table*}

\begin{table*}[ht]
\centering\caption{ROUGE-L descriptive results across the model configurations} 
\begin{tabular}{lllrrrrrrr}
  \hline
model & reasoning & method & min & q1 & q3 & SD & mean & median & max \\ 
  \hline
gpt-4.1 &  & 0-SHOT & 0.18 & 0.29 & 0.43 & 0.12 & 0.37 & 0.36 & 0.66 \\ 
  gpt-4.1 &  & RST1 & 0.19 & 0.38 & 0.57 & 0.13 & 0.46 & 0.48 & 0.69 \\ 
  gpt-4.1 &  & RST2 & 0.44 & 0.58 & 0.73 & 0.12 & 0.65 & 0.68 & 0.84 \\ 
  gpt-4.1-mini &  & 0-SHOT & 0.17 & 0.32 & 0.48 & 0.12 & 0.40 & 0.41 & 0.63 \\ 
  gpt-4.1-mini &  & RST1 & 0.27 & 0.37 & 0.59 & 0.14 & 0.51 & 0.51 & 0.77 \\ 
  gpt-4.1-mini &  & RST2 & 0.43 & 0.68 & 0.86 & 0.14 & 0.75 & 0.76 & 0.96 \\ 
  gpt-5.6-luna & high & 0-SHOT & 0.19 & 0.39 & 0.54 & 0.15 & 0.46 & 0.49 & 0.75 \\ 
  gpt-5.6-luna & high & RST1 & 0.28 & 0.44 & 0.64 & 0.15 & 0.55 & 0.53 & 0.79 \\ 
  gpt-5.6-luna & high & RST2 & 0.42 & 0.66 & 0.81 & 0.13 & 0.72 & 0.76 & 0.92 \\ 
  gpt-5.6-luna & none & 0-SHOT & 0.20 & 0.37 & 0.51 & 0.14 & 0.44 & 0.46 & 0.70 \\ 
  gpt-5.6-luna & none & RST1 & 0.26 & 0.47 & 0.59 & 0.13 & 0.53 & 0.52 & 0.80 \\ 
  gpt-5.6-luna & none & RST2 & 0.39 & 0.70 & 0.79 & 0.12 & 0.73 & 0.75 & 0.99 \\ 
  gpt-5.6-sol & high & 0-SHOT & 0.21 & 0.35 & 0.55 & 0.15 & 0.46 & 0.45 & 0.77 \\ 
  gpt-5.6-sol & high & RST1 & 0.20 & 0.45 & 0.58 & 0.13 & 0.50 & 0.52 & 0.72 \\ 
  gpt-5.6-sol & high & RST2 & 0.31 & 0.45 & 0.60 & 0.12 & 0.54 & 0.56 & 0.77 \\ 
  gpt-5.6-sol & none & 0-SHOT & 0.16 & 0.31 & 0.54 & 0.13 & 0.42 & 0.41 & 0.69 \\ 
  gpt-5.6-sol & none & RST1 & 0.23 & 0.43 & 0.56 & 0.12 & 0.48 & 0.49 & 0.70 \\ 
  gpt-5.6-sol & none & RST2 & 0.29 & 0.46 & 0.61 & 0.12 & 0.53 & 0.55 & 0.77 \\ 
  gpt-5.6-terra & high & 0-SHOT & 0.17 & 0.35 & 0.56 & 0.15 & 0.44 & 0.42 & 0.72 \\ 
  gpt-5.6-terra & high & RST1 & 0.29 & 0.45 & 0.63 & 0.13 & 0.52 & 0.51 & 0.76 \\ 
  gpt-5.6-terra & high & RST2 & 0.38 & 0.51 & 0.64 & 0.10 & 0.57 & 0.56 & 0.73 \\ 
  gpt-5.6-terra & none & 0-SHOT & 0.21 & 0.38 & 0.55 & 0.14 & 0.45 & 0.41 & 0.74 \\ 
  gpt-5.6-terra & none & RST1 & 0.30 & 0.42 & 0.60 & 0.11 & 0.52 & 0.54 & 0.76 \\ 
  gpt-5.6-terra & none & RST2 & 0.29 & 0.49 & 0.63 & 0.12 & 0.56 & 0.56 & 0.81 \\ 
   \hline
\end{tabular}
\end{table*}

\begin{table*}[ht]
\caption{G-Eval descriptive results across individual texts}
\label{aut_ids}
\centering
\begin{tabular}{lrrrrrrr}
  \toprule
ids & min & q1 & q3 & SD & mean & median & max \\ 
  \midrule
1 & 0.80 & 0.81 & 0.92 & 0.07 & 0.88 & 0.86 & 1.00 \\ 
  2 & 0.87 & 1.00 & 1.00 & 0.03 & 0.99 & 1.00 & 1.00 \\ 
  3 & 0.80 & 0.90 & 0.98 & 0.06 & 0.93 & 0.94 & 1.00 \\ 
  4 & 0.90 & 0.99 & 1.00 & 0.03 & 0.98 & 1.00 & 1.00 \\ 
  5 & 0.84 & 0.95 & 1.00 & 0.05 & 0.97 & 1.00 & 1.00 \\ 
  6 & 0.91 & 0.99 & 1.00 & 0.02 & 0.99 & 1.00 & 1.00 \\ 
  7 & 0.81 & 0.89 & 0.98 & 0.06 & 0.92 & 0.91 & 1.00 \\ 
  8 & 0.78 & 0.85 & 0.91 & 0.06 & 0.88 & 0.88 & 0.98 \\ 
  9 & 0.80 & 0.84 & 0.94 & 0.07 & 0.90 & 0.90 & 1.00 \\ 
  10 & 0.65 & 0.98 & 1.00 & 0.08 & 0.97 & 1.00 & 1.00 \\ 
  11 & 0.82 & 0.88 & 0.94 & 0.05 & 0.91 & 0.90 & 1.00 \\ 
  12 & 0.66 & 0.81 & 0.90 & 0.07 & 0.85 & 0.85 & 0.93 \\ 
  13 & 0.76 & 0.94 & 1.00 & 0.07 & 0.96 & 1.00 & 1.00 \\ 
  14 & 0.79 & 0.97 & 1.00 & 0.05 & 0.97 & 1.00 & 1.00 \\ 
  15 & 0.73 & 0.80 & 0.90 & 0.06 & 0.85 & 0.85 & 0.98 \\ 
  16 & 0.89 & 1.00 & 1.00 & 0.03 & 0.99 & 1.00 & 1.00 \\ 
  17 & 0.87 & 0.92 & 0.99 & 0.04 & 0.96 & 0.97 & 1.00 \\ 
  18 & 0.83 & 0.89 & 1.00 & 0.06 & 0.94 & 0.95 & 1.00 \\ 
  19 & 0.76 & 0.80 & 0.94 & 0.07 & 0.88 & 0.90 & 1.00 \\ 
  20 & 0.93 & 0.98 & 1.00 & 0.02 & 0.99 & 1.00 & 1.00 \\ 
  21 & 0.66 & 0.80 & 0.89 & 0.06 & 0.84 & 0.84 & 0.93 \\ 
  22 & 0.90 & 0.99 & 1.00 & 0.03 & 0.98 & 1.00 & 1.00 \\ 
  23 & 0.69 & 0.98 & 1.00 & 0.07 & 0.97 & 1.00 & 1.00 \\ 
  24 & 0.79 & 0.88 & 0.98 & 0.06 & 0.92 & 0.91 & 1.00 \\ 
  25 & 0.82 & 0.91 & 0.99 & 0.05 & 0.95 & 0.97 & 1.00 \\ 
   \bottomrule
\end{tabular} 
\end{table*}

\begin{table*}[ht]
\centering\caption{GPT-5.6 Sol Model Performance Statistics for $G$: the positive class is $0$.} 
\begin{tabular}{lr}
  \toprule
Metric & Value \\ 
  \midrule
Accuracy estimate, 95\% CI & 0.7733 (0.6979,  0.8376) \\ 
  Kappa & 0.4032 \\ 
  No Information Rate (NIR) & 0.6733 \\ 
  P-Value [Accuracy $>$ NIR] & 0.0048 \\ 
  McNemar's Test P-Value & 0.0000 \\ 
  Sensitivity (Recall) & 0.3878 \\ 
  Specificity & 0.9604 \\ 
  Positive Predictive Value (Precision) & 0.8261 \\ 
  Negative Predictive Value & 0.7638 \\ 
  Precision Alternate & 0.8261 \\ 
  Recall Alternate & 0.3878 \\ 
  F1 Score & 0.5278 \\ 
  F1 Score macro & 0.6893 \\ 
  Prevalence & 0.3267 \\ 
  Detection Rate & 0.1267 \\ 
  Detection Prevalence & 0.1533 \\ 
  Balanced Accuracy & 0.6741 \\ 
   \bottomrule
\end{tabular}
\end{table*}
\end{document}